\documentclass[10pt,journal,compsoc]{IEEEtran}

\ifCLASSOPTIONcompsoc
  \usepackage[nocompress,sort]{cite}
\else
  \usepackage{cite}
\fi 

\usepackage[utf8]{inputenc} % allow utf-8 input
\usepackage[T1]{fontenc}    % use 8-bit T1 fonts
\usepackage{url}            % simple URL typesetting
\usepackage{booktabs}       % professional-quality tables
\usepackage{amsfonts}       % blackboard math symbols
\usepackage{amsmath}        % blackboard math symbols
\usepackage{nicefrac}       % compact symbols for 1/2, etc.
\usepackage{microtype}      % microtypography
\usepackage{graphicx}
\usepackage{makecell}
\usepackage{multirow}
\usepackage{subcaption}
\newcommand{\ra}[1]{\renewcommand{\arraystretch}{#1}}
\usepackage[pagebackref=false,breaklinks=true,colorlinks,urlcolor=blue,citecolor=blue,linkcolor=blue,bookmarks=false]{hyperref}
\usepackage{tikz}
\usepackage{algorithmic}
\usepackage[linesnumbered,ruled,vlined]{algorithm2e}

\DeclareMathAlphabet      {\mathbfit}{OML}{cmm}{b}{it}

\usepackage{xcolor}
\usepackage[export]{adjustbox}

\usepackage{pifont}% http://ctan.org/pkg/pifont
\newcommand{\etal}{\textit{et~al}\mbox{.}}

\newcommand{\ie}{i.e.,\ }
\newcommand{\bbR}{{\mathbb{R}}}

\newcommand{\revised}[1]{#1}

\newcommand{\revision}[1]{\textcolor{black}{#1}}

\newlength{\twoimg}
\newlength{\threeimg}

\newlength{\elevenimg}

\newlength\paramargin
\newlength\figmargin
\newlength\secmargin
\newlength\figcapmargin

\begin{document}

\title{Cross-Resolution Adversarial Dual Network for Person Re-Identification and Beyond}

\author{Yu-Jhe~Li$^{*}$, Yun-Chun~Chen$^{*}$, Yen-Yu~Lin, and~Yu-Chiang~Frank~Wang
\IEEEcompsocitemizethanks{
\IEEEcompsocthanksitem Y.-J. Li is with the Graduate Institute of Communication Engineering, National Taiwan University, Taipei, Taiwan, E-mail: d08942008@ntu.edu.tw
\IEEEcompsocthanksitem Y.-C. Chen is with the Department of Computer Science, University of Toronto, ON, Canada, E-mail: ycchen@cs.toronto.edu
\IEEEcompsocthanksitem Y.-Y. Lin is with the Department of Computer Science, National Chiao Tung University, Hsinchu 300, Taiwan, E-mail: lin@cs.nctu.edu.tw
\IEEEcompsocthanksitem Y.-C. F. Wang is with the Department of Electrical Engineering and Graduate Institute of Communication Engineering, National Taiwan University, Taipei, Taiwan, E-mail: ycwang@ntu.edu.tw
%\IEEEcompsocthanksitem \note{XXX} is the corresponding author.
\IEEEcompsocthanksitem[*] The first two authors contributed equally to this work.}
}

\markboth{}
{Shell \MakeLowercase{\textit{et al.}}: Bare Demo of IEEEtran.cls for Computer Society Journals}

\IEEEtitleabstractindextext{

\begin{abstract}
Person re-identification (re-ID) aims at matching images of the same person across camera views. Due to varying distances between cameras and persons of interest, resolution mismatch can be expected, which would degrade re-ID performance in real-world scenarios. To overcome this problem, we propose a novel generative adversarial network to address cross-resolution person re-ID, allowing query images with varying resolutions. By advancing adversarial learning techniques, our proposed model learns resolution-invariant image representations while being able to recover the missing details in low-resolution input images. The resulting features can be jointly applied for improving re-ID performance due to preserving resolution invariance and recovering re-ID oriented discriminative details. Extensive experimental results on five standard person re-ID benchmarks confirm the effectiveness of our method and the superiority over the state-of-the-art approaches, especially when the input resolutions are not seen during training. Furthermore, the experimental results on two vehicle re-ID benchmarks also confirm the generalization of our model on cross-resolution visual tasks. The extensions of semi-supervised settings further support the use of our proposed approach to real-world scenarios and applications.
\end{abstract}

}

\maketitle

\IEEEdisplaynontitleabstractindextext

\IEEEpeerreviewmaketitle

\section{Introduction}

\begin{figure}[!ht]
  \centering\includegraphics[width=\linewidth]{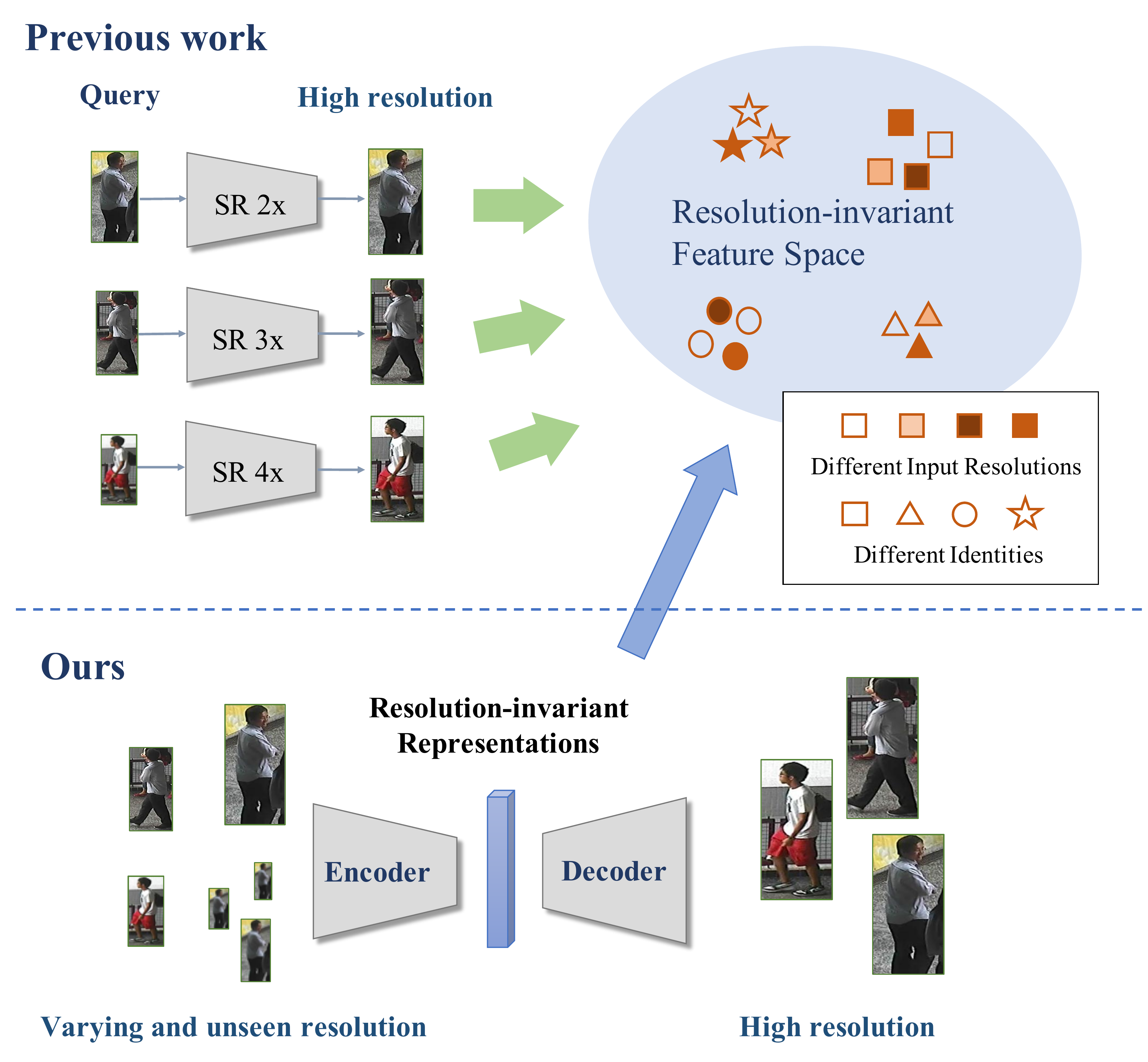}
  \vspace{-5.0mm}
  \caption{\textbf{Illustration and challenges of cross-resolution person re-ID.} (\emph{Top}) Existing methods for cross-resolution person re-ID typically leverage image super-resolution models with pre-defined up-sampling rates followed by person re-ID modules. Methods of this class, however, may not be applicable to query images of varying or unseen resolutions. (\emph{Bottom}) In contrast, our method learns resolution-invariant representations, allowing our model to re-identify persons in images of varying and even unseen resolutions.}
  \vspace{-4.0mm}
  \label{fig:teaser}
\end{figure}

\IEEEPARstart{P}{\revision{erson}} \revision{re-identification (re-ID)~\cite{zheng2016person,zhong2017re,wang2015zero,ye2020deep} aims at recognizing the same person across images taken by different cameras, and is an active research topic in computer vision and machine learning.} A variety of applications ranging from person tracking~\cite{andriluka2008people}, video surveillance system~\cite{khan2016person}, urban safety monitoring~\cite{garcia2015person}, to computational forensics~\cite{vezzani2013people} are highly correlated this research topic. Nevertheless, due to the presence of background clutter, occlusion, illumination or viewpoint changes, and even uncontrolled human pose variations, person re-ID remains a challenging task for practical applications.

Driven by the recent success of convolutional neural networks (CNNs), several learning based methods~\cite{lin2017improving,shen2018deep,hermans2017defense,zhong2017camera,si2018dual,chen2018group,zhang2019densely,hou2019interaction,zheng2019joint,zheng2019re} have been proposed to address the challenges in person re-ID. Despite promising performance, these methods are typically developed under the assumption that both query and gallery images are of \emph{similar} or \emph{sufficiently high} resolutions. This assumption, however, may not hold in practice since image resolutions would vary drastically due to the varying distances between cameras and persons of interest. For instance, query images captured by surveillance cameras are often of low resolution (LR) whereas those in the gallery set are carefully selected beforehand and are typically of high resolution (HR). As a result, direct matching of LR query images and HR gallery ones would lead to non-trivial \emph{resolution mismatch} problems~\cite{jing2015super,wang2016scale,jiao2018deep,wang2018cascaded}.

To address cross-resolution person re-ID, conventional methods typically learn a shared feature space for LR and HR images to mitigate the resolution mismatch problem~\cite{li2015multi,jing2015super,wang2016scale}. These approaches, however, adopt hand-engineered descriptors which cannot adapt themselves to the task at hand. The lack of an end-to-end learning pipeline might lead to sub-optimal person re-ID performance. To alleviate this issue, a number of approaches~\cite{wang2018cascaded,jiao2018deep} employing trainable descriptors are presented. These methods leverage image super-resolution (SR) models to convert LR input images into their HR versions, on which person re-ID is carried out. While performance improvements have been shown, these methods suffer from two limitations. First, each employed SR model is designed to upscale image resolutions by a particular factor. Therefore, these methods need to \emph{pre-determine} the resolutions of LR query images so that the corresponding SR models can be applied. However, designing SR models for each possible resolution input makes these methods hard to scale. Second, in real-world scenarios, query images can be with \emph{various} resolutions even with the resolutions that are \emph{not seen} during training. As illustrated in the top of Figure~\ref{fig:teaser}, query images with varying or unseen resolutions would restrict the applicability of the person re-ID methods that leverage SR models since one cannot assume the resolutions of the input images will be known in advance.

In this paper, we propose \emph{Cross-resolution Adversarial Dual Network} (CAD-Net) for cross-resolution person re-ID. The key characteristics of CAD-Net are two-fold. First, to address the resolution variations, CAD-Net derives the \emph{resolution-invariant representations} via adversarial learning. As shown in the bottom of Figure~\ref{fig:teaser}, the learned resolution-invariant representations allow our model to handle images of \emph{varying} and even \emph{unseen} resolutions. Second, CAD-Net learns to recover the missing details in LR input images. Together with the resolution-invariant features, our model generates HR images that are \emph{preferable for person re-ID}, achieving the state-of-the-art performance on cross-resolution person re-ID. It is worth noting that the above image resolution recovery and cross-resolution person re-ID are realized by a \emph{single} model learned in an \emph{end-to-end} fashion. 

Motivated by the multi-scale adversarial learning techniques in semantic segmentation~\cite{tsai2018learning} and person re-ID~\cite{chen2019learning}, which have been shown effective in deriving more robust feature representations, we employ multi-scale adversarial networks to align feature distributions between HR and LR images across different feature levels, resulting in consistent performance improvements over single-scale adversarial methods. On the other hand, since there are infinitely many HR images that reduce to the same LR image, it is difficult for a model to simultaneously handle the resolution variations and learn the mapping between LR and HR images. To alleviate this issue, we introduce a consistency loss in the HR feature space to enforce the consistency between the features of the recovered HR images and the corresponding HR ground-truth images, allowing our model to learn HR image representations that are more robust to the variations of HR image recovery. By jointly leveraging the above schemes, our method further improves the performance of cross-resolution person re-ID.

In addition to person re-ID, the resolution mismatch issue may occur in various applications such as vehicle re-ID~\cite{kanaci2018vehicle}. To demonstrate the wide applicability of our method, we also evaluate our method on cross-resolution vehicle re-ID and show that our CAD-Net performs favorably against existing cross-resolution vehicle re-ID approaches. Furthermore, to manifest that our formulation is not limited to cross-resolution setting, we show that our proposed algorithm improves person re-ID performance even when no significant resolution variations are present, achieving competitive performance compared to existing person re-ID approaches. Finally, as image labeling process is often labor intensive, we extend our CAD-Net to semi-supervised settings. Experimental results further support the use and extension of our method for such practical yet challenging settings.

The contributions of this paper are highlighted as follows: 

\begin{itemize}
  \item We propose an end-to-end trainable network which advances adversarial learning strategies for cross-resolution person re-ID.
  \item Our model learns resolution-invariant representations while being able to recover the missing details in LR input images, resulting in favorable performance in cross-resolution person re-ID.
  \item Our model is able to handle query images with varying or even unseen resolutions, without the need to pre-determine the input resolutions.
  \item Extensive experimental results on five person re-ID and two vehicle re-ID datasets show that our method achieves the state-of-the-art performance on both tasks in the cross-resolution setting, and further validate the effectiveness of our approach for real-world person re-ID applications in a semi-supervised manner.
\end{itemize}

In this work, we significantly extend our previous results~\cite{CAD-Net} and summarize the main differences in the following.

\begin{itemize}
  \item \textbf{Multi-scale adversarial learning for learning resolution-invariant representations.} Unlike our preliminary work that learns the resolution-invariant representations at a single scale, we adopt multi-scale adversarial network components in this work. The resultant model effectively aligns feature distributions in different levels and derives feature representations across image resolutions, achieving performance improvements over the single-scale model in cross-resolution person re-ID. We refer to our improved method as CAD-Net++.
  \item \textbf{HR feature space consistency loss.} To allow our model to handle the variations of HR image recovery, we introduce a feature consistency loss that enforces the consistency between the features of the recovered HR images and the corresponding HR ground-truth images. This loss further consistently improves the performance of cross-resolution person re-ID on all five datasets.
  \item \textbf{Applications.} In contrast to our preliminary work that focuses on a single task (i.e., cross-resolution person re-ID), we evaluate our proposed method under various settings with extensive ablation studies, including 1) cross-resolution person re-ID, 2) person re-ID with no significant resolution variations (we refer to this setting as standard person re-ID), 3) cross-resolution vehicle re-ID, and 4) semi-supervised cross-resolution person re-ID. Extensive experimental results confirm the effectiveness of our method in a wide range of scenarios.
\end{itemize}

\section{Related Work}

Person re-ID has been extensively studied in the literature. We review several topics relevant to our approach in this section.

{\flushleft {\bf Person re-ID.}}
Many existing methods, e.g., \cite{lin2017improving,shen2018deep,shen2018person,kalayeh2018human,cheng2016person,chang2018multi,chen2018group,li2018adaptation,sun2018beyond,suh2018part}, are developed to address various challenges in person re-ID, such as background clutter, viewpoint changes, and pose variations. For instance, Yang~\etal~\cite{zhong2017camera} learn a camera-invariant subspace to deal with the style variations caused by different cameras. Liu~\etal~\cite{liu2018pose} develop a pose-transferable framework based on generative adversarial network (GAN)~\cite{goodfellow2014generative} to yield pose-specific images for tackling pose variations. Several methods addressing background clutter leverage attention mechanisms to emphasize the discriminative parts~\cite{li2018harmonious,song2018mask,si2018dual}. \revision{In addition to these methods that learn global features, a few methods further utilize part-level information~\cite{sun2018beyond} to learn more fine-grained features, adopt human semantic parsing for learning local features~\cite{kalayeh2018human}, learn multi-scale re-ID representation~\cite{li2017person}, or derive part-aligned representations~\cite{suh2018part} for improving person re-ID.} 

Another research trend focuses on domain adaptation~\cite{ganin2015unsupervised,ganin2016domain,long2015learning,long2016unsupervised,chen2019crdoco,hoffman2017cycada} for person re-ID~\cite{wei2018person,image-image18,ge2018fd,li2019cross}. These methods either employ image-to-image translation modules (e.g., CycleGAN~\cite{zhu2017unpaired}) as a data augmentation technique to generate viewpoint specific images with labels~\cite{wei2018person,image-image18}, or leverage pose information to learn identity related but pose unrelated representations~\cite{ge2018fd} or pose-guided yet dataset-invariant representations~\cite{li2019cross} for cross-dataset person re-ID.

While promising performance has been demonstrated, the above approaches typically assume that both query and gallery images are of similar or sufficiently high resolutions, which might not be practical for real-world applications.

{\flushleft {\bf Cross-resolution person re-ID.}}
\revision{A number of methods have been proposed to address the resolution mismatch issue in person re-ID. These methods can be categorized into two groups depending on the adopted feature descriptors: 1) hand-crafted descriptor based methods~\cite{li2015multi,jing2015super,wang2016scale} and 2) trainable descriptor based methods~\cite{jiao2018deep,wang2018cascaded,chen2019learning,mao2019resolution,li2018toward}.} Methods in the first group typically use an engineered descriptor such as HOG~\cite{HoG} for feature extraction and then learn a shared feature space between HR and LR images. For instance, Li~\etal~\cite{li2015multi} jointly perform multi-scale distance metric learning and cross-scale image domain alignment. Jing~\etal~\cite{jing2015super} develop a semi-coupled low-rank dictionary learning framework to seek a mapping between HR and LR images. Wang~\etal~\cite{wang2016scale} learn a discriminating scale-distance function space by varying the image scale of LR images when matching with the HR ones. Nevertheless, these methods adopt hand-crafted descriptors, which cannot easily adapt the developed models to the tasks of interest, and thus may lead to sub-optimal person re-ID performance.

\revision{To alleviate this issue, several trainable descriptor based approaches are presented for cross-resolution person re-ID~\cite{jiao2018deep,wang2018cascaded,chen2019learning,mao2019resolution,li2018toward}.} The network of SING~\cite{jiao2018deep} is composed of several SR sub-networks and a person re-ID module to carry out LR person re-ID. On the other hand, CSR-GAN~\cite{wang2018cascaded} cascades multiple SR-GANs~\cite{ledig2017photo} and progressively recovers the details of LR images to address the resolution mismatch problem. Mao~\etal~\cite{mao2019resolution} develop a foreground-focus super-resolution model that learns to recover the resolution loss in LR input images followed by a resolution-invariant person re-ID module. In spite of their promising results, such methods rely on training pre-defined SR models~\cite{jiao2018deep,wang2018cascaded} or annotating the foreground mask for each training image to guide the learning of image recovery~\cite{mao2019resolution}. As mentioned earlier, the degree of resolution mismatch, i.e., the resolution difference between the query and gallery images, is typically \emph{unknown beforehand}. If the resolution of the input LR query is unseen during training, the above methods cannot be easily applied or might not lead to satisfactory performance. On the other hand, the dependence on foreground masks would make such methods hard to scale for real-world applications. Different from the above methods that employ SR models, a recent method motivated by the domain-invariant representations in domain adaptation~\cite{ganin2015unsupervised,ganin2016domain} is presented~\cite{chen2019learning}. By advancing adversarial learning strategies in the feature space, the RAIN method~\cite{chen2019learning} aligns the feature distributions of HR and LR images, allowing the model to be more robust to resolution variations.

Similar to RAIN~\cite{chen2019learning}, our method also performs feature distribution alignment between HR and LR images. Our model differs from RAIN~\cite{chen2019learning} in that our model learns to recover the missing details in LR input images and thus provides more discriminative evidence for person re-ID. By jointly observing features of both modalities in an end-to-end learning fashion, our model recovers HR images that are preferable for person re-ID, resulting in performance improvements on cross-resolution person re-ID. Experimental results demonstrate that our approach can be applied to input images of varying and even unseen resolutions using only a single model with favorable performance.

\begin{figure*}[!ht]
  \centering
  \includegraphics[width=\linewidth]{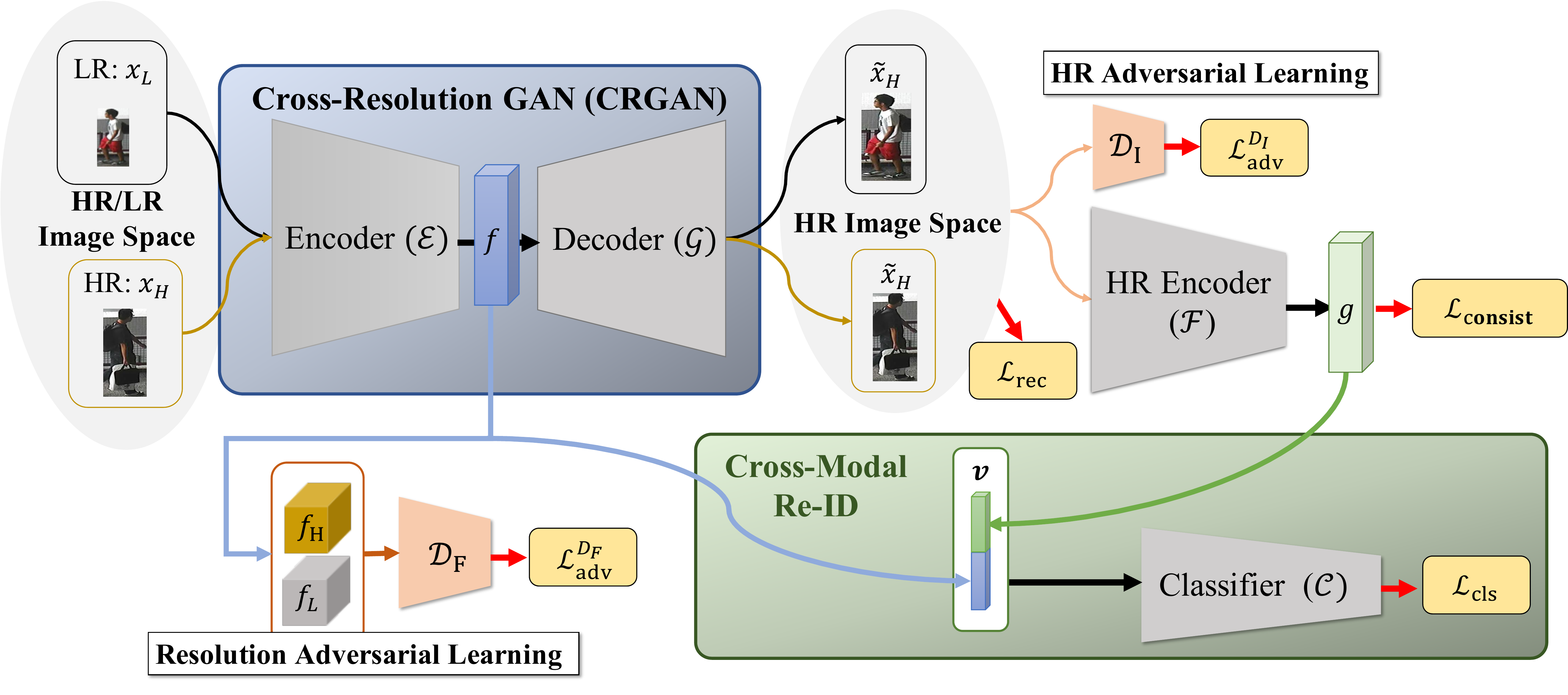}
  \vspace{-6.0mm}
  \caption{\textbf{Overview of the proposed Cross-resolution Adversarial Dual Network++ (CAD-Net++).} CAD-Net++ comprises cross-resolution GAN (CRGAN) (highlighted in blue) and cross-modal re-ID network (highlighted in green). The former learns resolution-invariant representations and recovers the missing details in LR input images, while the latter considers both feature modalities for cross-resolution person re-ID.}
  \vspace{-2.0mm}
  \label{fig:Model}
\end{figure*}

{\flushleft {\bf Cross-resolution vision applications.}}
The issues regarding cross-resolution handling have been studied in the literature. For face recognition, existing approaches typically rely on face hallucination algorithms~\cite{zhu2016deep,yu2017hallucinating} or SR mechanisms~\cite{kim2016accurate,dahl2017pixel,dong2016image} to super-resolve the facial details. Unlike the aforementioned methods that focus on synthesizing the facial details, our model learns to recover re-ID oriented discriminative details. For vehicle re-ID, the resolution mismatch issue is also a challenging yet under studied problem~\cite{kanaci2018vehicle}. While several efforts have been made~\cite{zhou2018aware,wang2017orientation,shen2017learning} to address the challenges (e.g., viewpoint or appearance variations) in vehicle re-ID, these resultant methods are developed under the assumption that both query and gallery images are of similar or sufficiently high resolutions. To carry out cross-resolution vehicle re-ID, MSVF~\cite{kanaci2018vehicle} designs a multi-branch network that learns a representation by fusing features from images of different scales. Our method differs from MSVF~\cite{kanaci2018vehicle} in three aspects. First, MSVF~\cite{kanaci2018vehicle} is tailored for cross-resolution vehicle re-ID while our method is developed to address cross-resolution person re-ID. Second, our model does not need to pre-determined the number of branches. Instead, our model carries out cross-resolution person re-ID using only a single model. Third, our model further learns to recover the missing details in LR input images. Through extensive experiments, we demonstrate that our algorithm performs favorably against existing cross-resolution vehicle re-ID approaches.

\section{Proposed Method}

In this section, we first provide an overview of our proposed approach. We then describe the details of each network component as well as the loss functions.

\subsection{Algorithmic Overview}

We first define the notations to be used in this paper. In the training stage, we assume we have access to a set of~$N$ HR images $X_H = \{x_i^H\}_{i=1}^N$ and its corresponding label set $Y_H = \{y_i^H\}_{i=1}^N$, where $x_i^H \in \bbR^{H \times W \times 3}$ and $y_i^H \in \bbR$ are the $i^\mathrm{th}$ HR image and its label, respectively. To allow our model to handle images of different resolutions, we generate a \emph{synthetic} LR image set $X_L = \{x_i^L\}_{i=1}^N$ by down-sampling each image in $X_H$, followed by resizing them back to the original image size via bilinear up-sampling (i.e., $x_i^L \in \bbR^{H \times W \times 3}$), where $x_i^L$ is the synthetic LR image of $x_i^H$. Obviously, the label set $Y_L$ for $X_L$ is identical to $Y_H$.

As shown in Figure~\ref{fig:Model}, our network comprises two components: the Cross-Resolution Generative Adversarial Network (CRGAN) and the Cross-Modal Re-ID network. To achieve cross-resolution person re-ID, our CRGAN simultaneously learns a resolution-invariant representation $f \in \bbR^{h \times w \times d}$ ($h \times w$ is the spatial size of $f$ whereas $d$ denotes the number of channels) from the input cross-resolution images, while producing the associated HR images as the decoder outputs. The recovered HR output image will be encoded as an HR representation $g \in \bbR^{h \times w \times d}$ by the HR encoder. For person re-ID, we first concatenate $f$ and $g$ to form a joint representation $\mathbfit{v} = [f, g] \in \bbR^{h \times w \times 2d}$. The classifier then takes the joint representation $\mathbfit{v}$ as input to perform person identity classification. The details of each component are elaborated in the following subsections.

As for testing, our network takes a query image resized to $H \times W \times 3$ as the input, and computes the joint representation $\mathbfit{v} = [f, g] \in \bbR^{h \times w \times 2d}$. We then apply global average pooling ($\mathrm{GAP}$) to $\mathbfit{v}$ for deriving a joint feature vector $\mathbfit{u} = \mathrm{GAP}(\mathbfit{v}) \in \bbR^{2d}$, which is applied to match the gallery images via nearest neighbor search with Euclidean distance. It is worth repeating that, the query image during testing can be with varying resolutions or with unseen ones during training (verified in experiments).

\subsection{Cross-Resolution GAN (CRGAN)}

In CRGAN, we have a cross-resolution encoder $\mathcal{E}$ which converts input images across different resolutions into resolution-invariant representations, followed by a high-resolution decoder $\mathcal{G}$ recovering the associated HR versions.

{\flushleft {\bf Cross-resolution encoder $\mathcal{E}$.}}
Since our goal is to perform cross-resolution person re-ID, we encourage the cross-resolution encoder $\mathcal{E}$ to extract resolution-invariant features for input images across resolutions (e.g., HR images in $X_H$ and LR ones in $X_L$). To achieve this, we advance adversarial learning strategies and deploy a resolution discriminator $\mathcal{D}_{F}$ in the latent \emph{feature space}. This discriminator $\mathcal{D}_{F}$ takes the feature maps $f_H$ and $f_L$ as inputs to determine whether the input feature maps are from $X_H$ or $X_L$. 

To be more precise, we define the feature-level adversarial loss $\mathcal{L}_\mathrm{adv}^{\mathcal{D}_{F}}$ as
\begin{equation}
  \begin{aligned}
  \mathcal{L}_\mathrm{adv}^{\mathcal{D}_{F}} = &~ \mathbb{E}_{x_H \sim X_H}[\log(\mathcal{D}_{F}(f_H))]\\
  + &~ \mathbb{E}_{x_L \sim X_L}[\log(1 - \mathcal{D}_{F}(f_L))],
  \end{aligned}
  \label{eq:adv_loss_feature}
\end{equation}
where $f_H = \mathcal{E}({x_H})$ and $f_L = \mathcal{E}({x_L}) \in \bbR^{h \times w \times d}$ denote the encoded HR and LR image features, respectively.\footnote{For simplicity, we omit the subscript $i$, denote HR and LR images as $x_H$ and $x_L$, and represent their corresponding labels as $y_H$ and $y_L$.}

While aligning feature distributions between HR and LR images at a single feature level has been shown effective to some extent in our previous results~\cite{CAD-Net}, similar to existing methods for semantic segmentation~\cite{tsai2018learning} and person re-ID~\cite{chen2019learning}, we adopt multi-scale adversarial networks and align feature distributions at multiple levels to learn more robust feature representations. In this work, we employ the ResNet-$50$~\cite{he2016deep} as the cross-resolution encoder $\mathcal{E}$, which has five residual blocks $\{R_1, R_2, R_3, R_4, R_5\}$. The feature maps extracted from the last activation layer of each residual block are denoted as $\{f^1, f^2, f^3, f^4, f^5\}$, where $f^j \in \bbR^{h_j \times w_j \times d_j}$ is of spatial size $h_j \times w_j$ and with $d_j$ channels.

\begin{figure}[t]
  \centering
  \includegraphics[width=0.9\linewidth]{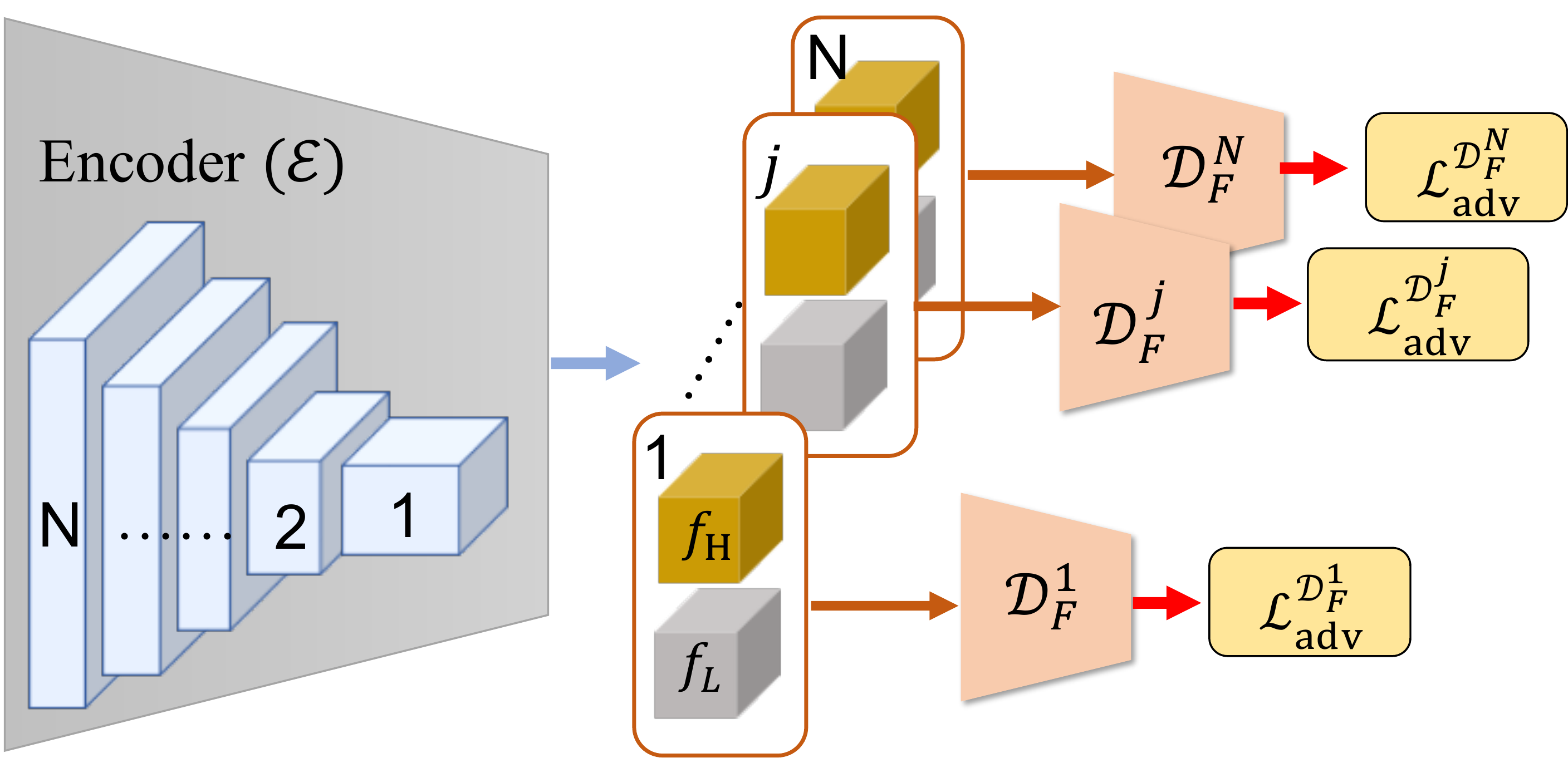}
  \caption{\textbf{Multi-scale adversarial learning.} We adopt multiple discriminators to effectively align feature distributions between HR and LR images at different feature levels. This multi-scale adversarial learning strategy allows our model to learn resolution-invariant representations that are more robust to resolution variations.}
  \label{fig:multi-discriminator}
  \vspace{-4.0mm}
\end{figure}

As shown in Figure~\ref{fig:multi-discriminator}, our multi-scale discriminator $\mathcal{D}_F^j$ takes the feature maps $f_H^j$ and $f_L^j$ extracted at the corresponding feature level as inputs, and determines whether the input feature map is from $X_H$ or $X_L$. Note that $j \in \{1, 2, 3, 4, 5\}$ is the index of the feature levels, and $f_H^j$ and $f_L^j$ denote the feature maps of $x_H$ and $x_L$, respectively.

To train the cross-resolution encoder $\mathcal{E}$ and the multi-scale resolution discriminators $\{\mathcal{D}_F^j\}$ with cross-resolution input images $x_H$ and $x_L$, we extend the adversarial loss in Eq.~\eqref{eq:adv_loss_feature} from single-scale to multi-scale adversarial learning and define the multi-scale feature-level adversarial loss $\mathcal{L}_\mathrm{adv}^{\mathcal{D}_{F}}$ as
\begin{equation}
  \begin{split}
  \mathcal{L}_\mathrm{adv}^{\mathcal{D}_{F}} = \sum_{j}^{}\bigg(&~ \mathbb{E}_{x_H \sim X_H}[\log(\mathcal{D}_F^{j}(f_H^j))] \\
  + &~ \mathbb{E}_{x_L \sim X_L}[\log(1 - \mathcal{D}_F^{j}(f_L^j))]\bigg).
  \end{split}
  \label{eq:adv_multi}
\end{equation}

With the multi-scale feature-level adversarial loss $\mathcal{L}_\mathrm{adv}^{\mathcal{D}_{F}}$, our multi-scale resolution discriminators $\{\mathcal{D}_F^j\}$ align the feature distributions across resolutions, carrying out the learning of resolution-invariant representations.

{\flushleft {\bf High-resolution decoder $\mathcal{G}$.}}
In addition to learning the resolution-invariant representation $f$, our CRGAN further synthesizes the associated HR images. This is to recover the missing details in LR inputs, together with the re-ID task to be performed later in the cross-modal re-ID network.

To achieve this goal, we have an HR decoder $\mathcal{G}$ in our CRGAN which reconstructs (or recovers) the HR images as the outputs. To accomplish this, we apply an HR reconstruction loss $\mathcal{L}_\mathrm{rec}$ between the reconstructed HR images and their corresponding HR ground-truth images. Specifically, the HR reconstruction loss $\mathcal{L}_\mathrm{rec}$ is defined as
\begin{equation}
  \label{eq:rec}
  \begin{aligned}
  \mathcal{L}_\mathrm{rec} = &~ \mathbb{E}_{x_H \sim X_H}[\|\mathcal{G}(f_H) - x_H\|_1]\\
  + &~ \mathbb{E}_{x_L \sim X_L}[\|\mathcal{G}(f_L) - x_{H}\|_1],
  \end{aligned}
\end{equation}
where the HR ground-truth image associated with $x_L$ is $x_H$. Following Huang~\etal~\cite{huang2018munit}, we adopt the $\ell_1$ norm in the HR reconstruction loss $\mathcal{L}_\mathrm{rec}$ as it preserves image sharpness. We note that both $X_H$ and $X_L$ will be shuffled during training. That is, images of the same identity but different resolutions will not necessarily be observed by the CRGAN at the same time.

It is worth noting that, while the aforementioned HR reconstruction loss $\mathcal{L}_\mathrm{rec}$ could reduce information loss in the latent feature space, we follow Ledig~\etal~\cite{ledig2017photo} and introduce skip connections between the cross-resolution encoder $\mathcal{E}$ and the HR decoder $\mathcal{G}$. This would facilitate the learning process of image reconstruction, as well as allowing more efficient gradient propagation.

To make the HR decoder $\mathcal{G}$ produce more perceptually realistic HR outputs and associate with the task of person re-ID, we further adopt adversarial learning in the \emph{image space} and introduce an HR image discriminator $\mathcal{D}_{I}$ which takes the recovered HR images (i.e., $\mathcal{G}(f_L)$ and $\mathcal{G}(f_H)$) and their corresponding HR ground-truth images as inputs to distinguish whether the input images are real or fake~\cite{ledig2017photo,wang2018cascaded}. Specifically, we define the image-level adversarial loss $\mathcal{L}_\mathrm{adv}^{\mathcal{D}_{I}}$ as
\begin{equation}\scriptsize
  \begin{aligned}
  \mathcal{L}_\mathrm{adv}^{\mathcal{D}_{I}} = &~ \mathbb{E}_{x_H \sim X_H}[\log(\mathcal{D}_{I}(x_H))] + \mathbb{E}_{x_L \sim X_L}[\log(1 - \mathcal{D}_{I}(\mathcal{G}(f_L)))] \\
  + &~ \mathbb{E}_{x_H \sim X_H}[\log(\mathcal{D}_{I}(x_H))] + \mathbb{E}_{x_H \sim X_H}[\log(1 - \mathcal{D}_{I}(\mathcal{G}(f_H)))].
  \end{aligned}
  \label{eq:adv_loss_image}
\end{equation}

It is also worth repeating that the goal of this HR decoder $\mathcal{G}$ is not simply to recover the missing details in LR inputs, but also to have such recovered HR images aligned with the learning task of interest (i.e., person re-ID). Namely, we encourage the HR decoder $\mathcal{G}$ to perform \textit{re-ID oriented} HR recovery, which is further realized by the following cross-modal re-ID network.

\begin{figure}[t]
  \centering
  \includegraphics[width=0.9\linewidth]{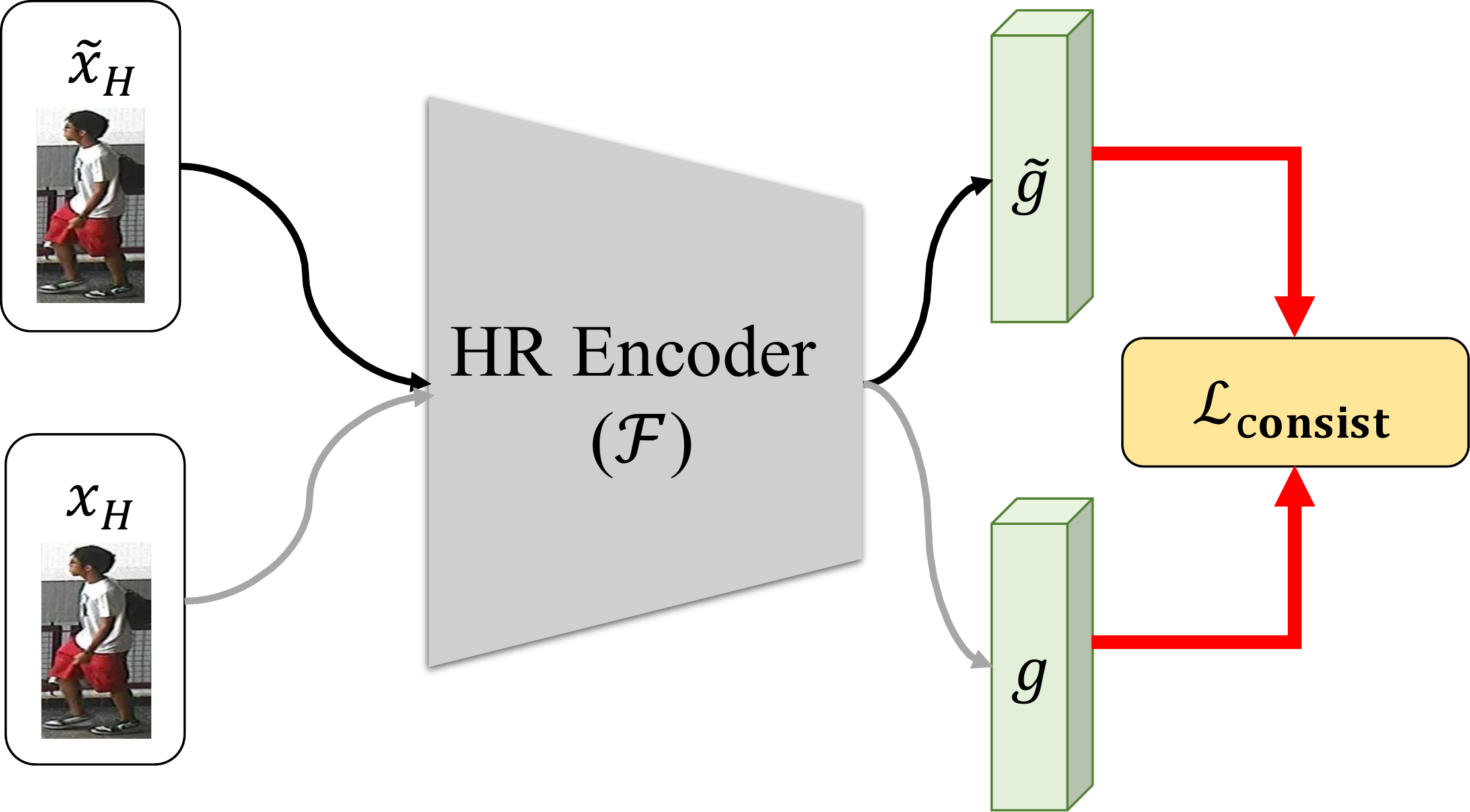}
  \caption{\textbf{Illustration of the feature consistency loss $\mathcal{L}_\mathrm{consist}$.} The HR encoder $\mathcal{F}$ takes the recovered HR image $\tilde{x}_H$ and the corresponding HR ground-truth image $x_H$ as inputs and derives their HR representations $\tilde{g}$ and $g$, respectively. We then introduce the feature consistency loss $\mathcal{L}_\mathrm{consist}$ to enforce the consistency between $\tilde{g}$ and $g$. This consistency loss allows our HR encoder $\mathcal{F}$ to learn HR representations that are more robust to the variations of HR image recovery.}
  \label{fig:consist-loss}
  \vspace{-3.0mm}
\end{figure}

\subsection{Cross-Modal Re-ID}

As shown in Figure~\ref{fig:Model}, the cross-modal re-ID network first applies an HR encoder $\mathcal{F}$, which takes the reconstructed HR image from the CRGAN as input, to derive the HR feature representation $g \in \bbR^{h \times w \times d}$. Then, a classifier $\mathcal{C}$ is learned to complete person re-ID.

While enforcing the HR reconstruction loss $\mathcal{L}_\mathrm{rec}$ and the image-level adversarial loss $\mathcal{L}_\mathrm{adv}^{\mathcal{D}_I}$ allows our model to map LR input images of various resolutions into their HR versions to some extent~\cite{CAD-Net}, it is still difficult for the model to simultaneously handle the resolution variations and learn the mapping between LR and HR images, especially when there are infinitely many mappings between LR and HR images. To address the variations of HR image recovery, we introduce a feature consistency loss $\mathcal{L}_\mathrm{consist}$ that enforces the consistency between the features of the recovered HR images and the corresponding HR ground-truth images. As illustrated in Figure~\ref{fig:consist-loss}, the HR encoder $\mathcal{F}$ takes the recovered HR image $\tilde{x}_H$ and its corresponding HR ground-truth image $x_H$ as inputs and derives the HR representations $\tilde{g} = \mathcal{F}(\tilde{x}_H)$ and $g = \mathcal{F}(x_H)$, respectively. We then enforce the consistency between $\tilde{g}$ and $g$ using the $\ell_1$ distance and define the feature consistency loss $\mathcal{L}_\mathrm{consist}$ as
\begin{equation}
  \label{eq:consist}
  \begin{aligned}
  \mathcal{L}_\mathrm{consist} = &~ \mathbb{E}_{x_H \sim X_H}[\|\mathcal{F}(\tilde{x}_H) - \mathcal{F}(x_H)\|_1]\\
  + &~ \mathbb{E}_{x_L \sim X_L}[\|\mathcal{F}(\tilde{x}_H) - \mathcal{F}(x_{H})\|_1].
  \end{aligned}
\end{equation}

Enforcing the feature consistency loss $\mathcal{L}_\mathrm{consist}$ allows the HR decoder $\mathcal{G}$ to derive the HR representation $g$ which is more robust to the variations of HR image recovery.

As for the input to the classifier $\mathcal{C}$, we jointly consider the feature representations of two different modalities for person identity classification, i.e., the resolution-invariant representation $f$ and the HR representation $g$. The former preserves content information, while the latter observes the recovered HR details for person re-ID. Thus, we have the classifier $\mathcal{C}$ take the concatenated feature representation $\mathbfit{v} = [f, g] \in \bbR^{h \times w \times 2d}$ as the input. In this work, the adopted classification loss $\mathcal{L}_\mathrm{cls}$ is the integration of the identity loss $\mathcal{L}_\mathrm{id}$ and the triplet loss $\mathcal{L}_\mathrm{tri}$~\cite{hermans2017defense}, and is defined as
\begin{equation}
  \begin{aligned}
  \mathcal{L}_\mathrm{cls} = \mathcal{L}_\mathrm{id} + \mathcal{L}_\mathrm{tri},
  \end{aligned}
  \label{eq:cls}
\end{equation}
where the identity loss $\mathcal{L}_\mathrm{id}$ computes the softmax cross entropy between the classification prediction and the corresponding ground-truth one hot vector, while the triplet loss $\mathcal{L}_\mathrm{tri}$ is introduced to enhance the discrimination ability during the re-ID process and is defined as
\begin{equation}
  \begin{aligned}
  \mathcal{L}_\mathrm{tri}
   = &~ \mathbb{E}_{(x_H,y_H) \sim (X_H,Y_H)}\max(0, \phi + d_\mathrm{pos}^H - d_\mathrm{neg}^H) \\
  + &~ \mathbb{E}_{(x_L,y_L) \sim (X_L,Y_L)}\max(0, \phi + d_\mathrm{pos}^L - d_\mathrm{neg}^L),
  \end{aligned}
  \label{eq:tri}
\end{equation}
where $d_\mathrm{pos}$ and $d_\mathrm{neg}$ are the distances between the positive (same label) and the negative (different labels) image pairs, respectively, and $\phi > 0$ serves as the margin.

It can be seen that, the above cross-resolution person re-ID framework is very different from existing one like CSR-GAN~\cite{wang2018cascaded}, which addresses image SR and person re-ID \emph{separately}. More importantly, the aforementioned identity loss $\mathcal{L}_\mathrm{id}$ not only updates the classifier $\mathcal{C}$, but also refines the HR decoder $\mathcal{G}$ in our CRGAN. This is the reason why our CRGAN is able to produce \emph{re-ID oriented} HR outputs, i.e., the recovered HR details preferable for person re-ID.

\begin{algorithm}[t]
  \caption{\revision{Training procedure of CAD-Net++}}
  \label{alg:cadnet}
  \small
  \KwData{Image sets $X_H$ and $X_L$ and label sets $Y_H$ and $Y_L$}
  \KwResult{Network parameters of CAD-Net++}
  $\mathcal{E}$ and $\mathcal{F}$ are initialized from weights pre-trained on ImageNet\\
  $\mathcal{G}$, $\{\mathcal{D}_F^j\}_{j=1}^2$, $\mathcal{D}_I$, and $\mathcal{C}$ are randomly initialized\\
  \For{each training iteration}{
    Construct $x_{H}$ and $y_{H}$ by sampling from $X_{H}$ and $Y_{H}$, respectively \\
    Construct $x_{L}$ and $y_{L}$ by sampling from $X_{L}$ and $Y_{L}$, respectively \\
    Extract features: $\{f_H^j\}_{j=1}^5 \leftarrow \mathcal{E}(x_{H})$ and $\{f_L^j\}_{j=1}^5 \leftarrow \mathcal{E}(x_{L})$ \\ 
    Reconstruct HR images: $\tilde{x}_{H}^H \leftarrow \mathcal{G}(f_{H}^1)$ and $\tilde{x}_{H}^L \leftarrow \mathcal{G}(f_{L}^1)$ \\
    Compute losses $\mathcal{L}_\mathrm{adv}^{\mathcal{D}_F}$ via Eq. (\ref{eq:adv_multi}), $\mathcal{L}_\mathrm{rec}$ via Eq. (\ref{eq:rec}), and $\mathcal{L}_\mathrm{adv}^{\mathcal{D}_I}$ via Eq. (\ref{eq:adv_loss_image}) \\
    \For{each iteration of updating CRGAN}{
      $\theta_\mathcal{E} \leftarrow \theta_\mathcal{E} + \nabla_{\theta_{\mathcal{E}}} (\lambda_\mathrm{adv}^{\mathcal{D}_{F}} \cdot \mathcal{L}_\mathrm{adv}^{\mathcal{D}_F} + \lambda_\mathrm{adv}^{\mathcal{D}_{I}} \cdot \mathcal{L}_\mathrm{adv}^{\mathcal{D}_I} - \lambda_\mathrm{rec} \cdot \mathcal{L}_\mathrm{rec}$) \\
      $\theta_\mathcal{G} \leftarrow \theta_\mathcal{G} + \nabla_{\theta_\mathcal{G}} (\lambda_\mathrm{adv}^{\mathcal{D}_{I}} \cdot \mathcal{L}_\mathrm{adv}^{\mathcal{D}_I} - \lambda_\mathrm{rec} \cdot \mathcal{L}_\mathrm{rec}$)
  	}
    \For{each iteration of updating the discriminators}{
      $\theta_{\mathcal{D}_F} \leftarrow \theta_{\mathcal{D}_F} - \nabla_{\theta_{\mathcal{D}_F}} \lambda_\mathrm{adv}^{\mathcal{D}_{F}} \cdot \mathcal{L}_\mathrm{adv}^{\mathcal{D}_F}$\\
      $\theta_{\mathcal{D}_I} \leftarrow \theta_{\mathcal{D}_I} - {\nabla}_{\theta_{\mathcal{D}_I}} \lambda_\mathrm{adv}^{\mathcal{D}_{I}} \cdot \mathcal{L}_\mathrm{adv}^{\mathcal{D}_I}$\\
  	}
  	Encode HR features: $\tilde{g} \leftarrow \mathcal{F}(\tilde{x}_H)$ and $g \leftarrow \mathcal{F}(x_H)$ \\
  	Compute loss $\mathcal{L}_\mathrm{consist}$ via Eq. (\ref{eq:consist}) \\
  	$\theta_{\mathcal{F}} \leftarrow \theta_{\mathcal{F}} - \nabla_{\theta_{\mathcal{F}}} \lambda_\mathrm{consist} \cdot \mathcal{L}_\mathrm{consist}$\\
  	Compute loss $\mathcal{L}_\mathrm{cls}$ via Eq. (\ref{eq:cls}) \\
  	$\theta_\mathcal{E} \leftarrow \theta_\mathcal{E} - \nabla_{\theta_{\mathcal{E}}} \lambda_\mathrm{cls} \cdot \mathcal{L}_\mathrm{cls}$\\
    $\theta_\mathcal{G} \leftarrow \theta_\mathcal{G} - \nabla_{\theta_{\mathcal{G}}} \lambda_\mathrm{cls} \cdot \mathcal{L}_\mathrm{cls}$\\
    $\theta_\mathcal{F} \leftarrow \theta_\mathcal{F} - \nabla_{\theta_{\mathcal{F}}} \lambda_\mathrm{cls} \cdot \mathcal{L}_\mathrm{cls}$\\
    }
%   \normalsize
\end{algorithm}

{\flushleft {\bf Full training objective.}}
The total loss $\mathcal{L}$ for training our proposed CAD-Net++ is summarized as follows:
\begin{equation}
  \begin{split}
  \mathcal{L} & = \mathcal{L}_\mathrm{cls} + \lambda_\mathrm{adv}^{\mathcal{D}_{F}}\cdot\mathcal{L}_\mathrm{adv}^{\mathcal{D}_{F}} + \lambda_\mathrm{rec}\cdot\mathcal{L}_\mathrm{rec} \\ & + \lambda_\mathrm{adv}^{\mathcal{D}_{I}}\cdot\mathcal{L}_\mathrm{adv}^{\mathcal{D}_{I}} + \lambda_\mathrm{consist}\cdot\mathcal{L}_\mathrm{consist},
  \end{split}
  \label{eq:fullobj}
\end{equation}
where $\lambda_\mathrm{adv}^{\mathcal{D}_{F}}$, $\lambda_\mathrm{rec}$, $\lambda_\mathrm{adv}^{\mathcal{D}_{I}}$, and $\lambda_\mathrm{consist}$ are the hyper-parameters used to control the relative importance of the corresponding losses. We note that $\mathcal{L}_\mathrm{adv}^{\mathcal{D}_{F}}$, $\mathcal{L}_\mathrm{rec}$, and $\mathcal{L}_\mathrm{adv}^{\mathcal{D}_{I}}$ are developed to learn the CRGAN, $\mathcal{L}_\mathrm{consist}$ is introduced to update the cross-modal re-ID component, and $\mathcal{L}_\mathrm{cls}$ is designed to update the entire framework.

To learn our network with the HR training images and their down-sampled LR ones, we minimize the HR reconstruction loss $\mathcal{L}_\mathrm{rec}$ for updating our CRGAN, the feature consistency loss $\mathcal{L}_\mathrm{consist}$ for updating the HR encoder $\mathcal{F}$, and the classification loss $\mathcal{L}_\mathrm{cls}$ for jointly updating the CRGAN and the cross-modal re-ID network. The image-level adversarial loss $\mathcal{L}_\mathrm{adv}^{\mathcal{D}_I}$ is computed for producing perceptually realistic HR images, while the multi-scale feature-level adversarial loss $\mathcal{L}_\mathrm{adv}^{\mathcal{D}_F}$ is optimized for learning resolution-invariant representations.

\revision{The training details of CAD-Net++ are summarized in Algorithm~\ref{alg:cadnet}. Specifically, we train our CAD-Net++ until all losses converge.}

\section{Experiments}

In this section, we first describe the implementation details, the adopted datasets for evaluation, and the experimental settings. We then present both quantitative and qualitative results, including ablation studies.

\subsection{Implementation Details}

We implement our model using PyTorch. The ResNet-$50$~\cite{he2016deep} pretrained on ImageNet is used to build the cross-resolution encoder $\mathcal{E}$ and the HR encoder $\mathcal{F}$. Note that since $\mathcal{E}$ and $\mathcal{F}$ work for different tasks, these two components do not share weights. The classifier $\mathcal{C}$ is composed of a global average pooling layer and a fully connected layer followed a softmax activation. The architecture of the resolution discriminator $\mathcal{D}_F$ is the same as that adopted by Tsai~\etal~\cite{tsai2018learning}. The structure of the HR image discriminator $\mathcal{D}_I$ is similar to the ResNet-$18$~\cite{he2016deep}. Our HR decoder $\mathcal{G}$ is similar to that proposed by Miyato~\etal~\cite{miyato2018cgans}. Components $\mathcal{D}_F$, $\mathcal{D}_I$, $\mathcal{G}$, and $\mathcal{C}$ are all randomly initialized. We use stochastic gradient descent to train the proposed model. For components $\mathcal{E}$, $\mathcal{G}$, $\mathcal{F}$, and $\mathcal{C}$, the learning rate, momentum, and weight decay are $1 \times 10^{-3}$, $0.9$, and $5 \times 10^{-4}$, respectively. For the two discriminators $\mathcal{D}_F$ and $\mathcal{D}_I$, the learning rate is set to $1 \times 10^{-4}$. The batch size is $32$. The margin $\phi$ in the triplet loss $\mathcal{L}_\mathrm{tri}$ is set to $2$. We set the hyper-parameters in all the experiments as follows: $\lambda_\mathrm{adv}^{\mathcal{D}_{F}}$ = 1, $\lambda_\mathrm{rec}$ = 1, $\lambda_\mathrm{adv}^{\mathcal{D}_{I}}$ = 1, and $\lambda_\mathrm{consist}$ = 1. All images of various resolutions are resized to $256 \times 128 \times 3$ in advance. We train our model on a single NVIDIA GeForce GTX $1080$ GPU with $12$ GB memory.

\subsection{Datasets}

We adopt five person re-ID datasets, including CUHK03~\cite{li2014deepreid}, VIPeR~\cite{gray2008viewpoint}, CAVIAR~\cite{Cheng:BMVC11}, Market-$1501$~\cite{zheng2015scalable}, and DukeMTMC-reID~\cite{zheng2017unlabeled}, and two vehicle re-ID datasets, including VeRi-$776$~\cite{liu2016deep} and VRIC~\cite{kanaci2018vehicle} for evaluation. The details of each dataset are described as follows.

{\flushleft {\bf CUHK03~\cite{li2014deepreid}.}}
The CUHK03 dataset is composed of $14,097$ images of $1,467$ identities with $5$ different camera views. Following CSR-GAN~\cite{wang2018cascaded}, we use the $1,367/100$ training/test identity split.

{\flushleft {\bf VIPeR~\cite{gray2008viewpoint}.}}
The VIPeR dataset contains $632$ person-image pairs captured by $2$ cameras. Following SING~\cite{jiao2018deep}, we randomly divide this dataset into two non-overlapping halves based on the identity labels. That is, images of a subject belong to either the training set or the test set.

{\flushleft {\bf CAVIAR~\cite{Cheng:BMVC11}.}}
The CAVIAR dataset is composed of $1,220$ images of $72$ person identities captured by $2$ cameras. Following SING~\cite{jiao2018deep}, we discard $22$ people who only appear in the closer camera, and split this dataset into two non-overlapping halves according to the identity labels.

{\flushleft {\bf Market-$1501$~\cite{zheng2015scalable}.}}
The Market-$1501$ dataset consists of $32,668$ images of $1,501$ identities with $6$ camera views. We use the widely adopted $751/750$ training/test identity split.

{\flushleft {\bf DukeMTMC-reID~\cite{zheng2017unlabeled}.}}
The DukeMTMC-reID dataset contains $36,411$ images of $1,404$ identities captured by $8$ cameras. We utilize the benchmarking $702/702$ training/test identity split.

{\flushleft {\bf VeRi-$776$~\cite{liu2016deep}.}}
The VeRi-$776$ dataset is divided into two subsets: a training set and a test set. The training set is composed of $37,781$ images of $576$ vehicles, and the test set has $11,579$ images of $200$ vehicles. Following the evaluation protocol in \cite{liu2016deep}, the image-to-track cross-camera search is performed, where we treat one image of a vehicle from one camera as the query, and search for tracks of the same vehicle in other cameras.

{\flushleft {\bf VRIC~\cite{kanaci2018vehicle}.}}
The VRIC dataset is a newly collected dataset, which consists of $60,430$ images of $5,656$ vehicle IDs collected from $60$ different cameras in traffic scenes. VRIC differs significantly from existing datasets in that vehicles were captured with variations in image resolution, motion blur, weather condition, and occlusion. The training set has $54,808$ images of $2,811$ vehicles, while the rest $5,622$ images of $2,811$ identities are used for testing.

\subsection{Experimental Settings and Evaluation Metrics}

We evaluate the proposed method under three different settings: (1) \emph{cross-resolution setting}~\cite{jiao2018deep,kanaci2018vehicle}, (2) \emph{standard setting}~\cite{ge2018fd,zhong2017camera}, and (3) \emph{semi-supervised setting}~\cite{chen2019learning}. For cross-resolution setting, the test (query) set is composed of LR images while the gallery set contains HR images only. For standard setting (i.e., re-ID with no significant resolution variations), both query and gallery sets contain HR images. For semi-supervised setting (i.e., re-ID with partially labeled datasets), we follow cross-resolution setting where the test (query) set consists LR images while the gallery set comprises HR images only.

In all of the experiments, we adopt the standard single-shot re-ID setting~\cite{jiao2018deep,liao2015person}. We note that the cross-resolution re-ID setting analyzes the \emph{robustness against the resolution variations}, while the standard re-ID setting examines if our method still improves re-ID if no significant resolution variations are present. The semi-supervised re-ID setting aims to investigate whether our proposed algorithm still exhibits sufficient ability in re-identifying images with less supervision.

We adopt the multi-scale resolution discriminators $\{\mathcal{D}_F^j\}$ which align feature distributions at different feature levels. To balance between learning efficiency and performance, we select the index of feature level with $j \in \{1, 2\}$, and denote our method as ``Ours (multi-scale)'' and the variant of our method with single-scale resolution discriminator ($j = 1$) as ``Ours (single-scale)''.

For performance evaluation, we adopt the average cumulative match characteristic as the evaluation metric. We note that the performance of our method can be further improved by applying pre-/post-processing methods, attention mechanisms, or re-ranking. For fair comparisons, no such techniques are used in all of our experiments.

\begin{table*}[t]
  %\small
  \ra{1.3}
  \begin{center}
  \caption{\textbf{Experimental results of cross-resolution person re-ID (\%).} The bold and underlined numbers indicate top two results, respectively.}
  \vspace{-2.0mm}
  \label{table:exp-ReID}
  \resizebox{\linewidth}{!} 
  {
  \begin{tabular}{l|ccc|ccc|ccc|ccc|ccc}
  \toprule
  \multirow{2}{*}{Method} & \multicolumn{3}{c|}{MLR-CUHK03} & \multicolumn{3}{c|}{MLR-VIPeR} & \multicolumn{3}{c|}{CAVIAR} & \multicolumn{3}{c|}{MLR-Market-1501} & \multicolumn{3}{c}{MLR-DukeMTMC-reID}\\
  & Rank 1 & Rank 5 & Rank 10 & Rank 1 & Rank 5 & Rank 10 & Rank 1 & Rank 5 & Rank 10 & Rank 1 & Rank 5 & Rank 10 & Rank 1 & Rank 5 & Rank 10 \\
  \midrule
  PCB~\cite{sun2018beyond} & 75.3 & 92.7 & 98.1 & 42.6 & 65.8 & 75.9 & 35.9 & 72.1 & 88.6 & 76.9 & 88.9 & 92.4 & 66.4 & 82.5 & 87.1 \\
  SPreID~\cite{kalayeh2018human} & 76.5 & 92.5 & 98.3 & 42.4 & 65.8 & 75.1 & 36.2 & 71.9 & 88.7 & 77.4 & 89.0 & 93.9 & 68.4 & 84.5 & 89.1 \\
  Part Aligned~\cite{suh2018part} & 73.4 & 92.1 & 97.5 & 40.2 & 62.3 & 73.1 & 35.7 & 71.4 & 87.9 & 75.6 & 88.5 & 92.2 & 67.5 & 83.1 & 87.2 \\
  CamStyle~\cite{zhong2017camera} & 69.1 & 89.6 & 93.9 & 34.4 & 56.8 & 66.6 & 32.1 & 72.3 & 85.9  & 74.5 & 88.6 & 93.0 & 64.0 & 78.1 & 84.4 \\
  FD-GAN~\cite{ge2018fd} &73.4 & 93.8 & 97.9 & 39.1 & 62.1 & 72.5 & 33.5 & 71.4 & 86.5  & 79.6 & 91.6 & 93.5 & 67.5 & 82.0 & 85.3 \\
  \midrule
  JUDEA~\cite{li2015multi} & 26.2 & 58.0 & 73.4 & 26.0 & 55.1 & 69.2 & 22.0 & 60.1 & 80.8 & - & - & - & - & - & - \\
  SLD$^2$L~\cite{jing2015super} & - & - & -& 20.3 & 44.0 & 62.0 & 18.4 & 44.8 & 61.2  & - & - & - & - & - & - \\
  SDF~\cite{wang2016scale} & 22.2 & 48.0 & 64.0 & 9.25 & 38.1 & 52.4 & 14.3 & 37.5 & 62.5 & - & - & - & - & - & - \\
  %
  %RAIN (single-level)~\cite{chen2019learning} & 77.6 & 96.2 & 98.5 & 41.2 & 66.3 & 75.6 & 41.5 & 75.3 & 85.6 & - & - & - & - & - & - \\
  %
  DenseNet-121~\cite{huang2017densely} & 70.8 & 91.3 & - & 31.4 & 63.1 & - & 31.1 & 65.5 & - & 60.0 & 78.8 & - & - & - & - \\
  SE-ResNet-50~\cite{hu2018squeeze} & 70.8 & 92.3 & - & 33.5 & 63.6 & - & 30.8 & 65.1 & - & 58.2 & 78.6 & - & - & - & - \\
  ResNet-50~\cite{he2016deep} & 67.4 & 91.7 & - & 29.9 & 62.2 & - & 29.6 & 64.0 & - & 57.0 & 78.7 & - & - & - & - \\
  FFSR~\cite{mao2019resolution} & 70.5 & 92.3 & - & 40.3 & 65.3 & - & 31.1 & 68.7 & - & 59.2 & 80.1 & - & - & - & - \\
  RIFE~\cite{mao2019resolution} & 69.7 & 91.5 & - & 33.9 & 63.6 & - & 35.7 & 74.9 & - & 62.6 & 82.4 & - & - & - & - \\
  FFSR+RIFE~\cite{mao2019resolution} & 73.3 & 92.6 & - & 41.6 & 64.9 & - & 36.4 & 72.0 & - & 66.9 & 84.7 & - & - & - & - \\
  RAIN~\cite{chen2019learning} & 78.9 & 97.3 & 98.7 & 42.5 & 68.3 & \textbf{79.6} & 42.0 & \textbf{77.3} & 89.6 & - & - & - & - & - & - \\
  \revision{FSRCNN-reID~\cite{dong2016accelerating}} & 68.1 & 90.5 & 95.3 & 34.0 & 57.9 & 67.2 & 33.9 & 72.4 & 89.2 & 75.0 & 88.3 & 92.1 & 66.3 & 80.7 & 85.2 \\
  SING~\cite{jiao2018deep} & 67.7 & 90.7 & 94.7 & 33.5 & 57.0 & 66.5 & 33.5 & 72.7 & 89.0  & 74.4 & 87.8 & 91.6 & 65.2 & 80.1 & 84.8 \\
  CSR-GAN~\cite{wang2018cascaded} & 71.3 & 92.1 & 97.4 & 37.2 & 62.3 & 71.6 & 34.7 & 72.5 & 87.4 & 76.4 & 88.5 & 91.9 & 67.6 & 81.4 & 85.1 \\
  \midrule %
  %
%   CAD-Net ($f$ only)~\cite{CAD-Net} & 77.6 & 96.2 & 98.5& 41.2 & 66.3 & 75.6 & 41.5 & 75.3 & 85.6 & 80.1 & 90.6 & 93.2 & 73.4 & 84.4 & 86.8 \\
%   %
%   CAD-Net ($g$ only)~\cite{CAD-Net} & 79.7 & 97.4 & 98.7 & 41.7 & 66.4 & 76.1 & 38.9 & 73.1 & 90.6 & 82.2 & 91.3 & 94.5 & 74.1 & 85.1 & 88.2 \\
%   %
  CAD-Net~\cite{CAD-Net} & 82.1 & 97.4 & 98.8 & 43.1 & 68.2 & 77.5 & 42.8 & 76.2 & 91.5 & 83.7 & 92.7 & 95.8 & 75.6 & 86.7 & 89.6 \\
  Ours (single-scale) & \underline{82.9} & \underline{97.8} & 98.9 & \underline{43.2} & \underline{68.4} & 77.6 & \underline{42.9} & 76.3 & \underline{91.8} & \underline{83.9} & \underline{92.8} & \underline{96.0} & \underline{76.1} & \underline{87.2} & \underline{89.9} \\
  Ours (multi-scale) ($f$ only) & 77.8 & 96.5 & 98.7 & 41.3 & 66.4 & 75.6 & 41.5 & 75.6 & 86.0 & 80.4 & 90.8 & 93.2 & 73.7 & 84.4 & 86.9 \\
  Ours (multi-scale) ($g$ only) & 81.9 & 97.1 & \underline{99.0} & 43.0 & 68.1 & 77.4 & 42.7 & 76.1 & \underline{91.8} & 83.8 & 92.6 & \underline{96.0} & 75.4 & 86.5 & 89.3 \\
  Ours (multi-scale) & \textbf{83.4} & \textbf{98.1} & \textbf{99.1} & \textbf{43.4} & \textbf{68.7} & \underline{78.2} & \textbf{43.1} & \underline{76.5} & \textbf{92.3} & \textbf{84.1} & \textbf{93.0} & \textbf{96.2} & \textbf{77.2} & \textbf{88.1} & \textbf{90.4} \\
  \bottomrule
  \end{tabular}
  }
  \end{center}
  \vspace{-3.0mm}
\end{table*}

\subsection{Evaluation of Cross-Resolution Setting}

We evaluate our proposed algorithm on both person re-ID~\cite{jiao2018deep,CAD-Net} and vehicle re-ID~\cite{kanaci2018vehicle} tasks.

\subsubsection{Cross-Resolution Person Re-ID}

Following SING~\cite{jiao2018deep}, we consider multiple low-resolution (MLR) person re-ID and evaluate the proposed method on \emph{four synthetic} and \emph{one real-world} benchmarks. To construct the synthetic MLR datasets (i.e., MLR-CUHK03, MLR-VIPeR, MLR-Market-$1501$, and MLR-DukeMTMC-reID), we follow SING~\cite{jiao2018deep} and down-sample images taken by one camera by a randomly selected down-sampling rate $r \in \{2, 3, 4\}$ (i.e., the size of the down-sampled image becomes $\frac{H}{r} \times \frac{W}{r} \times 3$), while the images taken by the other camera(s) remain unchanged. The CAVIAR dataset inherently contains realistic images of multiple resolutions, and is a \emph{genuine} and more challenging dataset for evaluating MLR person re-ID.

\revision{We compare our proposed approach (CAD-Net++) with methods developed for cross-resolution person re-ID, including JUDEA~\cite{li2015multi}, SLD$^2$L~\cite{jing2015super}, SDF~\cite{wang2016scale}, RAIN~\cite{chen2019learning}, DenseNet-121~\cite{huang2017densely}, SE-ResNet-50~\cite{hu2018squeeze}, ResNet-50~\cite{he2016deep}, FFSR~\cite{mao2019resolution}, RIFE~\cite{mao2019resolution}, FSRCNN-reID~\cite{dong2016accelerating} (FSRCNN followed by the same representation learning method as SING~\cite{jiao2018deep}), SING~\cite{jiao2018deep}, and CSR-GAN~\cite{wang2018cascaded}, and approaches developed for standard person re-ID, including PCB~\cite{sun2018beyond}, SPreID~\cite{kalayeh2018human}, Part Aligned~\cite{suh2018part}, CamStyle~\cite{zhong2017camera}, and FD-GAN~\cite{ge2018fd}.} For methods developed for cross-resolution person re-ID, the training set contains HR images and LR ones with all three down-sampling rates $r \in \{2, 3, 4\}$ for each person. For methods developed for standard person re-ID, the training set contains HR images for each identity only.

\revision{
{\flushleft {\bf Results.}}
Table~\ref{table:exp-ReID} reports the quantitative results recorded at ranks $1$, $5$, and $10$ on all five adopted datasets. For CSR-GAN~\cite{wang2018cascaded} on the MLR-CUHK03, CAVIAR, MLR-Market-$1501$, and MLR-DukeMTMC-reID datasets, and PCB~\cite{sun2018beyond}, SPreID~\cite{kalayeh2018human}, Part Aligned~\cite{suh2018part}, CamStyle~\cite{zhong2017camera}, FSRCNN-reID~\cite{dong2016accelerating}, and FD-GAN~\cite{ge2018fd} on all five datasets, their results are obtained by running the official code with the default implementation setup.} For SING~\cite{jiao2018deep}, we reproduce their results on the MLR-Market-$1501$ and MLR-DukeMTMC-reID datasets. 

Our method adopting either single-scale or multi-scale resolution discriminators performs favorably against all competing methods on all five adopted datasets. The performance gains can be ascribed to three main factors. First, unlike most existing person re-ID methods, our model performs cross-resolution person re-ID in an end-to-end learning fashion. Second, our method learns the resolution-invariant representations, allowing our model to recognize persons in images of different resolutions. Third, our model learns to recover the missing details in LR images, thus providing additional discriminative evidence for person re-ID.

\begin{table}[t]
  \ra{1.3}
  \begin{center}
  \caption{\textbf{Experimental results of cross-resolution vehicle re-ID (\%).} The bold and underlined numbers indicate top two results, respectively.}
  \vspace{-2.0mm}
  \label{table:exp-vehicle-reid-mlr}
  \resizebox{\linewidth}{!} 
  {
  \begin{tabular}{l|ccc|ccc}
  \toprule
  \multirow{2}{*}{Method} & \multicolumn{3}{c|}{MLR-VeRi776} & \multicolumn{3}{c}{VRIC} \\
  & Rank 1 & Rank 5 & mAP & Rank 1 & Rank 5 & mAP\\
  \midrule
  Siamese-Visual~\cite{shen2017learning} & 30.4 & 48.5 & 15.2  & 30.6 & 57.3 & 42.7 \\
  OIFE~\cite{wang2017orientation} & 57.4 & 76.6 & 40.2  & 24.6 & 51.0 & 38.5 \\  
  VAMI~\cite{zhou2018aware} & 59.1 & 78.7 & 43.9  & 30.5 & 59.2 & 43.8 \\  
  \midrule
  \revision{FSRCNN-reID~\cite{dong2016accelerating}} & 56.3 & 78.0 & 45.8 & 32.1 & 63.2 & 48.6 \\
  SING~\cite{jiao2018deep} & 55.2 & 77.3 & 45.1 & 30.8 & 60.4 & 46.6 \\  
  CSR-GAN~\cite{wang2018cascaded} & 58.4 & 80.1 & \underline{48.5} & 35.1 & 65.0 & \underline{48.6} \\  
  MSVF~\cite{kanaci2018vehicle} & \underline{64.1} & \underline{82.6} & 44.2 & \underline{46.6} & \underline{65.6} & 47.5 \\ 
  \midrule
  Ours & \bf{68.7} & \bf{85.3} & \bf{53.6} & \bf{50.1} & \bf{68.2} & \bf{50.4}\\
  \bottomrule
  \end{tabular}
  }
  \end{center}
  \vspace{-5.0mm}
\end{table}

{\flushleft {\bf Effect of multi-scale adversarial learning.}}
The effect of adopting multi-scale adversarial learning strategy can be observed by comparing two of the variant methods, i.e., Ours (single-scale) and Ours (multi-scale). We observe that adopting multi-scale adversarial learning strategy consistently improves the performance over adopting single-scale adversarial learning strategy on all five datasets. 

{\flushleft {\bf Effect of deriving joint representation.}}
The advantage of deriving joint representation $\mathbfit{v} = [f,g]$ can be assessed by comparing with two of our variant methods, i.e., Ours (multi-scale) ($f$ only) and Ours (multi-scale) ($g$ only). In the Ours (multi-scale) ($f$ only) method, the classifier $\mathcal{C}$ only takes the resolution-invariant representation $f$ as input. In the Ours (multi-scale) ($g$ only) method, the classifier $\mathcal{C}$ only takes the HR representation $g$ as input. We observe that deriving joint representation $\mathbfit{v}$ consistently improves the performance over these two variant/baseline methods. 

\begin{table}[t]
  \small
  \ra{1.3}
  \begin{center}
  \caption{\textbf{Experimental results of standard person re-ID (\%).} The bold and underlined numbers indicate top two results, respectively.}
  \vspace{-2.0mm}
  \label{table:exp-standard-reid}
  \resizebox{\linewidth}{!} 
  {
  \begin{tabular}{l|cc|cc}
  \toprule
  \multirow{2}{*}{Method} & \multicolumn{2}{c|}{Market-1501} & \multicolumn{2}{c}{DukeMTMC-reID} \\
  & Rank 1 & mAP & Rank 1 & mAP\\
  \midrule
  JLML~\cite{Li-2017-IJCAI} & 85.1 & 65.5 & - & - \\
  TriNet~\cite{Hermans-2017-arXiv} & 84.9 & 69.1 & - & - \\
  DML~\cite{Zhang-2018-CVPR} & 89.3 & 70.5 & - & - \\
  MGCAM~\cite{Song-2018-CVPR} & 83.8 & 74.3 & - & - \\
  DPFL~\cite{Cheng-2016-CVPR} & 88.9 & 73.1 & 79.2 & 60.6 \\
  PAN~\cite{Zheng-2018-TCSVT} & 82.8 & 63.4 & 71.6 & 51.5 \\
  PoseTransfer~\cite{Zhong-2018-CVPR} & 87.7 & 68.9 & 78.5 & 56.9 \\
  AlignedReID~\cite{zhang2017alignedreid} & 89.2 & 72.8 & 79.3 & 65.6 \\
  SVDNet~\cite{sun2017svdnet} & 82.3 & 62.1 & 76.7 & 56.8 \\
  CamStyle~\cite{zhong2017camera} & 89.2 &  71.6 & 78.6  &  57.6 \\
  PN-GAN~\cite{qian2017pose} & 89.4 & 72.6 & 73.6 &  53.2 \\
  \revision{SPreID~\cite{kalayeh2018human}} & 92.3 & 80.5 & 83.8 & 69.3\\
  \revision{FD-GAN~\cite{ge2018fd}} & 90.5 & 77.7 & 80.0 & 64.5 \\
  \revision{Part Aligned~\cite{suh2018part}} & 93.8 & 79.9 & 83.5 & 69.2 \\
  \revision{PCB~\cite{sun2018beyond}} & 93.2 & 81.7 & 82.9 & 67.1 \\
  \revision{DG-Net~\cite{zheng2019joint}} & 94.8 & \underline{86.0} & \underline{86.6} & \underline{74.8} \\
  \midrule
  \revision{Ours (FD-GAN)} & 91.5 & 78.1 & 81.8 & 65.6 \\
  \revision{Ours (Part Aligned)} & \underline{94.9} & 80.8 & 84.1 & 70.3 \\
  \revision{Ours (PCB)} & 94.4 & 82.5 & 84.3 & 67.9 \\
  \revision{Ours (DG-Net)} & \textbf{95.7} & \textbf{86.8} & \textbf{87.8} & \textbf{75.8} \\
  \bottomrule
  \end{tabular}
  }
  \end{center}
  \vspace{-4.0mm}
\end{table}

\subsubsection{Cross-Resolution Vehicle Re-ID}

Similar to cross-resolution person re-ID, we also consider multiple low-resolution (MLR) setting for vehicle re-ID and evaluate the proposed method on one \emph{synthetic} and one \emph{real-world} benchmarks. To construct the synthetic MLR-VeRi$776$ dataset, we down-sample images taken by one camera by a randomly selected down-sampling rate $r \in \{2, 3, 4\}$, whereas the images taken by the other camera(s) remain unchanged. The VRIC dataset is a \emph{genuine} and more challenging dataset for evaluating MLR vehicle re-ID and contains realistic images of multiple resolutions.

\revision{We compare our approach with cross-resolution re-ID methods, including FSRCNN-reID~\cite{dong2016accelerating}, SING~\cite{jiao2018deep}, CSR-GAN~\cite{wang2018cascaded}, and MSVF~\cite{kanaci2018vehicle}, and standard vehicle re-ID approaches, including Siamese-Visual~\cite{shen2017learning}, OIFE~\cite{wang2017orientation}, and VAMI~\cite{zhou2018aware}.} For cross-resolution re-ID methods, the training set contains HR images and LR ones with all three down-sampling rates $r \in \{2, 3, 4\}$ for each vehicle. For standard vehicle re-ID approaches, the training set comprises only HR images for each vehicle.

{\flushleft {\bf Results.}}
Table~\ref{table:exp-vehicle-reid-mlr} presents the quantitative results recorded at ranks $1$ and $5$, and mAP on the two adopted datasets. We observe that our proposed algorithm achieves the state-of-the-art performance on both datasets. While our method is designed for cross-resolution person re-ID, the favorable performance (about $4\%$ performance gains at rank 1 on both datasets) over all competing approaches (some of the competing methods are particularly tailored for cross-resolution vehicle re-ID task) demonstrates the generalization of our proposed algorithm.

\begin{figure}[t]
  \centering
  \includegraphics[width=\linewidth]{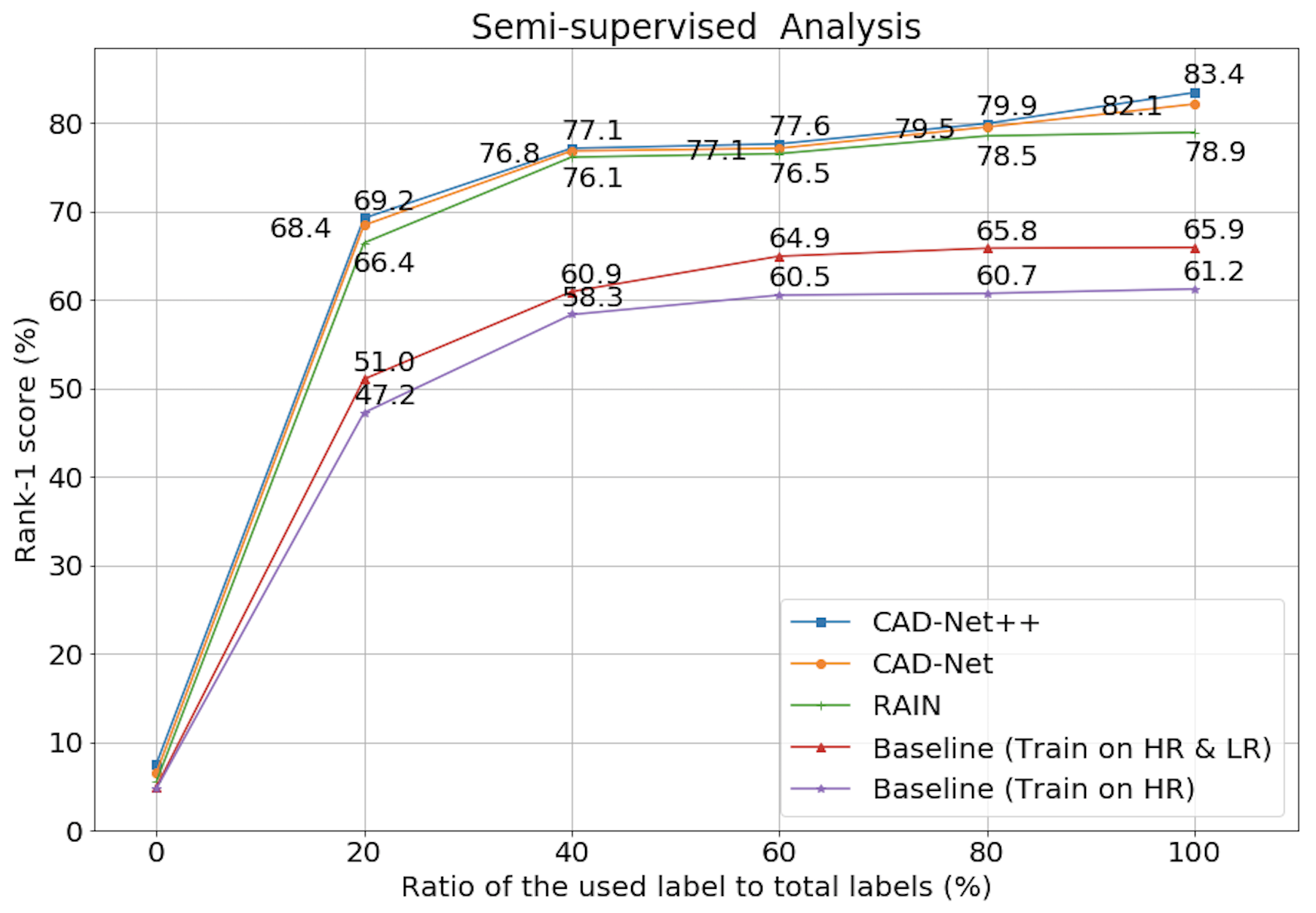}
  \caption{\textbf{Semi-supervised cross-resolution person re-ID on the MLR-CUHK03 dataset (\%).} Even if the training data is only partially labeled, our method still exhibits sufficient ability in re-identifying person images of various resolutions.}
  \label{fig:exp-semi}
  \vspace{-4.0mm}
\end{figure}

\subsection{Evaluation of Standard Setting}

\revision{To examine if our method still improves re-ID performance when no significant resolution variations are present, we consider standard person re-ID and compare with existing approaches, including JLML~\cite{Li-2017-IJCAI}, TriNet~\cite{Hermans-2017-arXiv}, DML~\cite{Zhang-2018-CVPR}, MGCAM~\cite{Song-2018-CVPR}, DPFL~\cite{Cheng-2016-CVPR}, PAN~\cite{Zheng-2018-TCSVT}, PoseTransfer~\cite{Zhong-2018-CVPR}, AlignedReID~\cite{zhang2017alignedreid}, SVDNet~\cite{sun2017svdnet}, CamStyle~\cite{zhong2017camera}, PN-GAN~\cite{qian2017pose}, SPreID~\cite{kalayeh2018human}, FD-GAN~\cite{ge2018fd}, Part Aligned~\cite{suh2018part}, PCB~\cite{sun2018beyond}, and DG-Net~\cite{zheng2019joint}, in which the training set contains HR images for each identity only. As for our model, we take the same training set and \emph{augment} it by down-sampling each image with three down-sampling rates $r \in \{2, 3, 4\}$ (i.e., our training set contains HR images and LR ones with $r \in \{2, 3, 4\}$) per person.}

\begin{table*}[t]
  \scriptsize
  \caption{\textbf{Quantitative results of cross-resolution person re-ID on the MLR-CUHK03 test set.} \emph{Left block}: resolutions are seen during training. \emph{Right block}: resolution is not seen during training. The bold and underlined numbers indicate top two results, respectively.}
  \vspace{-2.0mm}
  \centering
  \label{table:image-comparison}
  \resizebox{\linewidth}{!}
  {
  \begin{tabular}{l|ccc|c|ccc|c}
  \toprule
  \multirow{2}{*}{Method} & \multicolumn{4}{c|}{Down-sampling rate $r \in \{2, 3, 4\}$ (seen)} & \multicolumn{4}{c}{Down-sampling rate $r = 8$ (unseen)}\\\cmidrule{2-9} 
  & SSIM $\uparrow$ & PSNR $\uparrow$ & LPIPS~\cite{zhang2018unreasonable} $\downarrow$ & Rank 1 (\%) $\uparrow$ & SSIM $\uparrow$ & PSNR $\uparrow$ & LPIPS~\cite{zhang2018unreasonable} $\downarrow$ & Rank 1 (\%) $\uparrow$ \\
  \midrule
  CycleGAN~\cite{zhu2017unpaired} & 0.55 & 14.1 & 0.31 & 62.1 & 0.42 & 12.7 & 0.37 & 40.5\\
  SING~\cite{jiao2018deep} & 0.65 & 18.1 & 0.18 & 67.7 & 0.52 & 14.5 & 0.34 & 54.2\\
  CSR-GAN~\cite{wang2018cascaded} & \textbf{0.76} & \textbf{21.5} & 0.13 & 71.3 & 0.67 & 17.2 & 0.25 & 62.1\\
  \midrule
  CAD-Net~\cite{CAD-Net} & 0.73 & 20.2 & \underline{0.07} & \underline{82.1} & \underline{0.71} & \underline{19.8} & \underline{0.11} & \underline{78.6} \\
  Ours & \underline{0.75} & \underline{20.6} & \textbf{0.05} & \textbf{83.4} & \textbf{0.73} & \textbf{20.0} & \textbf{0.09} & \textbf{79.5}  \\
  \bottomrule
  \end{tabular}
  }
  \vspace{-2.0mm}
\end{table*}

\begin{figure*}[t]
  \centering
  \includegraphics[width=\linewidth]{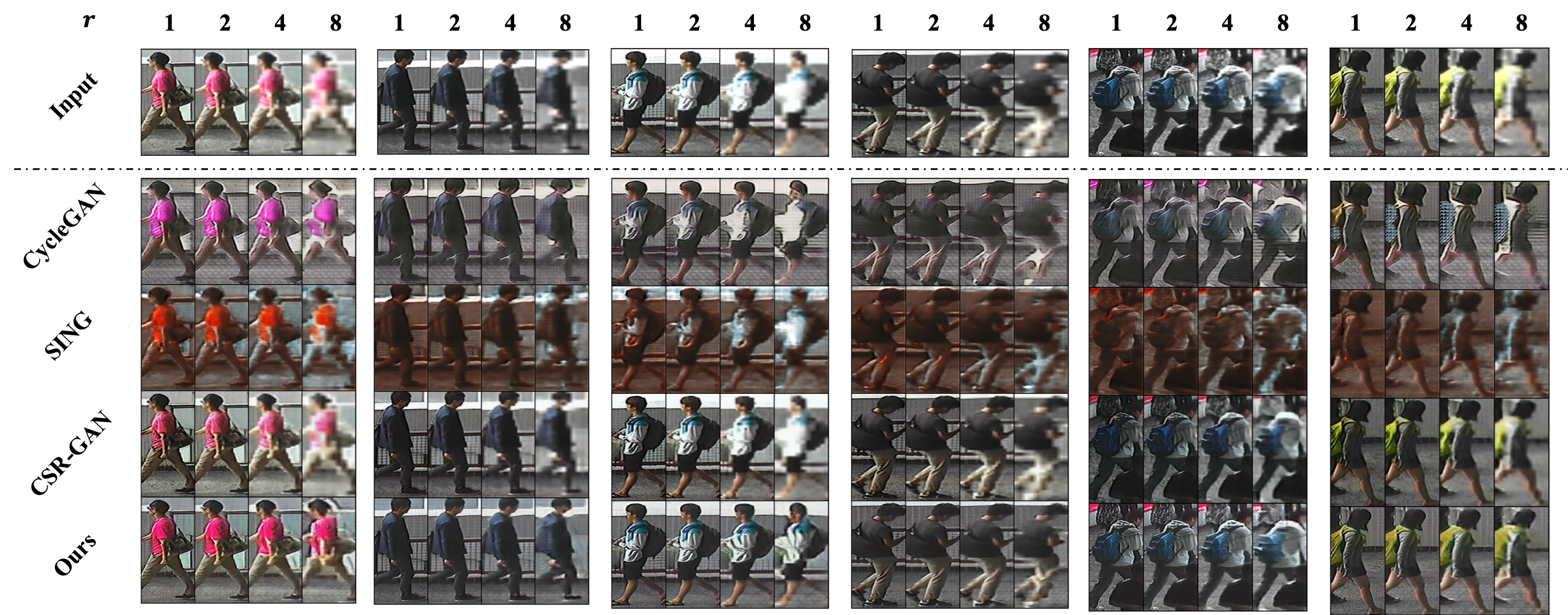}
  \vspace{-5.5mm}
  \caption{\revised{\textbf{Visual comparisons on the MLR-CUHK03 test set.} Given input images of various low resolutions (first row), we present the corresponding recovered HR images of CycleGAN~\cite{zhu2017unpaired}, SING~\cite{jiao2018deep}, CSR-GAN~\cite{wang2018cascaded}, and the proposed CRGAN.}}
  \label{fig:HR-img}
  \vspace{-3.0mm}
\end{figure*}

\revision{To demonstrate that our method can help improve the state-of-the-art methods, we initialize our HR encoder $\mathcal{F}$ with different pre-trained models as our method employs the same backbone (i.e., ResNet-$50$~\cite{he2016deep}) as most of these methods. Specifically, we initialize our HR encoder $\mathcal{F}$ with the pre-trained weights from FD-GAN~\cite{ge2018fd}, Part Aligned~\cite{suh2018part}, PCB~\cite{sun2018beyond}, and DG-Net~\cite{zheng2019joint}, and denote these variants of our method as Ours (FD-GAN), Ours (Part Aligned), Ours (PCB), and Ours (DG-Net), respectively.}

{\flushleft {\bf Results.}}
\revision{Table~\ref{table:exp-standard-reid} reports the quantitative results recorded at rank $1$ and the mAP on the two adopted datasets. We observe that our method results in further improvements over existing methods, achieving the state-of-the-art performance on the Market-$1501$~\cite{zheng2015scalable} and DukeMTMC-reID~\cite{zheng2017unlabeled} datasets in the standard person re-ID setting.}

The above quantitative results demonstrate that when significant resolution variations are present (i.e., cross-resolution setting), methods developed for standard re-ID suffer from the negative effect caused by the resolution mismatch issue. When considering the standard re-ID setting, the proposed method achieves further improvements over existing methods.

\subsection{Evaluation of Semi-Supervised Setting}

In the following, we conduct a series of semi-supervised experiments, and evaluate whether the proposed CAD-Net++ remains effective for cross-resolution person re-ID when only a subset of the labeled training data is available. Namely, less labeled training data can be used when computing the classification loss $\mathcal{L}_\mathrm{cls}$ (Eq. (\ref{eq:cls})).

We evaluate the proposed method on the MLR-CUHK$03$ dataset. For performance evaluation, we choose $k\%$ of the training data and keep their labels, while ignoring the labels of the rest, for $k\in\{0, 20, 40, 60, 80, 100\}$. Note that the unlabeled data are still utilized in optimizing the multi-scale feature-level adversarial loss (Eq. (\ref{eq:adv_multi})), the HR reconstruction loss (Eq. (\ref{eq:rec})), the image-level adversarial loss (Eq. (\ref{eq:adv_loss_image})), and the feature consistency loss (Eq. (\ref{eq:consist})). We compare our proposed approach with RAIN~\cite{chen2019learning} and two baseline methods: ``Baseline (train on HR)''~\cite{chen2019learning} and ``Baseline (train on HR \& LR)''~\cite{chen2019learning}.

Figure~\ref{fig:exp-semi} presents the model performance at rank $1$. We observe that without any label information, our method achieves $8\%$ at rank $1$. When the fraction of labeled data is increased to $20\%$, our model reaches $69.2\%$ at rank $1$, and is even better than SING~\cite{jiao2018deep} ($67.7\%$ at rank $1$) learned with $100\%$ labeled data. When the fraction of labeled data is set to $40\%$, our model achieves $77.1\%$ at rank $1$ and compares favorably against most existing approaches reported in Table~\ref{table:exp-ReID} that are learned with $100\%$ labeled data.

The promising results demonstrate that sufficient re-ID ability is exhibited by our method even if only a small portion of labeled training data are available. This favorable property increases the applicability of the proposed CAD-Net++ in real-world re-ID applications. We attribute this property to the elaborately developed loss functions. Except for the classification loss in Eq. (\ref{eq:cls}), all other loss functions can utilize unlabeled training data to regularize model training.

\subsection{Evaluation of the Recovered HR Images}

To demonstrate that our CRGAN is capable of recovering the missing details in LR input images of varying and even unseen resolutions, we evaluate the quality of the recovered HR images on the MLR-CUHK03 \emph{test set} using SSIM, PSNR, and LPIPS~\cite{zhang2018unreasonable} metrics. We employ the ImageNet-pretrained AlexNet~\cite{krizhevsky2012imagenet} when computing LPIPS. We compare our CRGAN with CycleGAN~\cite{zhu2017unpaired}, SING~\cite{jiao2018deep}, and CSR-GAN~\cite{wang2018cascaded}. For CycleGAN~\cite{zhu2017unpaired}, we train its model to learn a mapping between LR and HR images. We report the quantitative results of the recovered image quality and person re-ID in Table~\ref{table:image-comparison} with two different settings: (1) LR images of resolutions seen during training, i.e., $r \in \{2, 3, 4\}$ and (2) LR images of unseen resolution, i.e., $r = 8$.

For seen resolutions (i.e., left block), we observe that our results using SSIM and PSNR metrics are slightly worse than the CSR-GAN~\cite{wang2018cascaded} while compares favorably against SING~\cite{jiao2018deep} and CycleGAN~\cite{zhu2017unpaired}. However, our method performs favorably against these three methods using LPIPS metric and achieves the state-of-the-art performance when evaluating on cross-resolution person re-ID task. These results indicate that (1) SSIM and PSNR metrics are low-level pixel-wise metrics, which do not reflect high-level perceptual tasks and (2) the end-to-end learning of cross-resolution person re-ID would result in better re-ID performance and recover more perceptually realistic HR images as reflected by LPIPS. 

For unseen resolution (i.e., right block), our method performs favorably against all three competing methods on all the adopted evaluation metrics. These results suggest that our method is capable of handling unseen resolution (i.e., $r = 8$) with favorable performance in terms of both image quality and person re-ID. Note that we only train our model with HR images and LR ones with $r \in \{2, 3, 4\}$.

\begin{figure}[t]
  \centering
  \includegraphics[width=\linewidth]{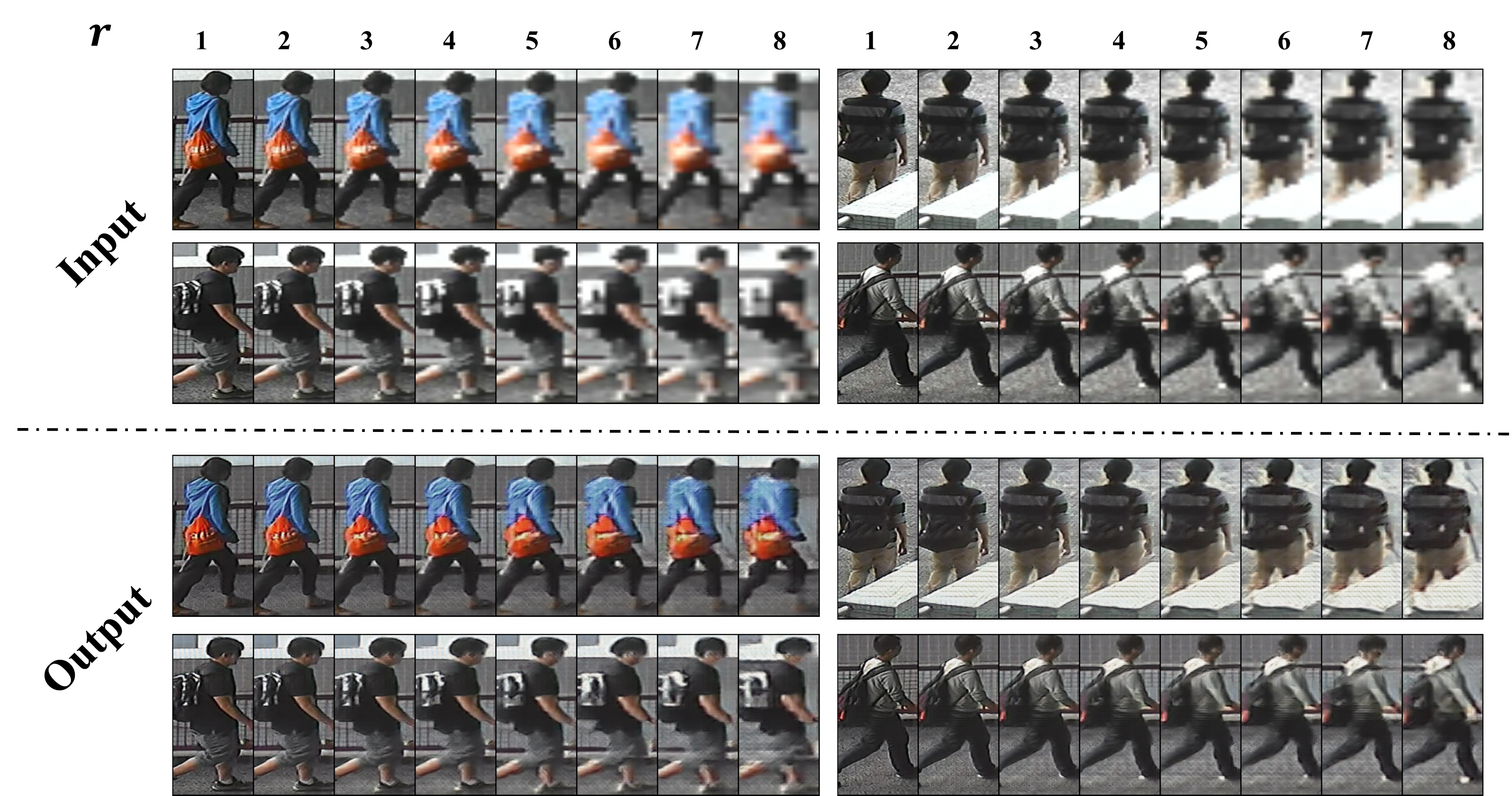}
  \vspace{-4.5mm}
  \caption{\textbf{Recovered HR images on the MLR-CUHK03 test set.} Given images of the same identity with eight different down-sampling rates, we present the corresponding HR images recovered by our CRGAN.}
  \label{fig:HR-uni}
  \vspace{-5.0mm}
\end{figure}

Figure~\ref{fig:HR-img} presents six examples. For each person, there are four different resolutions (i.e., $r \in \{1, 2, 4, 8\}$). Note that images with down-sampling rate $r = 1$ indicate that the images remain their original sizes and are the corresponding HR images of the LR ones. We observe that when LR images with down-sampling rate $r = 8$ are given, our model recovers the HR details with the highest quality among all competing methods. In addition, we present four examples in Figure~\ref{fig:HR-uni} to show that the proposed CRGAN is able to recover the missing details in LR images of \emph{various} resolutions. In each example, we have eight input images with down-sampling rates $r \in \{1, 2, 3, 4, 5, 6, 7, 8\}$, respectively, where $r \in \{1, 2, 3, 4\}$ are seen during training while the rest are unseen. The corresponding recovered HR images are displayed in Figure~\ref{fig:HR-uni}. Both quantitative and qualitative results above confirm that our model can handle \emph{a range of} seen resolutions and generalize well to \emph{unseen} resolutions using just one single model, i.e., CRGAN.

\begin{table}[t]
  \small
  \ra{1.3}
  \begin{center}
  \caption{\revised{\textbf{Ablation study of the loss functions on MLR-CUHK03.} The bold and underlined numbers indicate top two results, respectively.}}
  \vspace{-2.0mm}
  \label{table:exp-abla}
  \resizebox{\linewidth}{!} 
  {
  \begin{tabular}{l|ccc|c}
  \toprule
  %
  %\multirow{2}{*}{Method} & \multicolumn{3}{c|}{Quality of the recovered HR images} & Person re-ID (\%) \\
  %
  Method & SSIM $\uparrow$ & PSNR $\uparrow$  & LPIPS~\cite{zhang2018unreasonable} $\downarrow$ & Rank 1 (\%) $\uparrow$ \\
  \midrule
  Ours & \textbf{0.75} & \underline{20.6} & \textbf{0.05} & \textbf{83.4} \\
  Ours w/o $\mathcal{L}_\mathrm{adv}^{\mathcal{D}_{F}}$ & 0.54 & 14.2 & 0.34 & 67.6 \\
  Ours w/o $\mathcal{L}_\mathrm{rec}$& 0.45 & 12.9 & 0.40 & 66.7 \\
  Ours w/o $\mathcal{L}_\mathrm{adv}^{\mathcal{D}_{I}}$& 0.67 & 18.5 & 0.17 & 79.8 \\
  Ours w/o $\mathcal{L}_\mathrm{cls}$ & \underline{0.73} & \textbf{21.1} & \underline{0.12} & 1.7 \\
  Ours w/o $\mathcal{L}_\mathrm{consist}$ & \textbf{0.75} & 20.5 & \textbf{0.05} & \underline{82.6} \\
  \bottomrule
  \end{tabular}
  }
  \end{center}
  \vspace{-4.0mm}
\end{table}

\begin{table*}[t]
  \small
  \ra{1.3}
  \begin{center}
  \caption{\revision{\textbf{Ablation study of the feature consistency loss on cross-resolution person and vehicle re-ID tasks.} The bold numbers indicate the better results (\%).}}
  \vspace{-2.0mm}
  \label{table:ablation-consis-oss}
  \resizebox{\linewidth}{!} 
  {
  \begin{tabular}{l|cc|cc|cc|cc|cc|cc|cc}
  \toprule
  \multirow{3}{*}{Method} & \multicolumn{10}{c|}{Cross-resolution person re-ID} & \multicolumn{4}{c}{Cross-resolution vehicle re-ID} \\
  & \multicolumn{2}{c|}{MLR-CUHK03} & \multicolumn{2}{c|}{MLR-VIPeR} & \multicolumn{2}{c|}{CAVIAR} & \multicolumn{2}{c|}{MLR-Market-1501} & \multicolumn{2}{c|}{MLR-DukeMTMC-reID} & \multicolumn{2}{c|}{MLR-VeRi776} & \multicolumn{2}{c}{VRIC} \\
  & Rank 1 & Rank 5 & Rank 1 & Rank 5 & Rank 1 & Rank 5 & Rank 1 & Rank 5 & Rank 1 & Rank 5 & Rank 1 & Rank 5 & Rank 1 & Rank 5 \\
  \midrule
  Ours w/o $\mathcal{L}_\mathrm{consist}$ & 82.6 & 96.5 & 41.3 & 66.4 & 41.5 & 75.6 & 80.4 & 90.8 & 73.7 & 84.4 & 66.5 & 84.1 & 47.3 & 67.4 \\
  Ours & \textbf{83.4} & \textbf{98.1} & \textbf{43.4} & \textbf{68.7} & \textbf{43.1} & \textbf{76.5} & \textbf{84.1} & \textbf{93.0} & \textbf{77.2} & \textbf{88.1} & \textbf{68.7} & \textbf{85.3} & \textbf{50.1} & \textbf{68.2} \\
  \bottomrule
  \end{tabular}
  }
  \end{center}
%   \vspace{-2.0mm}
\end{table*}

\begin{table*}[t]
  \small
  \ra{1.3}
  \begin{center}
  \caption{\revision{\textbf{Ablation study of multi-scale adversarial learning on MLR-CUHK03.}}}
  \vspace{-2.0mm}
  \label{table:exp-multi-abla}
%   \resizebox{\linewidth}{!} 
  {
  \begin{tabular}{l|c|ccc|c}
  \toprule
  Feature level of $\mathcal{D}_F$ & Rank 1 (\%) $\uparrow$ & SSIM $\uparrow$ & PSNR $\uparrow$ & LPIPS~\cite{zhang2018unreasonable} $\downarrow$ & Number of parameters of $\mathcal{D}_F$ \\
  \midrule
  $j=1$ & {82.9} & {0.73} & {20.3} & {0.07} & 4.1 M \\
  $j \in \{1, 2\}$ & {83.4} & {0.75} & {20.6} & {0.05} & 10.5 M \\
  $j \in \{1, 2, 3\}$ & {83.6} & {0.77} & {20.8} & {0.05} & 19.6 M \\
  $j \in \{1, 2, 3, 4\}$ & {83.6} & {0.78} & {20.9} & {0.04} & 35.6 M \\
  $j \in \{1, 2, 3, 4, 5\}$  & {83.7} & {0.78} & {21.1} & {0.04} & 56.7 M \\
  \bottomrule
  \end{tabular}
  }
  \end{center}
  \vspace{-4.0mm}
\end{table*}

\subsection{Ablation Study}

\subsubsection{Loss Functions}

To analyze the importance of each developed loss function, we conduct an ablation study on the MLR-CUHK03 dataset. Table~\ref{table:exp-abla} reports the quantitative results of the recovered HR images and the performance of cross-resolution person re-ID recorded at rank $1$.

{\flushleft {\bf Multi-scale feature-level adversarial loss $\mathcal{L}_\mathrm{adv}^{\mathcal{D}_{F}}$.}}
Without $\mathcal{L}_\mathrm{adv}^{\mathcal{D}_{F}}$, our model does not learn the resolution-invariant representations and thus suffers from the resolution mismatch issue. Significant performance drops in the recovered image quality and re-ID performance occur, indicating the importance of our method for learning the resolution-invariant representations to address the problem of resolution mismatch.

{\flushleft {\bf HR reconstruction loss $\mathcal{L}_\mathrm{rec}$.}}
Once $\mathcal{L}_\mathrm{rec}$ is excluded, there is no explicit supervision to guide the CRGAN to perform image recovery, and the model suffers from information loss in compressing visual images into semantic feature maps. Severe performance drops in terms of the recovered image quality and re-ID performance are hence caused.

{\flushleft {\bf Image-level adversarial loss $\mathcal{L}_\mathrm{adv}^{\mathcal{D}_{I}}$.}}
When $\mathcal{L}_\mathrm{adv}^{\mathcal{D}_{I}}$ is turned off, our model is not encouraged to produce perceptually realistic HR images as reflected by LPIPS, resulting in the performance drop of $3.6\%$ at rank $1$.

{\flushleft {\bf Classification loss $\mathcal{L}_\mathrm{cls}$.}}
Although our model is still able to perform image recovery without $\mathcal{L}_\mathrm{cls}$, our model cannot perform discriminative learning for person re-ID since data labels are not used during training. Thus, significant performance drop in person re-ID occurs.

\revision{
{\flushleft {\bf Consistency loss $\mathcal{L}_\mathrm{consist}$.}}
As shown in Table~\ref{table:exp-abla}, once $\mathcal{L}_\mathrm{consist}$ is disabled, the quality of the recovered HR images almost remains the same but a performance drop of $0.8\%$ at rank $1$ is occurred. To further evaluate the contribution of the feature consistency loss $\mathcal{L}_\mathrm{consist}$, we report the results of the Ours w/o $\mathcal{L}_\mathrm{consist}$ method on all five cross-resolution person re-ID datasets (i.e., MLR-CUHK$03$, MLR-VIPeR, CAVIAR, MLR-Market-$1501$, and MLR-DukeMTMC-reID) and all two cross-resolution vehicle re-ID datasets (i.e., MLR-VeRi$776$ and VRIC) in Table~\ref{table:ablation-consis-oss}. We observe that without the feature consistency loss $\mathcal{L}_\mathrm{consist}$, our model suffers from performance drops on all seven datasets. While the performance improvement in person re-ID contributed by the feature consistency loss $\mathcal{L}_\mathrm{consist}$ is marginal on the MLR-CUHK$03$ dataset, our results show that incorporating such a simple loss function consistently improves the performance on all seven datasets of both tasks without increasing the model complexity and capacity (i.e., the number of model parameters remains the same). In particular, on the MLR-Market-$1501$ and MLR-DukeMTMC-reID datasets, introducing the feature consistency loss $\mathcal{L}_\mathrm{consist}$ results in $3.7\%$ and $3.5\%$ performance improvement at Rank $1$, respectively.
}

The ablation study demonstrates that the losses $\mathcal{L}_\mathrm{adv}^{\mathcal{D}_{F}}$, $\mathcal{L}_\mathrm{rec}$, and $\mathcal{L}_\mathrm{cls}$ are crucial to our method, while the losses $\mathcal{L}_\mathrm{adv}^{\mathcal{D}_{I}}$ and $\mathcal{L}_\mathrm{consist}$ are helpful for further improving the performance.

\vspace{-4.0mm}
\revision{\subsubsection{Multi-scale adversarial learning}}

\revision{For learning the resolution-invariant representations $f$, we perform an ablation study on the MLR-CUHK$03$ dataset to analyze the effect of introducing feature-level discriminators $\{\mathcal{D}_F^j\}$ at different feature levels (scales) to align feature distributions between HR and LR images. We report the results recorded at rank $1$ for cross-resolution person re-ID, evaluate the quality of the recovered HR images using SSIM, PSNR, and LPIPS~\cite{zhang2018unreasonable} metrics, and report the numbers of parameters of the feature-level discriminators $\{\mathcal{D}_F^j\}$. As shown in Table~\ref{table:exp-multi-abla}, the performance in rank $1$ of cross-resolution person re-ID and the quality of the recovered HR images improve as the number of feature distribution alignments increases. We observe that when introducing five feature-level discriminators $\{\mathcal{D}_F^j\}_{j=1}^{5}$ to align feature distributions at all five feature levels, our model reaches the best performance with the best quality of the recovered HR images, but the number of parameters of the feature-level discriminators $\{\mathcal{D}_F^j\}_{j=1}^{5}$ increases accordingly. To balance the parameter number (efficiency) and performance, we select the setting with the first two levels (i.e., $j \in \{1, 2\}$) for our method, and denote it as ``Ours (multi-level)'' for performance comparison.}

\begin{figure}[t]
  \vspace{-2.0mm}
  \centering
  \includegraphics[width=1.0\linewidth]{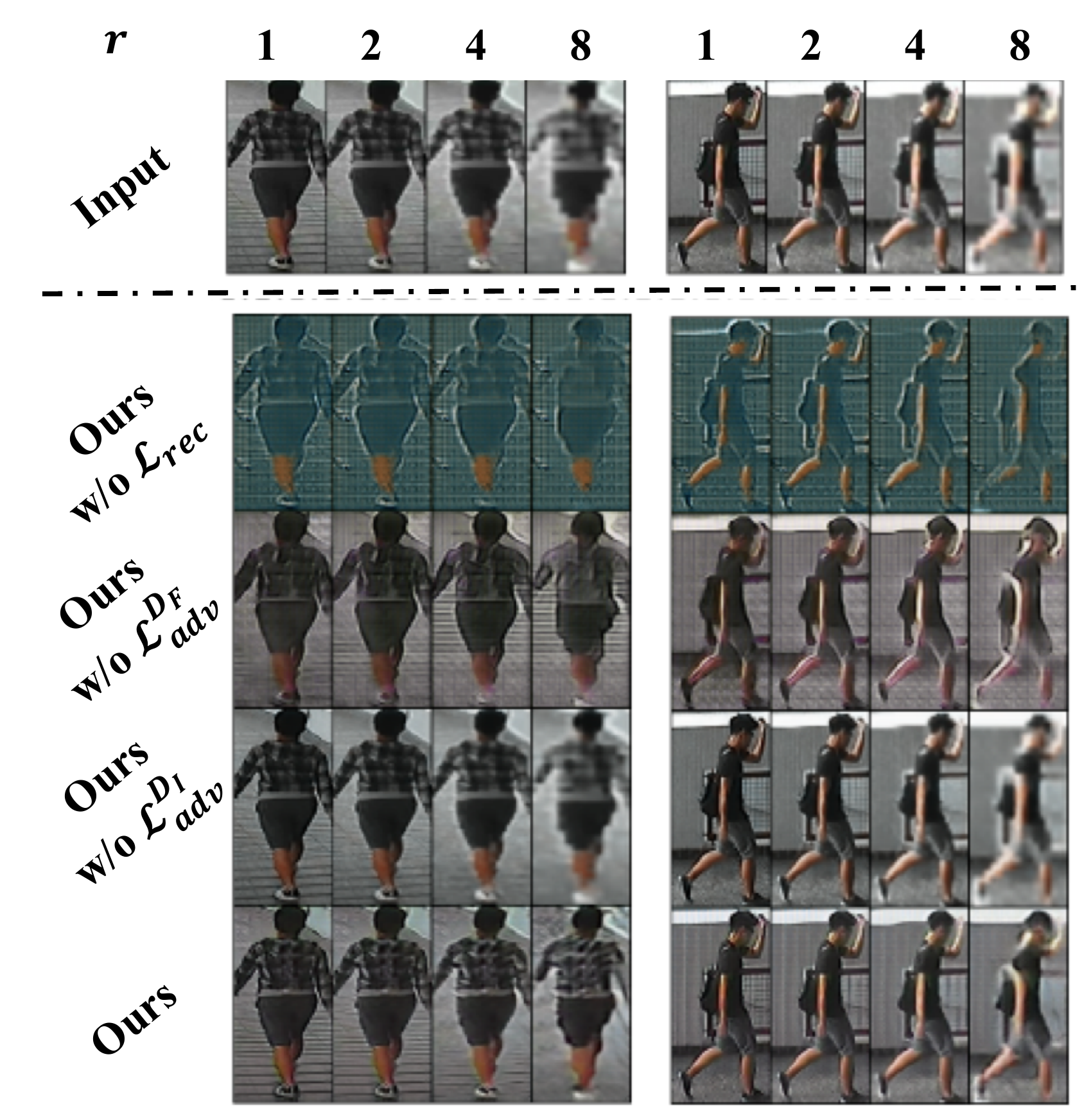}
  \vspace{-6.0mm}
  \caption{\revision{\textbf{Ablation study of the recovered HR images.} We present the recovered HR images obtained from our method as well as the ablation methods on the MLR-CUHK03 \emph{test set} (i.e., cross-resolution person re-ID setting).}}
  \label{fig:abl}
  \vspace{-5.0mm}
\end{figure}

\setlength{\threeimg}{0.33\textwidth}
\begin{figure*}[t]
  \centering
  \begin{subfigure}[b]{\threeimg}
    \centering
    \includegraphics[width=\linewidth]{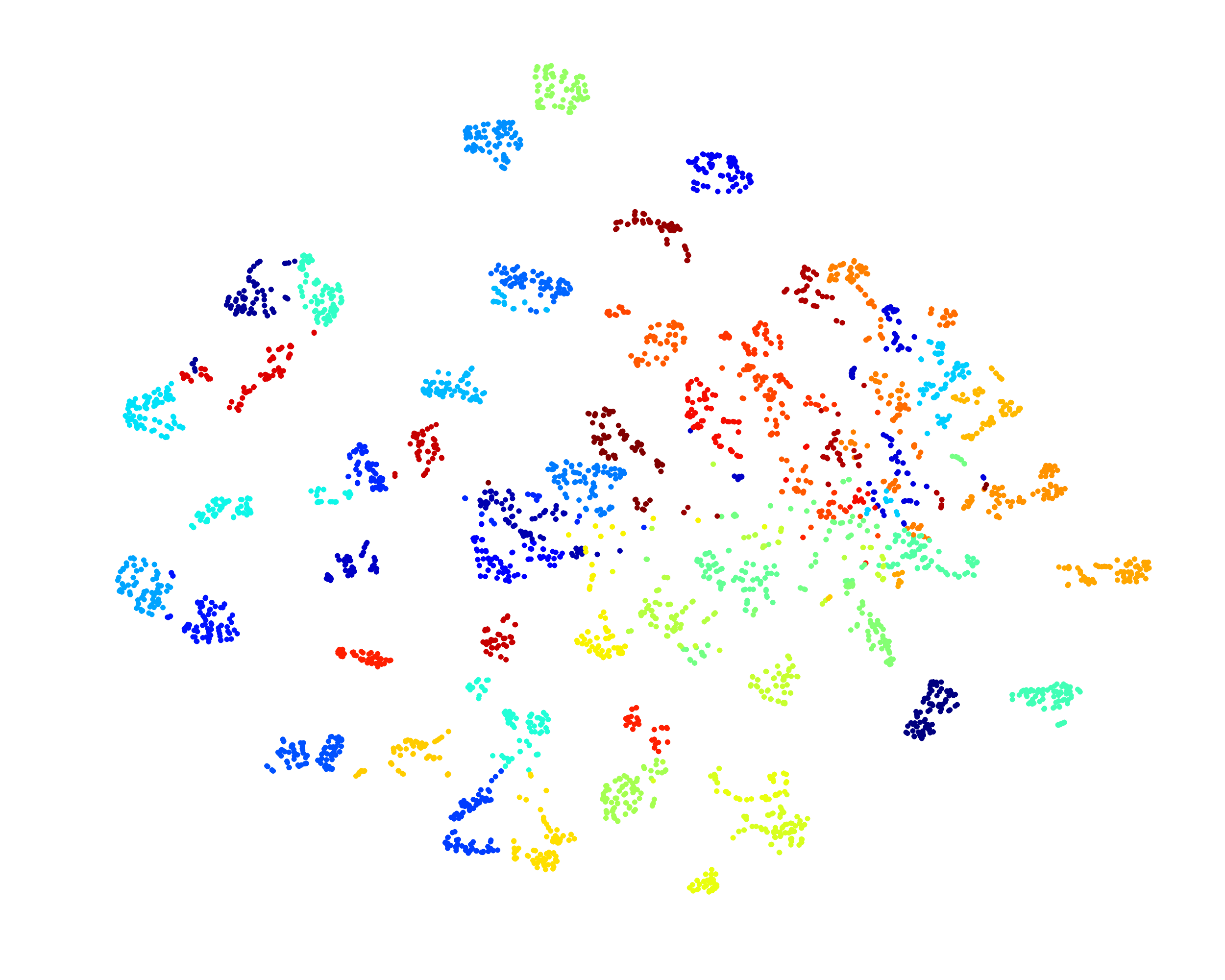}
    \vspace{-6.5mm}
    \caption{Ours w/o $\mathcal{L}_\mathrm{adv}^{\mathcal{D}_{F}}$: colorized w.r.t \textbf{identity}.}
    \label{fig:tsne-baseline}
  \end{subfigure}
  \hfill
  \begin{subfigure}[b]{\threeimg}
    \centering
    \includegraphics[width=\linewidth]{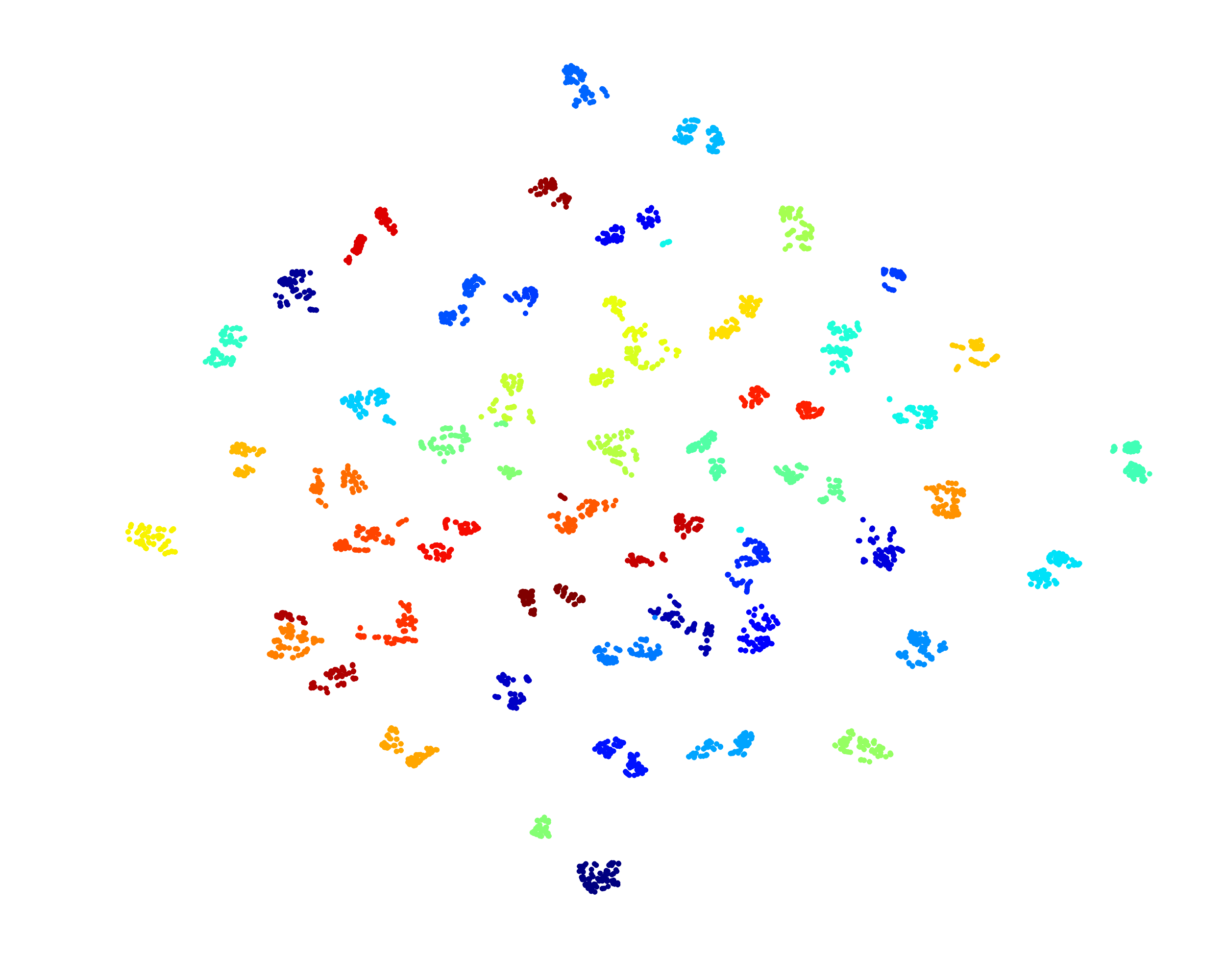}
    \vspace{-6.5mm}
    \caption{Ours: colorized w.r.t \textbf{identity}.}
    \label{fig:tsne-identity}
  \end{subfigure}
  \hfill
  \begin{subfigure}[b]{\threeimg}
    \centering
    \includegraphics[width=\linewidth]{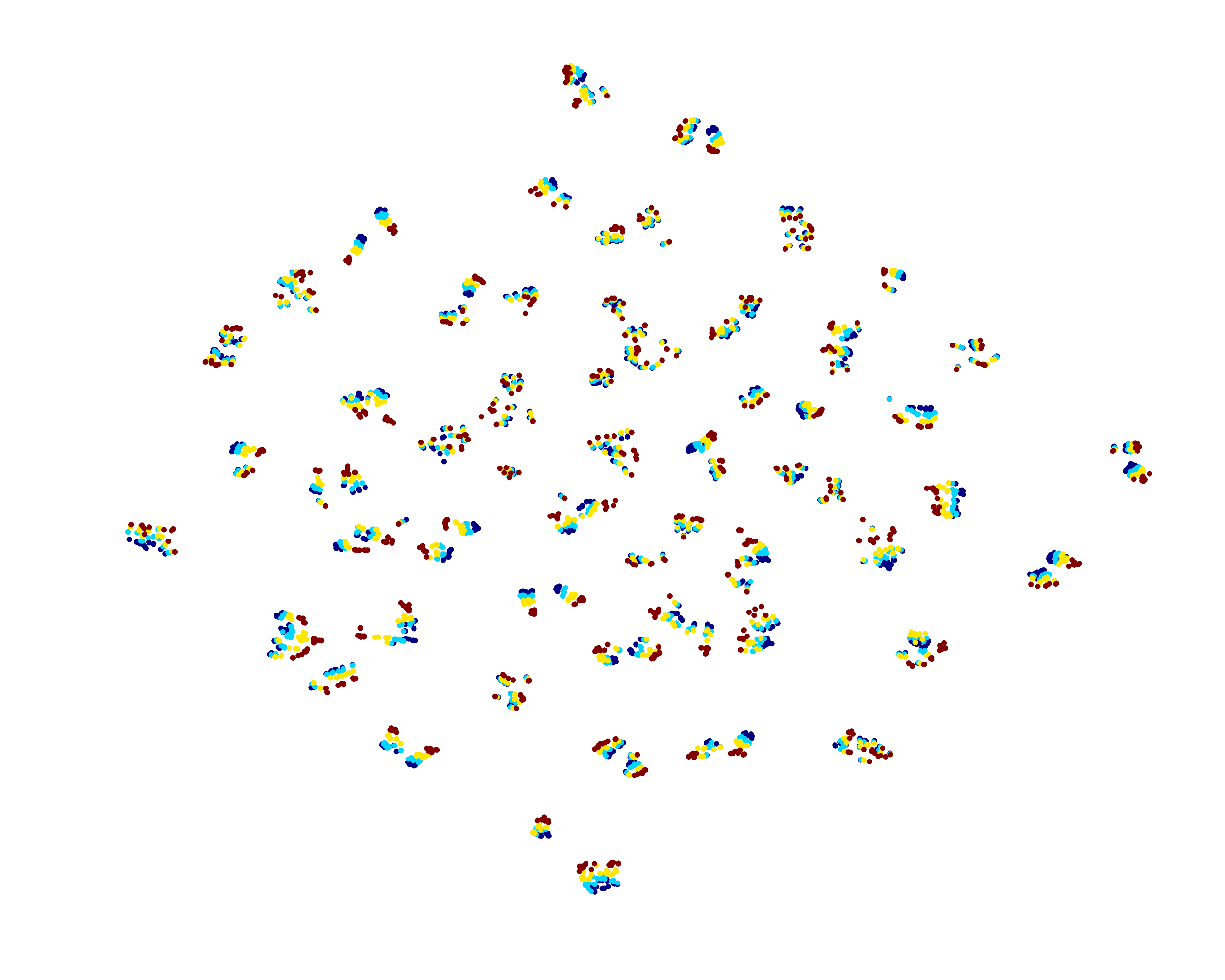}
    \vspace{-6.5mm}
    \caption{Ours: colorized w.r.t \textbf{resolution}.}
    \label{fig:tsne-resolution}
  \end{subfigure}
  \vspace{-5.0mm}
  \caption{\textbf{2D visualization of the resolution-invariant feature vector $\mathbfit{w}$ on the MLR-CUHK03 test set via t-SNE.} Data of different identities (each in a unique color) derived by our model \emph{without} and \emph{with} observing the feature-level adversarial loss $\mathcal{L}_\mathrm{adv}^{\mathcal{D}_{F}}$ are shown in (a) and (b), respectively. The same data but with resolution-specific colorization, \ie one color for each down-sampling rate $r \in \{1, 2, 4, 8\}$, are depicted in (c). Note that images with $r = 8$ are not seen during training.}
  \label{fig:tsne}
  \vspace{-2.0mm}
\end{figure*}

\begin{figure*}[t]
  \centering
  \begin{subfigure}[b]{0.33\linewidth}
    \includegraphics[width=\linewidth]{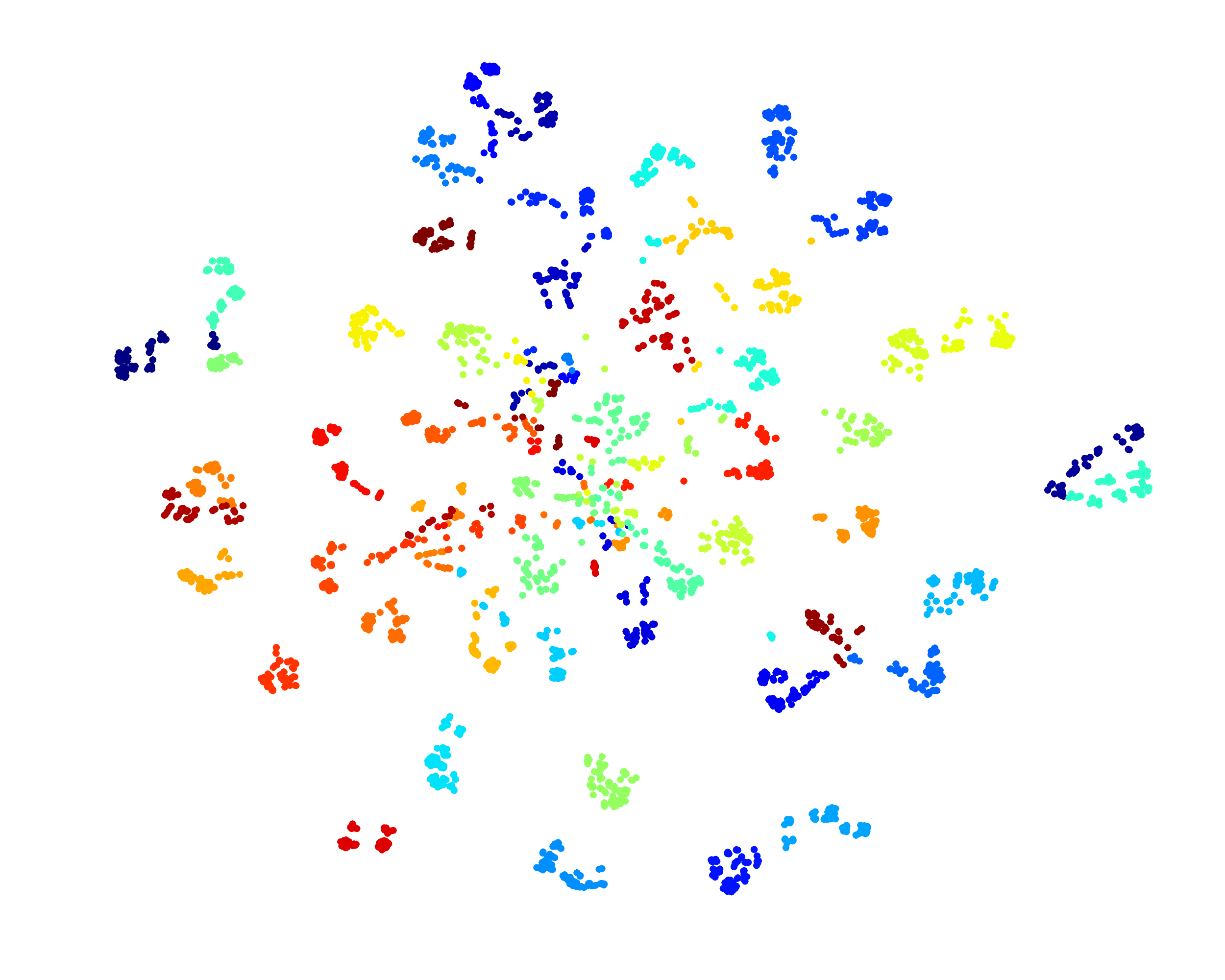}\\
    \vspace{-6.5mm}
    \caption{SING~\cite{jiao2018deep}}
    \label{fig:tsne-identity-naive}
  \end{subfigure}
  \hfill
  \begin{subfigure}[b]{0.33\linewidth}
    \includegraphics[width=\linewidth]{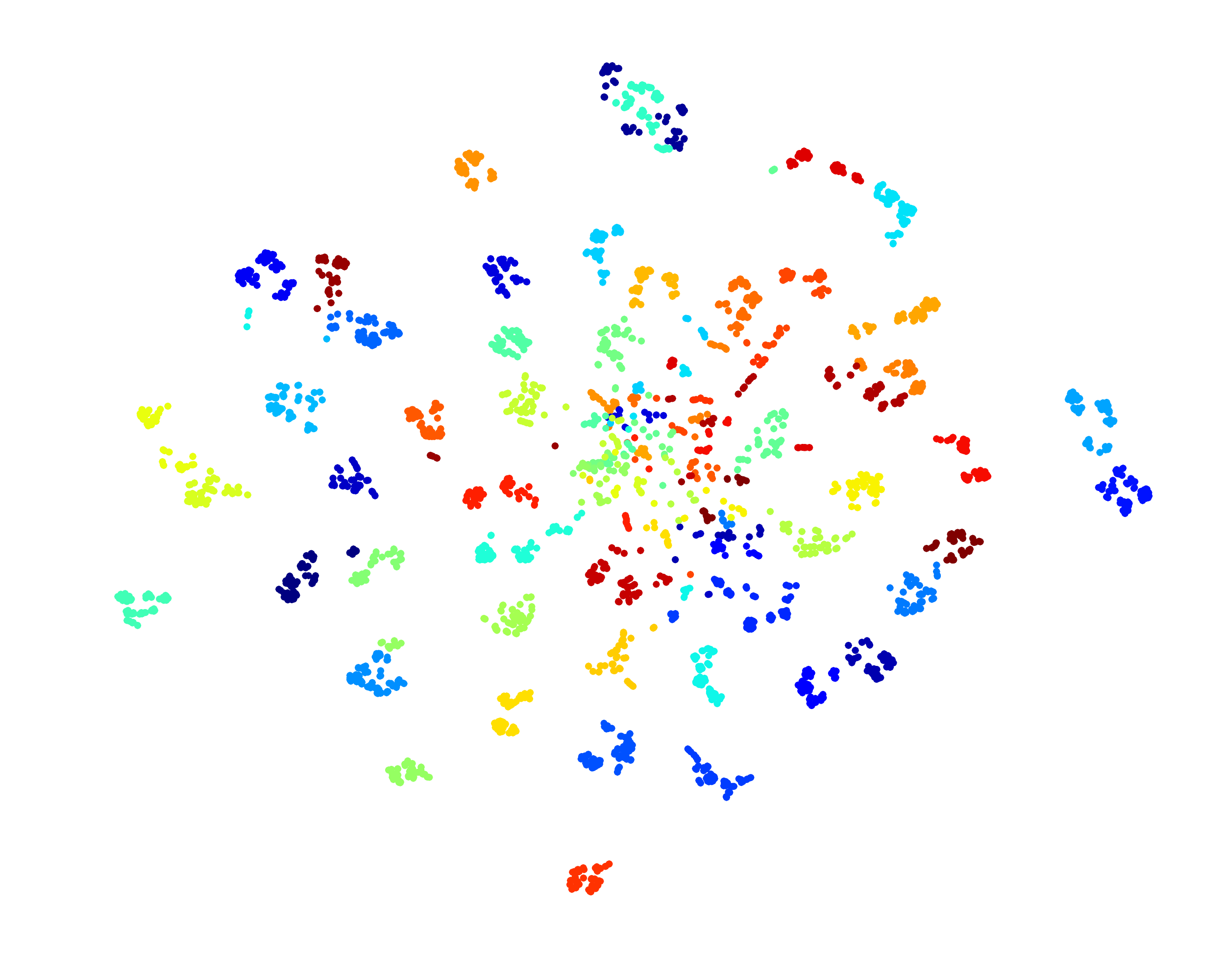}\\
    \vspace{-6.5mm}
    \caption{CSR-GAN~\cite{wang2018cascaded}}
    \label{fig:tsne-identity-better}
  \end{subfigure}
  \hfill
  \begin{subfigure}[b]{0.33\linewidth}
    \includegraphics[width=\linewidth]{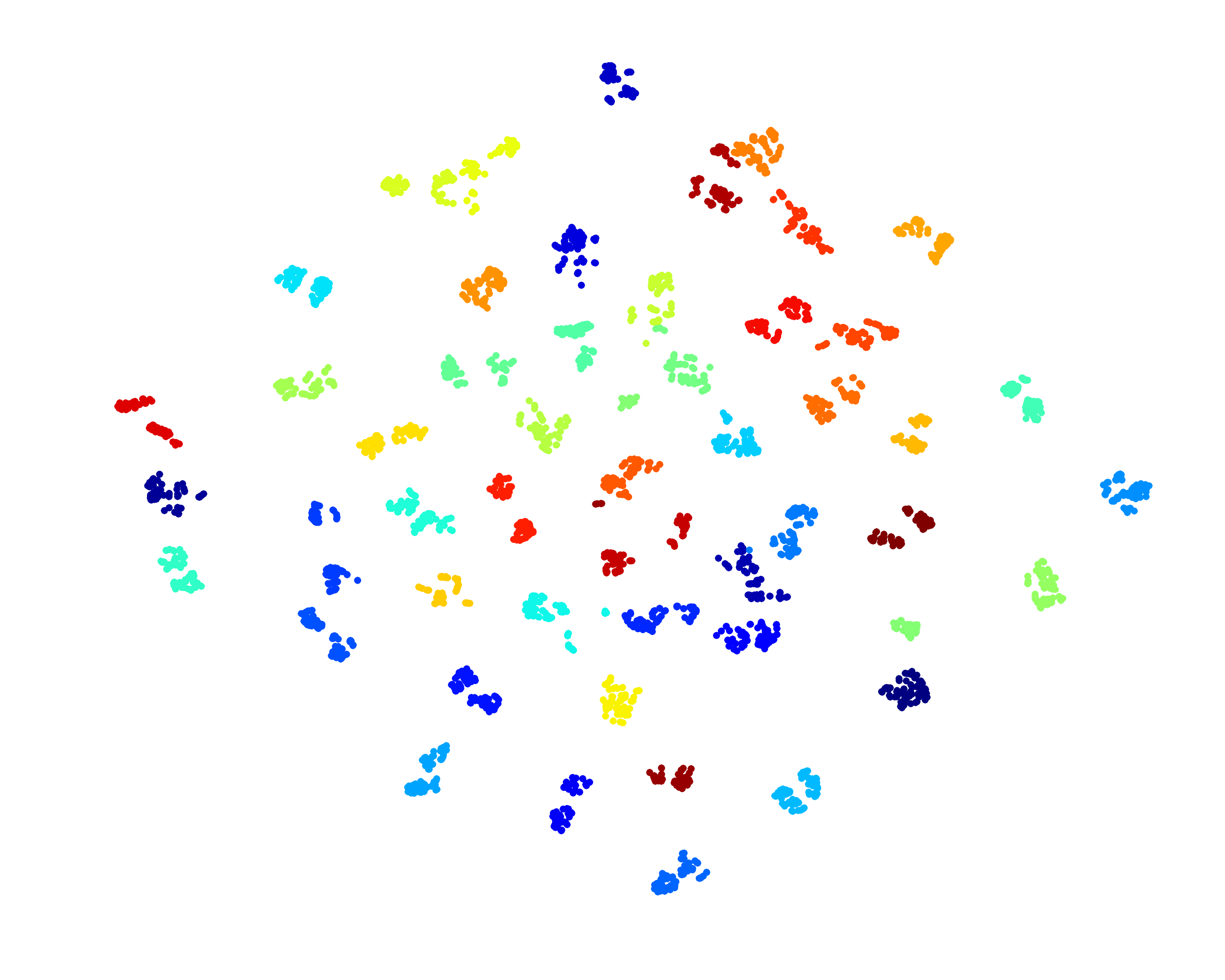}\\
    \vspace{-6.5mm}
    \caption{Ours}
    \label{fig:tsne-identity-ours}
  \end{subfigure}
  \vspace{-5.0mm}
  \caption{\revised{\textbf{Visual comparison of the re-ID feature vectors.} We present the visualization of the re-ID feature vectors of three methods, including (a) SING~\cite{jiao2018deep}, (b) CSR-GAN~\cite{wang2018cascaded}, and (c) our method, on the MLR-CUHK03 test set via t-SNE. There are 50 different identities, each of which is shown in a unique color.}}
  \label{fig:tsne-identity-comp}
  \vspace{-3.0mm}
\end{figure*}

\begin{figure*}[t!]
  \centering
  \begin{subfigure}[b]{0.33\linewidth}
    \includegraphics[width=\linewidth]{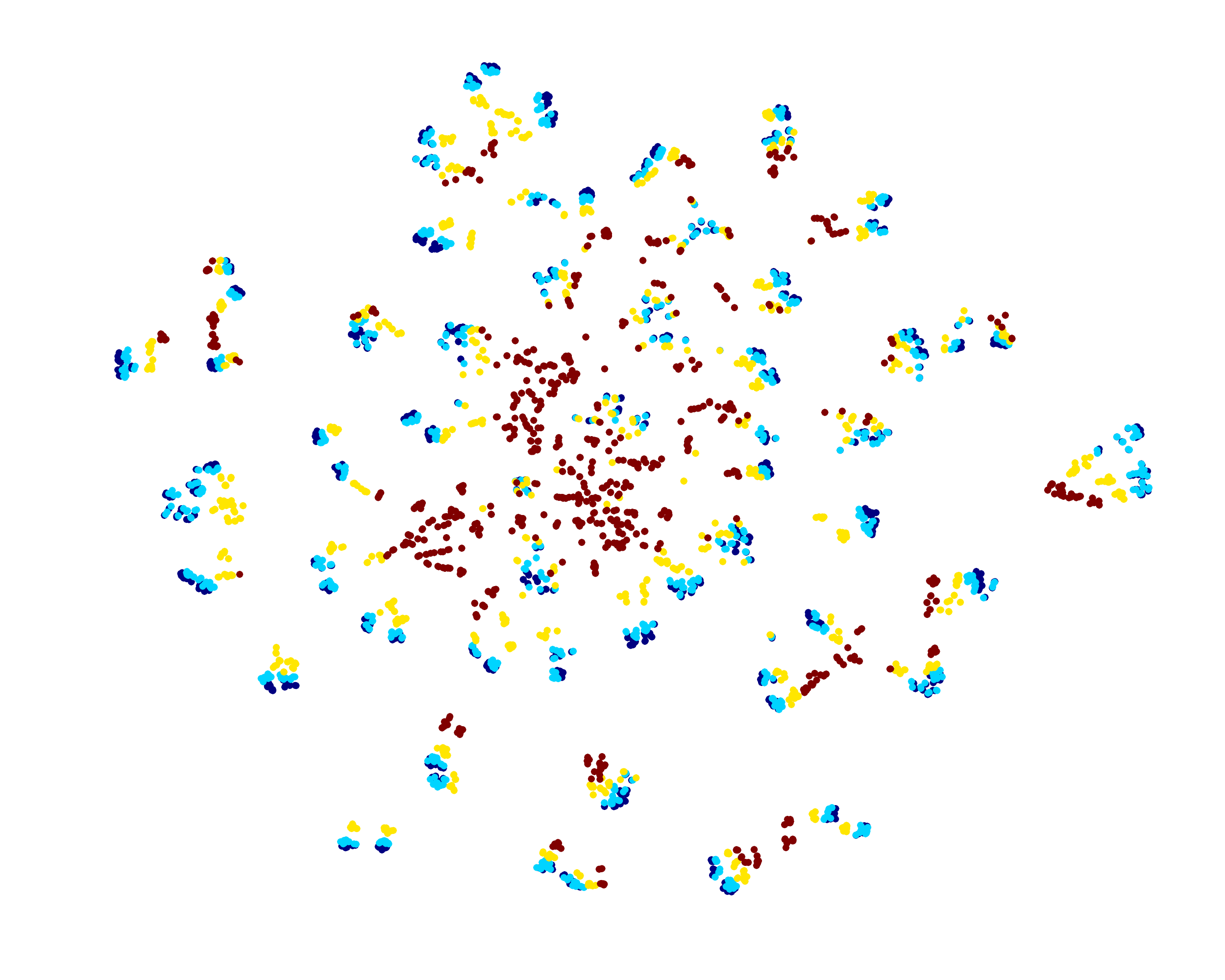}\\
    \vspace{-6.5mm}
    \caption{SING~\cite{jiao2018deep}}
    \label{fig:tsne-reso-naive}
  \end{subfigure}
  \hfill
  \begin{subfigure}[b]{0.33\linewidth}
    \includegraphics[width=\linewidth]{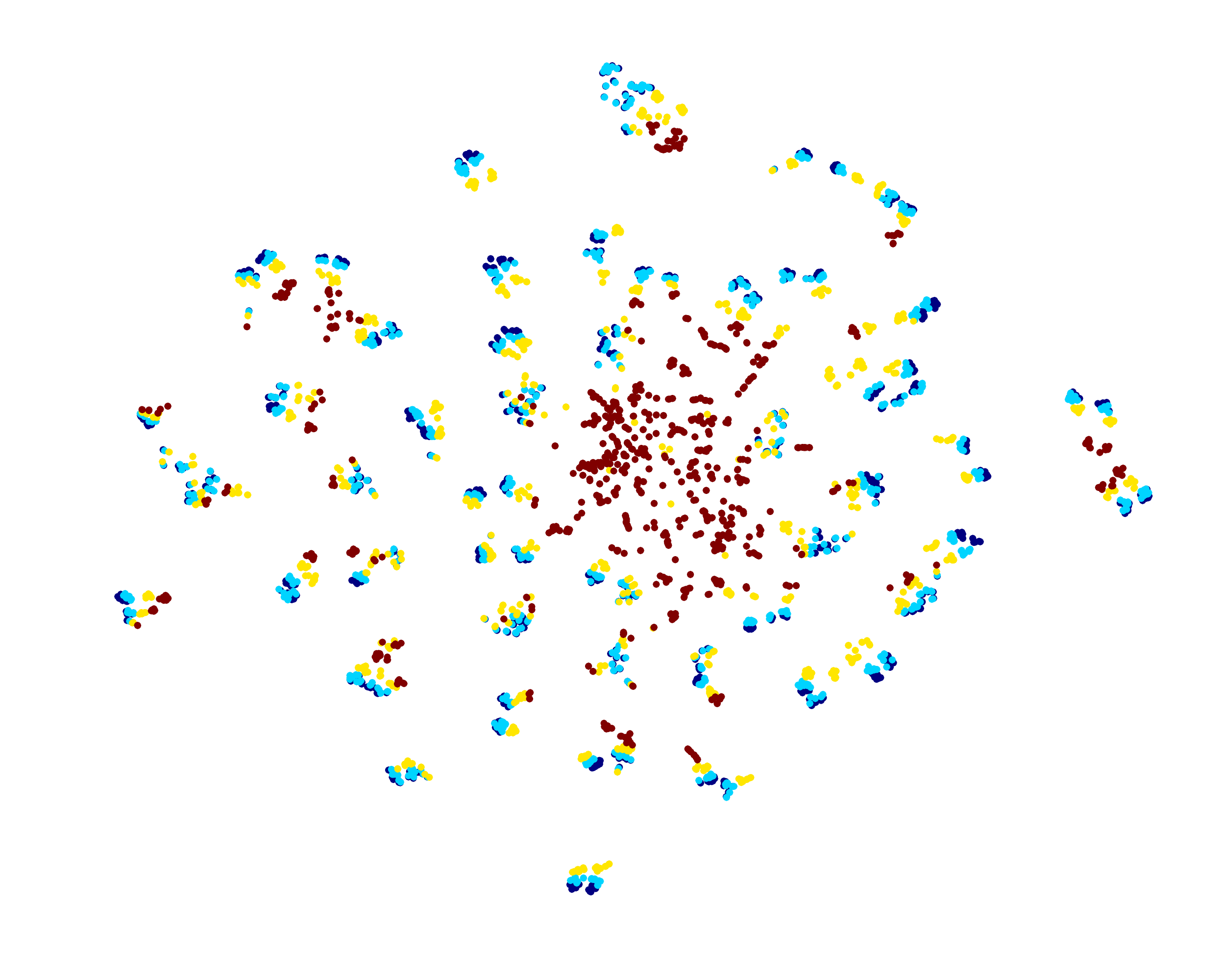}\\
    \vspace{-6.5mm}
    \caption{CSR-GAN~\cite{wang2018cascaded}}
    \label{fig:tsne-reso-better}
  \end{subfigure}
  \hfill
  \begin{subfigure}[b]{0.33\linewidth}
    \includegraphics[width=\linewidth]{img/Reso_tsne_50_ours_n_bar.pdf}\\
    \vspace{-6.5mm}
    \caption{Ours}
    \label{fig:tsne-reso-ours}
  \end{subfigure}
  \vspace{-5.0mm}
  \caption{\revised{\textbf{Visual comparison of the re-ID feature vectors.} We visualize the re-ID feature vectors of three methods, including (a) SING~\cite{jiao2018deep}, (b) CSR-GAN~\cite{wang2018cascaded}, and (c) our method, on the MLR-CUHK03 test set via t-SNE. We consider four different down-sampling rates $r \in \{1, 2, 4, 8\}$, each of which is marked by a unique color. Note that images with $r = 8$ are marked in red and are unseen during training.}}
  \label{fig:tsne-reso-comp}
  \vspace{-3.0mm}
\end{figure*}

\setlength{\elevenimg}{0.04\textwidth}
\setlength{\twoimg}{0.495\textwidth}
\begin{figure*}[t]
  \ra{1.3}
  \centering
  \begin{subfigure}{\twoimg}
    \begin{subfigure}[b]{\dimexpr\elevenimg+20pt\relax}
      \makebox[20pt]{\raisebox{15pt}{\rotatebox[origin=c]{90}{\cite{jiao2018deep}}}}%
      \includegraphics[width=0.95\elevenimg,cfbox=white 1pt 1pt]{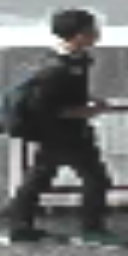}\\
      \makebox[20pt]{\raisebox{15pt}{\rotatebox[origin=c]{90}{\cite{wang2018cascaded}}}}%
      \includegraphics[width=0.95\elevenimg,cfbox=white 1pt 1pt]{rank-sample/Query_100_4.png}\\
      \makebox[20pt]{\raisebox{15pt}{\rotatebox[origin=c]{90}{Ours}}}%
      \includegraphics[width=0.95\elevenimg,cfbox=white 1pt 1pt]{rank-sample/Query_100_4.png}\\
      \vspace{-4.5mm}
      \caption*{\hspace{7mm} \scriptsize Query}
    \end{subfigure}
    \hskip .25em
    \tikz{\draw[-,blue, densely dashed, thick](0,-1.8) -- (0,3.25);}
    \hskip .25em
    \begin{subfigure}[b]{\elevenimg}
      \includegraphics[width=0.95\linewidth,cfbox=green 1pt 1pt]{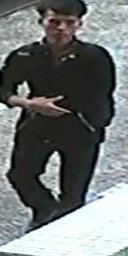}\\
      \includegraphics[width=0.95\linewidth,cfbox=green 1pt 1pt]{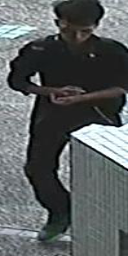}\\
      \includegraphics[width=0.95\linewidth,cfbox=green 1pt 1pt]{rank-sample/111.png}\\
      \vspace{-4.5mm}
      \caption*{\scriptsize Top 1}
    \end{subfigure}
    \hskip .5em
    \begin{subfigure}[b]{\elevenimg}
      \includegraphics[width=0.95\linewidth,cfbox=green 1pt 1pt]{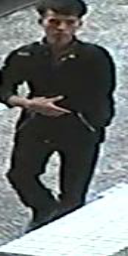}\\ 
      \includegraphics[width=0.95\linewidth,cfbox=green 1pt 1pt]{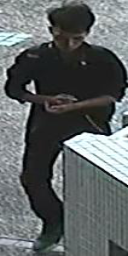}\\
      \includegraphics[width=0.95\linewidth,cfbox=green 1pt 1pt]{rank-sample/106.png}\\
      \vspace{-4.5mm}
      \caption*{\scriptsize Top 2}
    \end{subfigure}
    \hskip .5em
    \begin{subfigure}[b]{\elevenimg}
      \includegraphics[width=0.95\linewidth,cfbox=green 1pt 1pt]{rank-sample/106.png}\\
      \includegraphics[width=0.95\linewidth,cfbox=green 1pt 1pt]{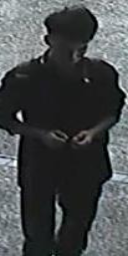}\\
      \includegraphics[width=0.95\linewidth,cfbox=green 1pt 1pt]{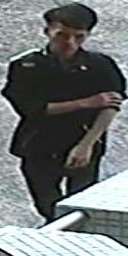}\\
      \vspace{-4.5mm}
      \caption*{\scriptsize Top 3}
    \end{subfigure}
    \hskip .5em
    \begin{subfigure}[b]{\elevenimg}
    \includegraphics[width=0.95\linewidth,cfbox=green 1pt 1pt]{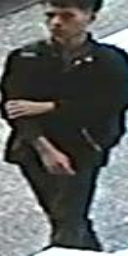}\\
      \includegraphics[width=0.95\linewidth,cfbox=green 1pt 1pt]{rank-sample/108.png}\\
      \includegraphics[width=0.95\linewidth,cfbox=green 1pt 1pt]{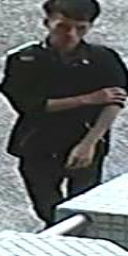}\\
      \vspace{-4.5mm}
      \caption*{\scriptsize Top 4}
    \end{subfigure}
    \hskip .5em
    \begin{subfigure}[b]{\elevenimg}
      \includegraphics[width=0.95\linewidth,cfbox=green 1pt 1pt]{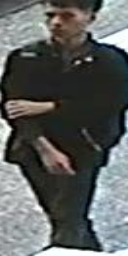}\\
      \includegraphics[width=0.95\linewidth,cfbox=red 1pt 1pt]{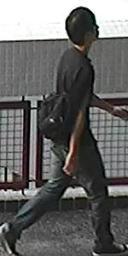}\\
      \includegraphics[width=0.95\linewidth,cfbox=green 1pt 1pt]{rank-sample/109.png}\\
      \vspace{-4.5mm}
      \caption*{\scriptsize Top 5}
    \end{subfigure}
    \hskip .5em
    \begin{subfigure}[b]{\elevenimg}
      \includegraphics[width=0.95\linewidth,cfbox=green 1pt 1pt]{rank-sample/111.png}\\
      \includegraphics[width=0.95\linewidth,cfbox=green 1pt 1pt]{rank-sample/113.png}\\
      \includegraphics[width=0.95\linewidth,cfbox=green 1pt 1pt]{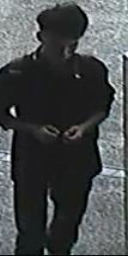}\\
      \vspace{-4.5mm}
      \caption*{\scriptsize Top 6}
    \end{subfigure}
    \hskip .5em
    \begin{subfigure}[b]{\elevenimg}
      \includegraphics[width=0.95\linewidth,cfbox=red 1pt 1pt]{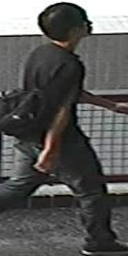}\\
      \includegraphics[width=0.95\linewidth,cfbox=green 1pt 1pt]{rank-sample/107.png}\\
      \includegraphics[width=0.95\linewidth,cfbox=green 1pt 1pt]{rank-sample/107.png}\\
      \vspace{-4.5mm}
      \caption*{\scriptsize Top 7}
    \end{subfigure}
    \caption{Down-sampling rate $r = 4$ (seen resolution).}
    \label{fig:rank-person-4}
  \end{subfigure}
  \hfill
  \begin{subfigure}{\twoimg}
    \begin{subfigure}[b]{\dimexpr\elevenimg+20pt\relax}
      \makebox[20pt]{\raisebox{15pt}{\rotatebox[origin=c]{90}{\cite{jiao2018deep}}}}%
      \includegraphics[width=0.95\elevenimg,cfbox=white 1pt 1pt]{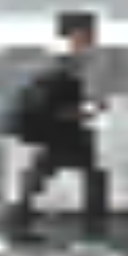}\\
      \makebox[20pt]{\raisebox{15pt}{\rotatebox[origin=c]{90}{\cite{wang2018cascaded}}}}%
      \includegraphics[width=0.95\elevenimg,cfbox=white 1pt 1pt]{rank-sample/Query_100_8.png}\\
      \makebox[20pt]{\raisebox{15pt}{\rotatebox[origin=c]{90}{Ours}}}%
      \includegraphics[width=0.95\elevenimg,cfbox=white 1pt 1pt]{rank-sample/Query_100_8.png}\\
      \vspace{-4.5mm}
      \caption*{\hspace{7mm} \scriptsize Query}
    \end{subfigure}
    \hskip .25em
    \tikz{\draw[-,blue, densely dashed, thick](0,-1.8) -- (0,3.25);}
    \hskip .25em
    \begin{subfigure}[b]{\elevenimg}
      \includegraphics[width=0.95\linewidth,cfbox=red 1pt 1pt]{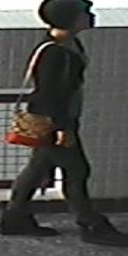}\\
      \includegraphics[width=0.95\linewidth,cfbox=red 1pt 1pt]{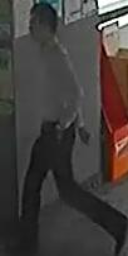}\\
      \includegraphics[width=0.95\linewidth,cfbox=green 1pt 1pt]{rank-sample/111.png}\\
      \vspace{-4.5mm}
      \caption*{\scriptsize Top 1}
    \end{subfigure}
    \hskip .5em
    \begin{subfigure}[b]{\elevenimg}
      \includegraphics[width=0.95\linewidth,cfbox=red 1pt 1pt]{rank-sample/618.png}\\ 
      \includegraphics[width=0.95\linewidth,cfbox=red 1pt 1pt]{rank-sample/623.png}\\
      \includegraphics[width=0.95\linewidth,cfbox=green 1pt 1pt]{rank-sample/106.png}\\
      \vspace{-4.5mm}
      \caption*{\scriptsize Top 2}
    \end{subfigure}
    \hskip .5em
    \begin{subfigure}[b]{\elevenimg}
      \includegraphics[width=0.95\linewidth,cfbox=red 1pt 1pt]{rank-sample/623.png}\\
      \includegraphics[width=0.95\linewidth,cfbox=red 1pt 1pt]{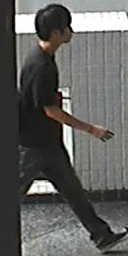}\\
      \includegraphics[width=0.95\linewidth,cfbox=green 1pt 1pt]{rank-sample/109.png}\\
      \vspace{-4.5mm}
      \caption*{\scriptsize Top 3}
    \end{subfigure}
    \hskip .5em
    \begin{subfigure}[b]{\elevenimg}
    \includegraphics[width=0.95\linewidth,cfbox=red 1pt 1pt]{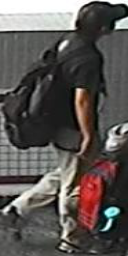}\\
      \includegraphics[width=0.95\linewidth,cfbox=red 1pt 1pt]{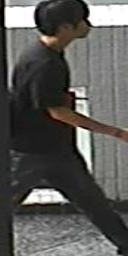}\\
      \includegraphics[width=0.95\linewidth,cfbox=green 1pt 1pt]{rank-sample/108.png}\\
      \vspace{-4.5mm}
      \caption*{\scriptsize Top 4}
    \end{subfigure}
    \hskip .5em
    \begin{subfigure}[b]{\elevenimg}
      \includegraphics[width=0.95\linewidth,cfbox=red 1pt 1pt]{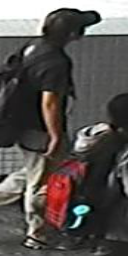}\\
      \includegraphics[width=0.95\linewidth,cfbox=red 1pt 1pt]{rank-sample/618.png}\\
      \includegraphics[width=0.95\linewidth,cfbox=red 1pt 1pt]{rank-sample/618.png}\\
      \vspace{-4.5mm}
      \caption*{\scriptsize Top 5}
    \end{subfigure}
    \hskip .5em
    \begin{subfigure}[b]{\elevenimg}
      \includegraphics[width=0.95\linewidth,cfbox=red 1pt 1pt]{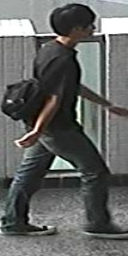}\\
      \includegraphics[width=0.95\linewidth,cfbox=red 1pt 1pt]{rank-sample/626.png}\\
      \includegraphics[width=0.95\linewidth,cfbox=green 1pt 1pt]{rank-sample/113.png}\\
      \vspace{-4.5mm}
      \caption*{\scriptsize Top 6}
    \end{subfigure}
    \hskip .5em
    \begin{subfigure}[b]{\elevenimg}
      \includegraphics[width=0.95\linewidth,cfbox=red 1pt 1pt]{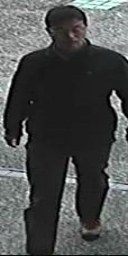}\\
      \includegraphics[width=0.95\linewidth,cfbox=green 1pt 1pt]{rank-sample/108.png}\\
      \includegraphics[width=0.95\linewidth,cfbox=green 1pt 1pt]{rank-sample/107.png}\\
      \vspace{-4.5mm}
      \caption*{\scriptsize Top 7}
    \end{subfigure}
    \caption{Down-sampling rate $r = 8$ (unseen resolution).}
    \label{fig:rank-person-8}
  \end{subfigure}
  \vspace{-1.5mm}
  \caption{\revised{\textbf{Top-ranked gallery images of cross-resolution person re-ID.} Given an LR query image with a down-sampling rate $r = 4$ (\emph{left}) or $r = 8$ (\emph{right}) of the MLR-CUHK03 test set, we present the top 7 gallery images retrieved by SING~\cite{jiao2018deep} (\emph{top row}), CSR-GAN~\cite{wang2018cascaded} (\emph{middle row}), and our method (\emph{bottom row}). Images enclosed by green and red rectangles denote correct and incorrect matches, respectively. Note that images with the down-sampling rate $r = 8$ are not seen during training.}}
  \label{fig:rank-person}
  \vspace{-2.0mm}
\end{figure*}

\setlength{\elevenimg}{0.04\textwidth}
\setlength{\twoimg}{0.495\textwidth}
\begin{figure*}[t]
  \ra{1.3}
  \centering
  \begin{subfigure}{\twoimg}
    \begin{subfigure}[b]{\dimexpr\elevenimg+20pt\relax}
      \makebox[20pt]{\raisebox{15pt}{\rotatebox[origin=c]{90}{\cite{jiao2018deep}}}}%
      \includegraphics[width=0.95\elevenimg,cfbox=white 1pt 1pt]{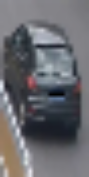}\\
      \makebox[20pt]{\raisebox{15pt}{\rotatebox[origin=c]{90}{\cite{wang2018cascaded}}}}%
      \includegraphics[width=0.95\elevenimg,cfbox=white 1pt 1pt]{rank-sample/63_q_4.png}\\
      \makebox[20pt]{\raisebox{15pt}{\rotatebox[origin=c]{90}{Ours}}}%
      \includegraphics[width=0.95\elevenimg,cfbox=white 1pt 1pt]{rank-sample/63_q_4.png}\\
      \vspace{-4.5mm}
      \caption*{\hspace{7mm} \scriptsize Query}
    \end{subfigure}
    \hskip .25em
    \tikz{\draw[-,blue, densely dashed, thick](0,-1.8) -- (0,3.25);}
    \hskip .25em
    \begin{subfigure}[b]{\elevenimg}
      \includegraphics[width=0.95\linewidth,cfbox=green 1pt 1pt]{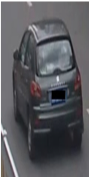}\\
      \includegraphics[width=0.95\linewidth,cfbox=green 1pt 1pt]{rank-sample/63_1.png}\\
      \includegraphics[width=0.95\linewidth,cfbox=green 1pt 1pt]{rank-sample/63_1.png}\\
      \vspace{-4.5mm}
      \caption*{\scriptsize Top 1}
    \end{subfigure}
    \hskip .5em
    \begin{subfigure}[b]{\elevenimg}
      \includegraphics[width=0.95\linewidth,cfbox=green 1pt 1pt]{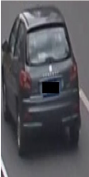}\\ 
      \includegraphics[width=0.95\linewidth,cfbox=green 1pt 1pt]{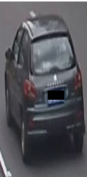}\\
      \includegraphics[width=0.95\linewidth,cfbox=green 1pt 1pt]{rank-sample/63_2.png}\\
      \vspace{-4.5mm}
      \caption*{\scriptsize Top 2}
    \end{subfigure}
    \hskip .5em
    \begin{subfigure}[b]{\elevenimg}
      \includegraphics[width=0.95\linewidth,cfbox=green 1pt 1pt]{rank-sample/63_3.png}\\
      \includegraphics[width=0.95\linewidth,cfbox=green 1pt 1pt]{rank-sample/63_2.png}\\
      \includegraphics[width=0.95\linewidth,cfbox=green 1pt 1pt]{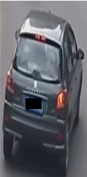}\\
      \vspace{-4.5mm}
      \caption*{\scriptsize Top 3}
    \end{subfigure}
    \hskip .5em
    \begin{subfigure}[b]{\elevenimg}
    \includegraphics[width=0.95\linewidth,cfbox=green 1pt 1pt]{rank-sample/63_4.png}\\
      \includegraphics[width=0.95\linewidth,cfbox=green 1pt 1pt]{rank-sample/63_4.png}\\
      \includegraphics[width=0.95\linewidth,cfbox=green 1pt 1pt]{rank-sample/63_3.png}\\
      \vspace{-4.5mm}
      \caption*{\scriptsize Top 4}
    \end{subfigure}
    \hskip .5em
    \begin{subfigure}[b]{\elevenimg}
      \includegraphics[width=0.95\linewidth,cfbox=red 1pt 1pt]{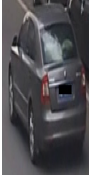}\\
      \includegraphics[width=0.95\linewidth,cfbox=green 1pt 1pt]{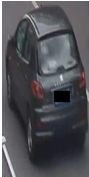}\\
      \includegraphics[width=0.95\linewidth,cfbox=green 1pt 1pt]{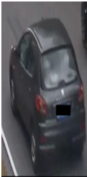}\\
      \vspace{-4.5mm}
      \caption*{\scriptsize Top 5}
    \end{subfigure}
    \hskip .5em
    \begin{subfigure}[b]{\elevenimg}
      \includegraphics[width=0.95\linewidth,cfbox=green 1pt 1pt]{rank-sample/63_5.png}\\
      \includegraphics[width=0.95\linewidth,cfbox=green 1pt 1pt]{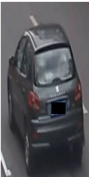}\\
      \includegraphics[width=0.95\linewidth,cfbox=green 1pt 1pt]{rank-sample/63_6.png}\\
      \vspace{-4.5mm}
      \caption*{\scriptsize Top 6}
    \end{subfigure}
    \hskip .5em
    \begin{subfigure}[b]{\elevenimg}
      \includegraphics[width=0.95\linewidth,cfbox=red 1pt 1pt]{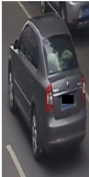}\\
      \includegraphics[width=0.95\linewidth,cfbox=red 1pt 1pt]{rank-sample/518_1.png}\\
      \includegraphics[width=0.95\linewidth,cfbox=red 1pt 1pt]{rank-sample/518_1.png}\\
      \vspace{-4.5mm}
      \caption*{\scriptsize Top 7}
    \end{subfigure}
    \caption{Down-sampling rate $r = 4$ (seen resolution).}
    \label{fig:rank-vehicle-4}
  \end{subfigure}
  \hfill
  \begin{subfigure}{\twoimg}
    \begin{subfigure}[b]{\dimexpr\elevenimg+20pt\relax}
      \makebox[20pt]{\raisebox{15pt}{\rotatebox[origin=c]{90}{\cite{jiao2018deep}}}}%
      \includegraphics[width=0.95\elevenimg,cfbox=white 1pt 1pt]{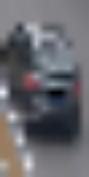}\\
      \makebox[20pt]{\raisebox{15pt}{\rotatebox[origin=c]{90}{\cite{wang2018cascaded}}}}%
      \includegraphics[width=0.95\elevenimg,cfbox=white 1pt 1pt]{rank-sample/63_q_8.png}\\
      \makebox[20pt]{\raisebox{15pt}{\rotatebox[origin=c]{90}{Ours}}}%
      \includegraphics[width=0.95\elevenimg,cfbox=white 1pt 1pt]{rank-sample/63_q_8.png}\\
      \vspace{-4.5mm}
      \caption*{\hspace{7mm} \scriptsize Query}
    \end{subfigure}
    \hskip .25em
    \tikz{\draw[-,blue, densely dashed, thick](0,-1.8) -- (0,3.25);}
    \hskip .25em
    \begin{subfigure}[b]{\elevenimg}
      \includegraphics[width=0.95\linewidth,cfbox=red 1pt 1pt]{rank-sample/518_2.png}\\
      \includegraphics[width=0.95\linewidth,cfbox=red 1pt 1pt]{rank-sample/518_2.png}\\
      \includegraphics[width=0.95\linewidth,cfbox=green 1pt 1pt]{rank-sample/63_1.png}\\
      \vspace{-4.5mm}
      \caption*{\scriptsize Top 1}
    \end{subfigure}
    \hskip .5em
    \begin{subfigure}[b]{\elevenimg}
      \includegraphics[width=0.95\linewidth,cfbox=red 1pt 1pt]{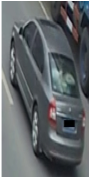}\\ 
      \includegraphics[width=0.95\linewidth,cfbox=green 1pt 1pt]{rank-sample/63_3.png}\\
      \includegraphics[width=0.95\linewidth,cfbox=green 1pt 1pt]{rank-sample/63_2.png}\\
      \vspace{-4.5mm}
      \caption*{\scriptsize Top 2}
    \end{subfigure}
    \hskip .5em
    \begin{subfigure}[b]{\elevenimg}
      \includegraphics[width=0.95\linewidth,cfbox=red 1pt 1pt]{rank-sample/63_3.png}\\
      \includegraphics[width=0.95\linewidth,cfbox=red 1pt 1pt]{rank-sample/518_3.png}\\
      \includegraphics[width=0.95\linewidth,cfbox=green 1pt 1pt]{rank-sample/63_4.png}\\
      \vspace{-4.5mm}
      \caption*{\scriptsize Top 3}
    \end{subfigure}
    \hskip .5em
    \begin{subfigure}[b]{\elevenimg}
    \includegraphics[width=0.95\linewidth,cfbox=green 1pt 1pt]{rank-sample/63_4.png}\\
      \includegraphics[width=0.95\linewidth,cfbox=green 1pt 1pt]{rank-sample/63_4.png}\\
      \includegraphics[width=0.95\linewidth,cfbox=green 1pt 1pt]{rank-sample/63_3.png}\\
      \vspace{-4.5mm}
      \caption*{\scriptsize Top 4}
    \end{subfigure}
    \hskip .5em
    \begin{subfigure}[b]{\elevenimg}
      \includegraphics[width=0.95\linewidth,cfbox=red 1pt 1pt]{rank-sample/518_2.png}\\
      \includegraphics[width=0.95\linewidth,cfbox=green 1pt 1pt]{rank-sample/63_6.png}\\
      \includegraphics[width=0.95\linewidth,cfbox=green 1pt 1pt]{rank-sample/63_5.png}\\
      \vspace{-4.5mm}
      \caption*{\scriptsize Top 5}
    \end{subfigure}
    \hskip .5em
    \begin{subfigure}[b]{\elevenimg}
      \includegraphics[width=0.95\linewidth,cfbox=green 1pt 1pt]{rank-sample/63_5.png}\\
      \includegraphics[width=0.95\linewidth,cfbox=red 1pt 1pt]{rank-sample/518_3.png}\\
      \includegraphics[width=0.95\linewidth,cfbox=green 1pt 1pt]{rank-sample/63_6.png}\\
      \vspace{-4.5mm}
      \caption*{\scriptsize Top 6}
    \end{subfigure}
    \hskip .5em
    \begin{subfigure}[b]{\elevenimg}
      \includegraphics[width=0.95\linewidth,cfbox=red 1pt 1pt]{rank-sample/518_1.png}\\
      \includegraphics[width=0.95\linewidth,cfbox=red 1pt 1pt]{rank-sample/518_1.png}\\
      \includegraphics[width=0.95\linewidth,cfbox=red 1pt 1pt]{rank-sample/518_1.png}\\
      \vspace{-4.5mm}
      \caption*{\scriptsize Top 7}
    \end{subfigure}
    \caption{Down-sampling rate $r = 8$ (unseen resolution).}
    \label{fig:rank-vehicle-8}
  \end{subfigure}
  \vspace{-1.5mm}
  \caption{\revised{\textbf{Top-ranked gallery images of cross-resolution vehicle re-ID.} Given an LR query image with a down-sampling rate $r = 4$ (\emph{left}) or $r = 8$ (\emph{right}) of the MLR-VeRi776 test set, we present the top 7 gallery images retrieved by SING~\cite{jiao2018deep} (\emph{top row}), CSR-GAN~\cite{wang2018cascaded} (\emph{middle row}), and our method (\emph{bottom row}). Images enclosed by green and red rectangles denote correct and incorrect matches, respectively. Note that images with the down-sampling rate $r = 8$ are not seen during training.}}
  \label{fig:rank-vehicle}
  \vspace{-3.0mm}
\end{figure*}

\vspace{3.0mm}

\subsubsection{The Recovered HR Images}

Here, we present the recovered HR images generated by our method and the ablation methods on the MLR-CUHK03 \emph{test set} using the cross-resolution person re-ID setting.

We visualize two examples of the recovered HR images in Figure~\ref{fig:abl}. In each example, the recovered HR images with different input down-sampling rates $r = \{1, 2, 4, 8\}$ are shown. We observe that without applying the HR reconstruction loss $\mathcal{L}_\mathrm{rec}$ (Eq. (\ref{eq:rec})), our model is not able to recover high-quality HR images. Without the multi-scale feature-level adversarial loss $\mathcal{L}_\mathrm{adv}^{\mathcal{D}_{F}}$ (Eq. (\ref{eq:adv_multi})), our model does not learn resolution-invariant representations. Thus, our model cannot recover the HR details of images of an unseen resolution, e.g., $r = 8$. While our model can still reconstruct the feature maps to their HR images without the image-level adversarial loss $\mathcal{L}_\mathrm{adv}^{\mathcal{D}_{I}}$ (Eq. (\ref{eq:adv_loss_image})), the recovered HR images may not look perceptually realistic, especially for input images with the down-sampling rate $r = 8$.

The examples in Figure~\ref{fig:abl} demonstrate that the HR reconstruction loss $\mathcal{L}_\mathrm{rec}$ is essential to image recovery. The multi-scale feature-level adversarial loss $\mathcal{L}_\mathrm{adv}^{\mathcal{D}_{F}}$ enables our model to deal with images of unseen resolutions while the image-level adversarial loss $\mathcal{L}_\mathrm{adv}^{\mathcal{D}_{I}}$ encourages our model to recover perceptually realistic HR images.

\setlength{\threeimg}{0.65\linewidth}
\begin{figure*}[t]
  \centering
  \begin{subfigure}[b]{\threeimg}
    \centering
    \includegraphics[width=\linewidth]{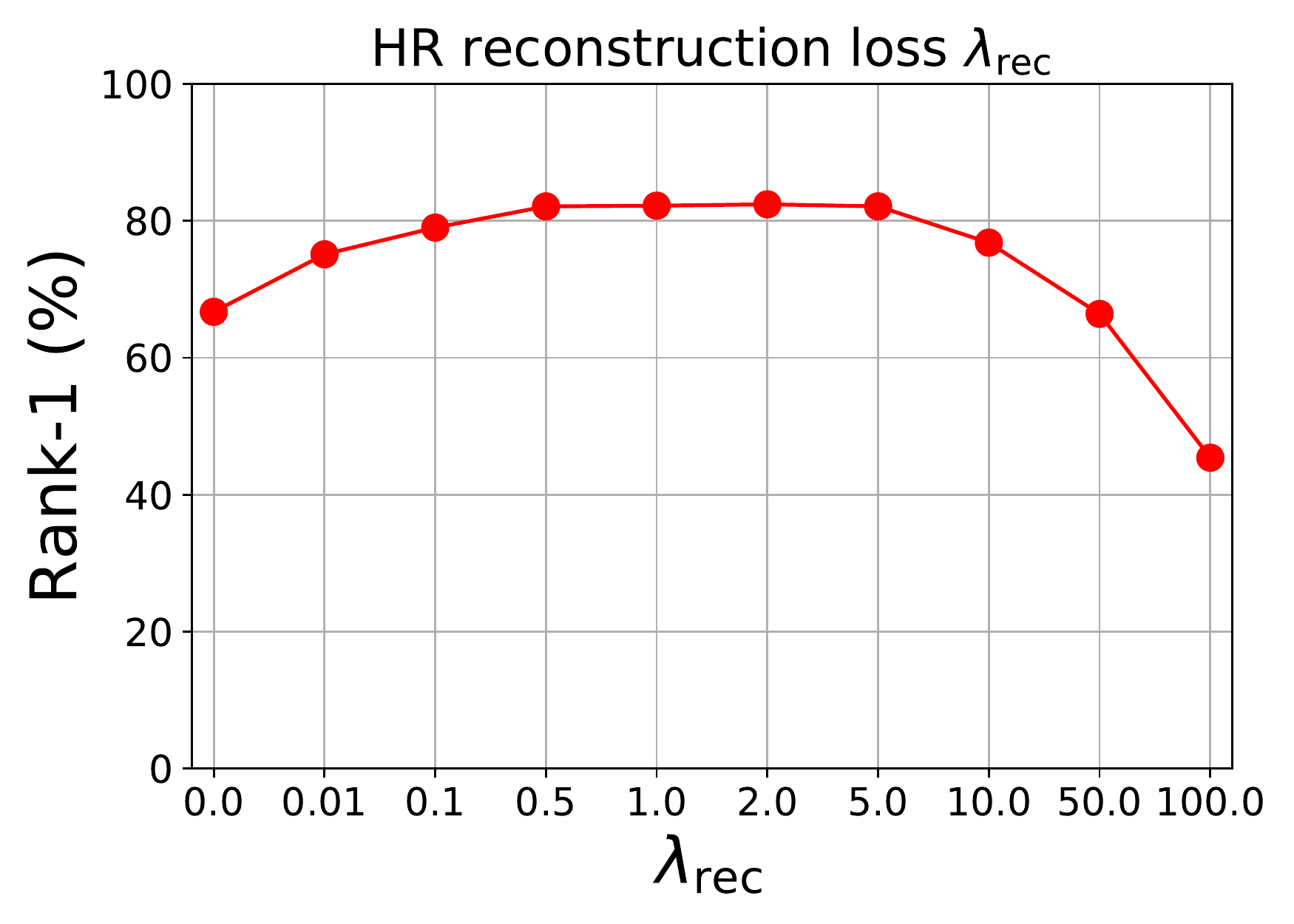}
    \vspace{-6.0mm}
    \caption{HR reconstruction loss $\lambda_\mathrm{rec}$.}
    \label{fig:lambda-rec}
  \end{subfigure}
  \hfill
  \begin{subfigure}[b]{\threeimg}
    \centering
    \includegraphics[width=\linewidth]{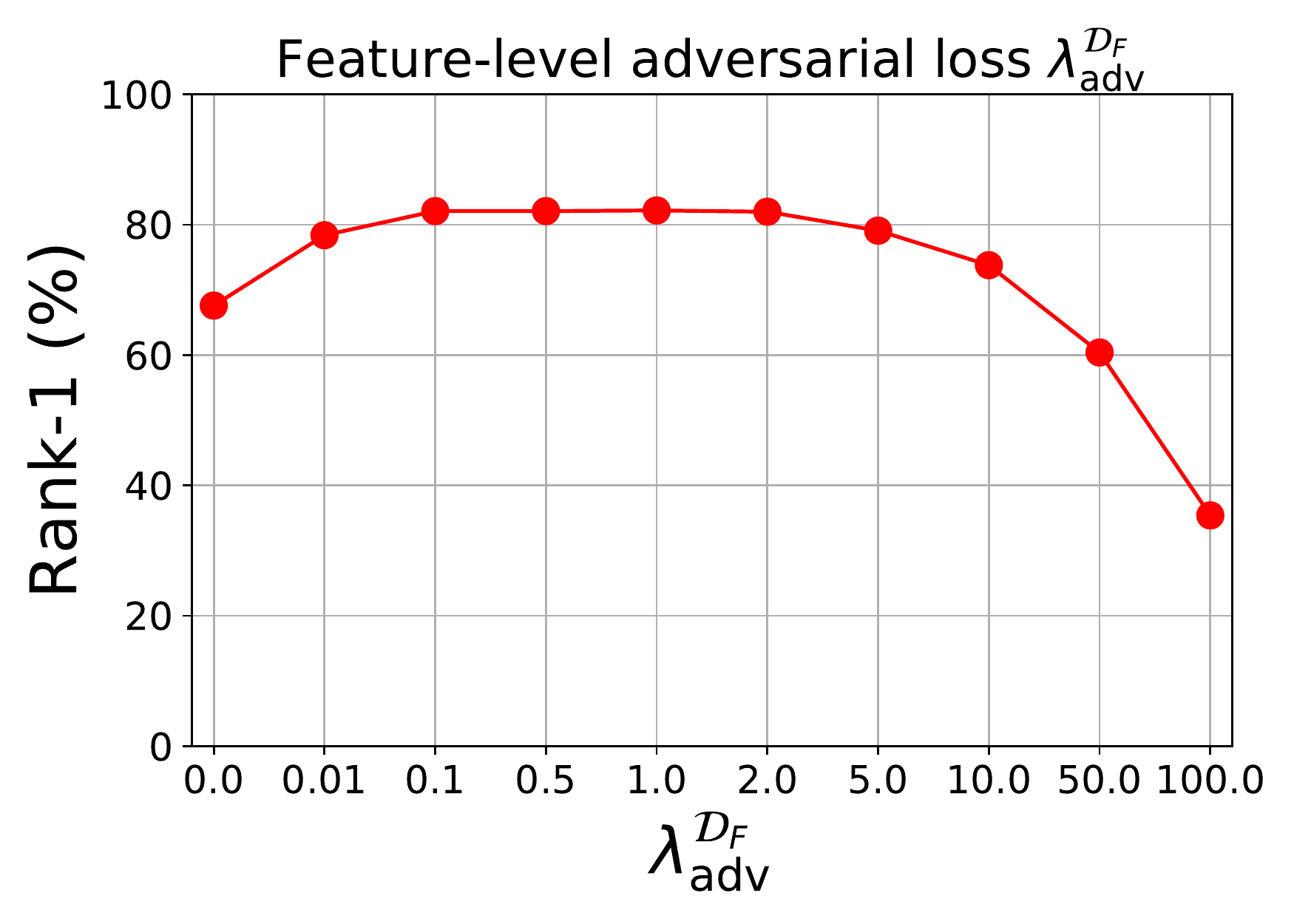}
    \vspace{-6.0mm}
    \caption{Feature-level adversarial loss $\lambda_\mathrm{adv}^{D_F}$.}
    \label{fig:lambda-adv-F}
  \end{subfigure}
  \hfill
  \begin{subfigure}[b]{\threeimg}
    \centering
    \includegraphics[width=\linewidth]{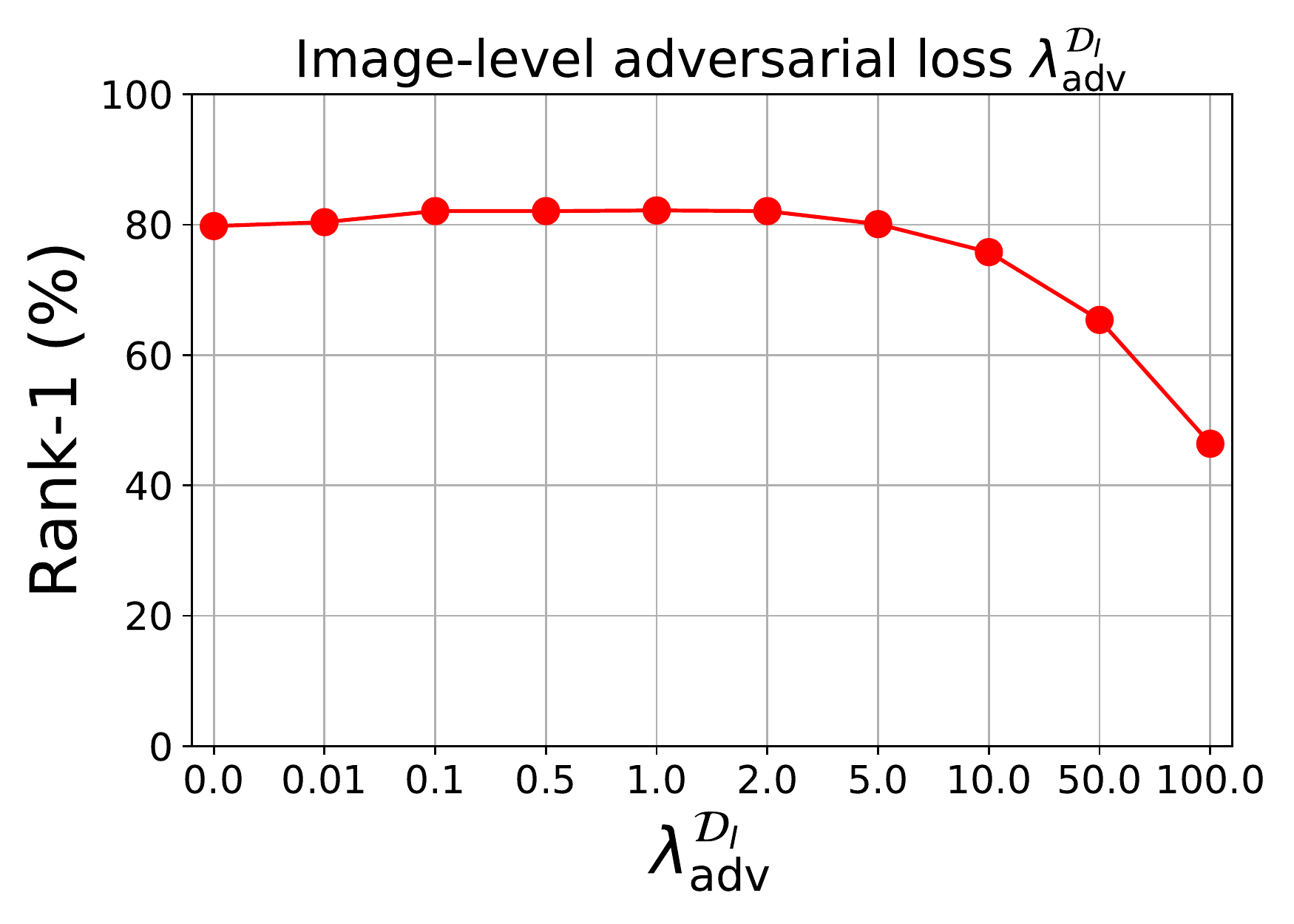}
    \vspace{-6.0mm}
    \caption{Image-level adversarial loss $\lambda_\mathrm{adv}^{D_I}$.}
    \label{fig:lambda-adv-I}
  \end{subfigure}
  \vspace{-1.0mm}
  \caption{\revised{\textbf{Sensitivity analysis of hyper-parameters.} We conduct a sensitivity analysis of hyper-parameters on the MLR-CUHK03 validation set using the cross-resolution person re-ID setting. We observe that the performance of our proposed method is generally stable when the hyper-parameters are set within a reasonable range.}}
  \label{fig:lambda}
  \vspace{-3.0mm}
\end{figure*}

\subsection{Resolution-Invariant Representation}

\subsubsection{Effect of Feature-level Adversarial Loss}

To demonstrate the effectiveness of our model in deriving the resolution-invariant representations $f$, we first apply global average pooling to $f$ to obtain the resolution-invariant feature vector $\mathbfit{w} = \mathrm{GAP}(f) \in \bbR^d$. We then visualize $\mathbfit{w}$ on the MLR-CUHK03 \emph{test set} in Figure~\ref{fig:tsne}.

We select $50$ different identities, each of which is indicated by a unique color, as shown in Figure~\ref{fig:tsne-baseline} and Figure~\ref{fig:tsne-identity}. In Figure~\ref{fig:tsne-baseline}, we observe that without the multi-scale feature-level adversarial loss $\mathcal{L}_\mathrm{adv}^{\mathcal{D}_F}$ (Eq. (\ref{eq:adv_multi})), our model cannot establish a well-separated feature space. When loss $\mathcal{L}_\mathrm{adv}^{\mathcal{D}_F}$ is imposed, the projected feature vectors are well separated as shown in Figure~\ref{fig:tsne-identity}. These two figures indicate that without loss $\mathcal{L}_\mathrm{adv}^{\mathcal{D}_F}$, our model does not learn resolution-invariant representations, thus implicitly suffering from the negative impact induced by the resolution mismatch issue.

We note that the projected feature vectors in Figure~\ref{fig:tsne-identity} are well separated, suggesting that sufficient re-ID ability can be exhibited by our model. On the other hand, for Figure~\ref{fig:tsne-resolution}, we colorize each image resolution with a unique color in each identity cluster (four different down-sampling rates $r \in \{1, 2, 4, 8\}$). We observe that the projected feature vectors of the same identity but different down-sampling rates are all well clustered. We note that images with down-sampling rate $r = 8$ are not presented in the training set (i.e., unseen resolution).

The above visualizations demonstrate that our model learns resolution-invariant representations and generalizes well to unseen image resolution (e.g., $r = 8$) for cross-resolution person re-ID.

\subsubsection{Visual Comparisons of the Re-ID Feature Vector}

We visualize the feature vector for person re-ID on the MLR-CUHK$03$ \emph{test set} via t-SNE and present visual comparisons with SING~\cite{jiao2018deep} and CSR-GAN~\cite{wang2018cascaded}. The comparisons are conducted on a subset of $50$ identities. Note that for our method, we use the joint feature vector $\mathbfit{u} = \mathrm{GAP}(\mathbfit{v}) \in \bbR^{2d}$ for person re-ID.

Figure~\ref{fig:tsne-identity-comp} shows the visual results of the three competing methods, including SING~\cite{jiao2018deep} in Figure~\ref{fig:tsne-identity-naive}, CSR-GAN~\cite{wang2018cascaded} in Figure~\ref{fig:tsne-identity-better}, and our method in Figure~\ref{fig:tsne-identity-ours}. In each figure, we plot the projected re-ID feature vectors of each identity with a specific color. We observe that both SING~\cite{jiao2018deep} and CSR-GAN~\cite{wang2018cascaded} do not separate instances of different identities very well. In contrast, as shown in Figure~\ref{fig:tsne-identity-ours}, our method successfully recognizes most identities, implying the reliable re-ID ability of our proposed method.

In Figure~\ref{fig:tsne-reso-comp}, we adopt resolution-specific coloring and display the visual results of SING~\cite{jiao2018deep} in Figure~\ref{fig:tsne-reso-naive}, CSR-GAN~\cite{wang2018cascaded} in Figure~\ref{fig:tsne-reso-better}, and our method in Figure~\ref{fig:tsne-reso-ours}. As shown in Figure~\ref{fig:tsne-reso-naive} and Figure~\ref{fig:tsne-reso-better}, SING~\cite{jiao2018deep} and CSR-GAN~\cite{wang2018cascaded} tend to mix images yielded with the down-sampling rate $r = 8$, i.e., those in red, even if these images are of different categories. The visual results indicate that both SING~\cite{jiao2018deep} and CSR-GAN~\cite{wang2018cascaded} suffer from the resolution mismatch problem. Our method, on the other hand, learns resolution-invariant representations. As shown in Figure~\ref{fig:tsne-reso-ours}, images even with \emph{unseen} different down-sampling rates (e.g., $r = 8$) are well clustered with respect to the identities.

The above visual comparisons verify that through learning resolution-invariant representations, our method works well on images of diverse and even unseen resolutions.

\subsection{Top-Ranked Gallery Images}

{\flushleft {\bf Person re-ID.}}
As shown in Figure~\ref{fig:rank-person}, given an LR query image with down-sampling rate $r = 4$ or $r = 8$, we present the first $7$ top-ranked HR gallery images in Figure~\ref{fig:rank-person-4} and Figure~\ref{fig:rank-person-8}, respectively. We compare our method (bottom row) with two approaches developed for cross-resolution person re-ID SING~\cite{jiao2018deep} (top row) and CSR-GAN~\cite{wang2018cascaded} (middle row). The green and red boundaries indicate correct and incorrect matches, respectively. In Figure~\ref{fig:rank-person-4}, all three approaches including ours achieves almost correct matches. However, from the results in the top row of Figure~\ref{fig:rank-person-8}, we observe that SING~\cite{jiao2018deep} does not have any correct matches, while CSR-GAN~\cite{wang2018cascaded} achieves $1$ out of $7$ correct matches. Our method, on the contrary, achieves $6$ out of $7$ correct matches, which again verifies the effectiveness and robustness of our model. Note that the resolution ($r = 8$) of the query image is not seen during training.

{\flushleft {\bf Vehicle re-ID.}}
Similarly, as shown in Figure~\ref{fig:rank-vehicle}, given an LR query image with down-sampling rates $r = 4$ and $r = 8$, we present the first $7$ top-ranked HR gallery images in Figure~\ref{fig:rank-vehicle-4} and Figure~\ref{fig:rank-vehicle-8}, respectively. We also compare our method (bottom row) with two existing methods, i.e, SING~\cite{jiao2018deep} (top row) and CSR-GAN~\cite{wang2018cascaded} (middle row). In Figure~\ref{fig:rank-vehicle-4}, all three approaches achieve satisfactory re-ID results. We then consider the case where the resolution ($r = 8$) of the query image is unseen during training. In Figure~\ref{fig:rank-vehicle-8}, SING~\cite{jiao2018deep} and CSR-GAN~\cite{wang2018cascaded} only have $2$ and $3$ out of $7$ correct matches, respectively. In contrast, our method achieves $6$ out of $7$ correct matches. The comparison with existing methods also supports the applicability of our method to cross-resolution vehicle re-ID.

\subsection{Sensitivity Analysis of Hyper-parameters}

We further analyze the sensitivity of our model against the hyper-parameters introduced in the total loss $\mathcal{L}$ (Eq. (\ref{eq:fullobj})). We conduct sensitivity analysis by varying the value of each hyper-parameter and report the results on the MLR-CUHK$03$ \emph{validation set} using the cross-resolution person re-ID setting. Figure~\ref{fig:lambda} presents the experimental results.

For $\lambda_\mathrm{rec}$ and $\lambda_\mathrm{adv}^{\mathcal{D}_F}$, if their values are set to $0$, our model suffers from performance drops, as shown Figure~\ref{fig:lambda-rec} and Figure~\ref{fig:lambda-adv-F}. When $\lambda_\mathrm{rec}$ and $\lambda_\mathrm{adv}^{\mathcal{D}_F}$ lie in a certain range (near $1$), the performance of our method is improved and remains stable. However, once their values are too large, e.g., $100$, significant performance drop occurs since the corresponding losses (i.e., the HR reconstruction loss $\mathcal{L}_\mathrm{rec}$ and the multi-scale feature-level adversarial loss $\mathcal{L}_\mathrm{adv}^{\mathcal{D}_F}$) dominate the total loss. In Figure~\ref{fig:lambda-adv-I}, we observe that when the value of $\lambda_\mathrm{adv}^{\mathcal{D}_I}$ is set to $0$, modest performance drop happens. If $\lambda_\mathrm{adv}^{\mathcal{D}_I}$ varies within a reasonable region, e.g., $0.01\sim2$, the performance remains stable. However, when the value of $\lambda_\mathrm{adv}^{\mathcal{D}_I}$ is too large, e.g., larger than $10$, severe performance drop occurs since the image-level adversarial loss $\lambda_\mathrm{adv}^{\mathcal{D}_I}$ dominates the total loss.

In sum, the performance of our model remains stable when the values of the hyper-parameters lie in a certain range (near $1$). As a result, we set $\lambda_\mathrm{rec} = 1$, 
$\lambda_\mathrm{adv}^{\mathcal{D}_F} = 1$, and $\lambda_\mathrm{adv}^{\mathcal{D}_I} = 1$ without further fine-tuning them, though applying grid search or random search for hyper-parameter optimization might lead to further performance improvement.

\section{Conclusions}

We have presented an \emph{end-to-end trainable} generative adversarial network, CAD-Net++, for addressing the resolution mismatch issue in person re-ID. The core technical novelty lies in the unique design of the proposed CRGAN which learns the \emph{resolution-invariant} representations while being able to recover \emph{re-ID oriented} HR details preferable for person re-ID. Our cross-modal re-ID network jointly considers the information from two feature modalities, resulting in improved re-ID performance. \revised{Extensive experimental results show that our approach performs favorably against existing cross-resolution person re-ID methods on five challenging benchmarks, achieves competitive performance against existing approaches even when no significant resolution variations are present, and produces perceptually higher quality HR images using only a \emph{single} model. Visualization of the resolution-invariant representations further verifies our ability in handling query images with \emph{varying} or even \emph{unseen} resolutions. Furthermore, we demonstrate the applicability of our method through cross-resolution vehicle re-ID task. Experimental results confirm the generalization of our model on cross-resolution visual tasks. The extensions to semi-supervised settings also demonstrate the superiority of our method over existing approaches. Thus, the use of our model for practical re-ID applications can be strongly supported.}

%\ifCLASSOPTIONcompsoc
%  % The Computer Society usually uses the plural form
%\else
%  % regular IEEE prefers the singular form
%\fi


\ifCLASSOPTIONcaptionsoff
  \newpage
\fi

\bibliographystyle{IEEEtran}
\bibliography{IEEEabrv,reference.bib}

\begin{thebibliography}{10}
\providecommand{\url}[1]{#1}
\csname url@samestyle\endcsname
\providecommand{\newblock}{\relax}
\providecommand{\bibinfo}[2]{#2}
\providecommand{\BIBentrySTDinterwordspacing}{\spaceskip=0pt\relax}
\providecommand{\BIBentryALTinterwordstretchfactor}{4}
\providecommand{\BIBentryALTinterwordspacing}{\spaceskip=\fontdimen2\font plus
\BIBentryALTinterwordstretchfactor\fontdimen3\font minus
  \fontdimen4\font\relax}
\providecommand{\BIBforeignlanguage}[2]{{%
\expandafter\ifx\csname l@#1\endcsname\relax
\typeout{** WARNING: IEEEtran.bst: No hyphenation pattern has been}%
\typeout{** loaded for the language `#1'. Using the pattern for}%
\typeout{** the default language instead.}%
\else
\language=\csname l@#1\endcsname
\fi
#2}}
\providecommand{\BIBdecl}{\relax}
\BIBdecl

\bibitem{zheng2016person}
L.~Zheng, Y.~Yang, and A.~G. Hauptmann, ``Person re-identification: Past,
  present and future,'' \emph{arXiv}, 2016.

\bibitem{zhong2017re}
Z.~Zhong, L.~Zheng, D.~Cao, and S.~Li, ``Re-ranking person re-identification
  with k-reciprocal encoding,'' in \emph{CVPR}, 2017.

\bibitem{wang2015zero}
Z.~Wang, R.~Hu, C.~Liang, Y.~Yu, J.~Jiang, M.~Ye, J.~Chen, and Q.~Leng,
  ``Zero-shot person re-identification via cross-view consistency,''
  \emph{TMM}, 2015.

\bibitem{ye2020deep}
M.~Ye, J.~Shen, G.~Lin, T.~Xiang, L.~Shao, and S.~C. Hoi, ``Deep learning for
  person re-identification: A survey and outlook,'' \emph{arXiv}, 2020.

\bibitem{andriluka2008people}
M.~Andriluka, S.~Roth, and B.~Schiele, ``People-tracking-by-detection and
  people-detection-by-tracking,'' in \emph{CVPR}, 2008.

\bibitem{khan2016person}
F.~M. Khan and F.~Br{\'e}mond, ``Person re-identification for real-world
  surveillance systems,'' \emph{arXiv}, 2016.

\bibitem{garcia2015person}
J.~Garcia, N.~Martinel, C.~Micheloni, and A.~Gardel, ``Person re-identification
  ranking optimisation by discriminant context information analysis,'' in
  \emph{ICCV}, 2015.

\bibitem{vezzani2013people}
R.~Vezzani, D.~Baltieri, and R.~Cucchiara, ``People reidentification in
  surveillance and forensics: A survey,'' \emph{ACM Computing Surveys (CSUR)},
  2013.

\bibitem{lin2017improving}
Y.~Lin, L.~Zheng, Z.~Zheng, Y.~Wu, and Y.~Yang, ``Improving person
  re-identification by attribute and identity learning,'' \emph{arXiv}, 2017.

\bibitem{shen2018deep}
Y.~Shen, H.~Li, T.~Xiao, S.~Yi, D.~Chen, and X.~Wang, ``Deep group-shuffling
  random walk for person re-identification,'' in \emph{CVPR}, 2018.

\bibitem{hermans2017defense}
A.~Hermans, L.~Beyer, and B.~Leibe, ``In defense of the triplet loss for person
  re-identification,'' \emph{arXiv}, 2017.

\bibitem{zhong2017camera}
Z.~Zhong, L.~Zheng, Z.~Zheng, S.~Li, and Y.~Yang, ``Camera style adaptation for
  person re-identification,'' in \emph{CVPR}, 2018.

\bibitem{si2018dual}
J.~Si, H.~Zhang, C.-G. Li, J.~Kuen, X.~Kong, A.~C. Kot, and G.~Wang, ``Dual
  attention matching network for context-aware feature sequence based person
  re-identification,'' in \emph{CVPR}, 2018.

\bibitem{chen2018group}
D.~Chen, D.~Xu, H.~Li, N.~Sebe, and X.~Wang, ``Group consistent similarity
  learning via deep crf for person re-identification,'' in \emph{CVPR}, 2018.

\bibitem{zhang2019densely}
Z.~Zhang, C.~Lan, W.~Zeng, and Z.~Chen, ``Densely semantically aligned person
  re-identification,'' in \emph{CVPR}, 2019.

\bibitem{hou2019interaction}
R.~Hou, B.~Ma, H.~Chang, X.~Gu, S.~Shan, and X.~Chen,
  ``Interaction-and-aggregation network for person re-identification,'' in
  \emph{CVPR}, 2019.

\bibitem{zheng2019joint}
Z.~Zheng, X.~Yang, Z.~Yu, L.~Zheng, Y.~Yang, and J.~Kautz, ``Joint
  discriminative and generative learning for person re-identification,'' in
  \emph{CVPR}, 2019.

\bibitem{zheng2019re}
M.~Zheng, S.~Karanam, Z.~Wu, and R.~J. Radke, ``Re-identification with
  consistent attentive siamese networks,'' in \emph{CVPR}, 2019.

\bibitem{jing2015super}
X.-Y. Jing, X.~Zhu, F.~Wu, X.~You, Q.~Liu, D.~Yue, R.~Hu, and B.~Xu,
  ``Super-resolution person re-identification with semi-coupled low-rank
  discriminant dictionary learning,'' in \emph{CVPR}, 2015.

\bibitem{wang2016scale}
Z.~Wang, R.~Hu, Y.~Yu, J.~Jiang, C.~Liang, and J.~Wang, ``Scale-adaptive
  low-resolution person re-identification via learning a discriminating
  surface,'' in \emph{IJCAI}, 2016.

\bibitem{jiao2018deep}
J.~Jiao, W.-S. Zheng, A.~Wu, X.~Zhu, and S.~Gong, ``Deep low-resolution person
  re-identification,'' in \emph{AAAI}, 2018.

\bibitem{wang2018cascaded}
Z.~Wang, M.~Ye, F.~Yang, X.~Bai, and S.~Satoh, ``Cascaded sr-gan for
  scale-adaptive low resolution person re-identification.'' in \emph{IJCAI},
  2018.

\bibitem{li2015multi}
X.~Li, W.-S. Zheng, X.~Wang, T.~Xiang, and S.~Gong, ``Multi-scale learning for
  low-resolution person re-identification,'' in \emph{ICCV}, 2015.

\bibitem{tsai2018learning}
Y.-H. Tsai, W.-C. Hung, S.~Schulter, K.~Sohn, M.-H. Yang, and M.~Chandraker,
  ``Learning to adapt structured output space for semantic segmentation,'' in
  \emph{CVPR}, 2018.

\bibitem{chen2019learning}
Y.-C. Chen, Y.-J. Li, X.~Du, and Y.-C.~F. Wang, ``Learning resolution-invariant
  deep representations for person re-identification,'' in \emph{AAAI}, 2019.

\bibitem{kanaci2018vehicle}
A.~Kanac{\i}, X.~Zhu, and S.~Gong, ``Vehicle re-identification in context,'' in
  \emph{GCPR}, 2018.

\bibitem{CAD-Net}
Y.-J. Li, Y.-C. Chen, Y.-Y. Lin, X.~Du, and Y.-C.~F. Wang, ``Recover and
  identify: A generative dual model for cross-resolution person
  re-identification,'' in \emph{ICCV}, 2019.

\bibitem{shen2018person}
Y.~Shen, H.~Li, S.~Yi, D.~Chen, and X.~Wang, ``Person re-identification with
  deep similarity-guided graph neural network,'' in \emph{ECCV}, 2018.

\bibitem{kalayeh2018human}
M.~M. Kalayeh, E.~Basaran, M.~G{\"o}kmen, M.~E. Kamasak, and M.~Shah, ``Human
  semantic parsing for person re-identification,'' in \emph{CVPR}, 2018.

\bibitem{cheng2016person}
D.~Cheng, Y.~Gong, S.~Zhou, J.~Wang, and N.~Zheng, ``Person re-identification
  by multi-channel parts-based cnn with improved triplet loss function,'' in
  \emph{CVPR}, 2016.

\bibitem{chang2018multi}
X.~Chang, T.~M. Hospedales, and T.~Xiang, ``Multi-level factorisation net for
  person re-identification,'' in \emph{CVPR}, 2018.

\bibitem{li2018adaptation}
Y.-J. Li, F.-E. Yang, Y.-C. Liu, Y.-Y. Yeh, X.~Du, and Y.-C. Frank~Wang,
  ``Adaptation and re-identification network: An unsupervised deep transfer
  learning approach to person re-identification,'' in \emph{CVPRW}, 2018.

\bibitem{sun2018beyond}
Y.~Sun, L.~Zheng, Y.~Yang, Q.~Tian, and S.~Wang, ``Beyond part models: Person
  retrieval with refined part pooling (and a strong convolutional baseline),''
  in \emph{ECCV}, 2018.

\bibitem{suh2018part}
Y.~Suh, J.~Wang, S.~Tang, T.~Mei, and K.~Mu~Lee, ``Part-aligned bilinear
  representations for person re-identification,'' in \emph{ECCV}, 2018.

\bibitem{liu2018pose}
J.~Liu, B.~Ni, Y.~Yan, P.~Zhou, S.~Cheng, and J.~Hu, ``Pose transferrable
  person re-identification,'' in \emph{CVPR}, 2018.

\bibitem{goodfellow2014generative}
I.~Goodfellow, J.~Pouget-Abadie, M.~Mirza, B.~Xu, D.~Warde-Farley, S.~Ozair,
  A.~Courville, and Y.~Bengio, ``Generative adversarial nets,'' in
  \emph{NeurIPS}, 2014.

\bibitem{li2018harmonious}
W.~Li, X.~Zhu, and S.~Gong, ``Harmonious attention network for person
  re-identification,'' in \emph{CVPR}, 2018.

\bibitem{song2018mask}
C.~Song, Y.~Huang, W.~Ouyang, and L.~Wang, ``Mask-guided contrastive attention
  model for person re-identification,'' in \emph{CVPR}, 2018.

\bibitem{li2017person}
S.~Li, M.~Shao, and Y.~Fu, ``Person re-identification by cross-view multi-level
  dictionary learning,'' \emph{TPAMI}, 2017.

\bibitem{ganin2015unsupervised}
Y.~Ganin and V.~Lempitsky, ``Unsupervised domain adaptation by
  backpropagation,'' in \emph{ICML}, 2015.

\bibitem{ganin2016domain}
Y.~Ganin, E.~Ustinova, H.~Ajakan, P.~Germain, H.~Larochelle, F.~Laviolette,
  M.~Marchand, and V.~Lempitsky, ``Domain-adversarial training of neural
  networks,'' \emph{JMLR}, 2016.

\bibitem{long2015learning}
M.~Long, Y.~Cao, J.~Wang, and M.~I. Jordan, ``Learning transferable features
  with deep adaptation networks,'' in \emph{ICML}, 2015.

\bibitem{long2016unsupervised}
M.~Long, H.~Zhu, J.~Wang, and M.~I. Jordan, ``Unsupervised domain adaptation
  with residual transfer networks,'' in \emph{NeurIPS}, 2016.

\bibitem{chen2019crdoco}
Y.-C. Chen, Y.-Y. Lin, M.-H. Yang, and J.-B. Huang, ``Crdoco: Pixel-level
  domain transfer with cross-domain consistency,'' in \emph{CVPR}, 2019.

\bibitem{hoffman2017cycada}
J.~Hoffman, E.~Tzeng, T.~Park, J.-Y. Zhu, P.~Isola, K.~Saenko, A.~A. Efros, and
  T.~Darrell, ``Cycada: Cycle-consistent adversarial domain adaptation,'' in
  \emph{ICML}, 2018.

\bibitem{wei2018person}
L.~Wei, S.~Zhang, W.~Gao, and Q.~Tian, ``Person transfer gan to bridge domain
  gap for person re-identification,'' in \emph{CVPR}, 2018.

\bibitem{image-image18}
W.~Deng, L.~Zheng, Q.~Ye, G.~Kang, Y.~Yang, and J.~Jiao, ``Image-image domain
  adaptation with preserved self-similarity and domain-dissimilarity for person
  reidentification,'' in \emph{CVPR}, 2018.

\bibitem{ge2018fd}
Y.~Ge, Z.~Li, H.~Zhao, G.~Yin, S.~Yi, X.~Wang, and H.~Li, ``Fd-gan: Pose-guided
  feature distilling gan for robust person re-identification,'' in
  \emph{NeurIPS}, 2018.

\bibitem{li2019cross}
Y.-J. Li, C.-S. Lin, Y.-B. Lin, and Y.-C.~F. Wang, ``Cross-dataset person
  re-identification via unsupervised pose disentanglement and adaptation,'' in
  \emph{ICCV}, 2019.

\bibitem{zhu2017unpaired}
J.-Y. Zhu, T.~Park, P.~Isola, and A.~A. Efros, ``Unpaired image-to-image
  translation using cycle-consistent adversarial networks,'' in \emph{ICCV},
  2017.

\bibitem{mao2019resolution}
S.~Mao, S.~Zhang, and M.~Yang, ``Resolution-invariant person
  re-identification,'' in \emph{IJCAI}, 2019.

\bibitem{li2018toward}
K.~Li, Z.~Ding, S.~Li, and Y.~Fu, ``Toward resolution-invariant person
  reidentification via projective dictionary learning,'' \emph{TNNLS}, 2018.

\bibitem{HoG}
N.~Dalal and B.~Triggs, ``Histograms of oriented gradients for human
  detection,'' in \emph{CVPR}, 2005.

\bibitem{ledig2017photo}
C.~Ledig, L.~Theis, F.~Husz{\'a}r, J.~Caballero, A.~Cunningham, A.~Acosta,
  A.~P. Aitken, A.~Tejani, J.~Totz, Z.~Wang \emph{et~al.}, ``Photo-realistic
  single image super-resolution using a generative adversarial network.'' in
  \emph{CVPR}, 2017.

\bibitem{zhu2016deep}
S.~Zhu, S.~Liu, C.~C. Loy, and X.~Tang, ``Deep cascaded bi-network for face
  hallucination,'' in \emph{ECCV}, 2016.

\bibitem{yu2017hallucinating}
X.~Yu and F.~Porikli, ``Hallucinating very low-resolution unaligned and noisy
  face images by transformative discriminative autoencoders,'' in \emph{CVPR},
  2017.

\bibitem{kim2016accurate}
J.~Kim, J.~Kwon~Lee, and K.~Mu~Lee, ``Accurate image super-resolution using
  very deep convolutional networks,'' in \emph{CVPR}, 2016.

\bibitem{dahl2017pixel}
R.~Dahl, M.~Norouzi, and J.~Shlens, ``Pixel recursive super resolution,'' in
  \emph{ICCV}, 2017.

\bibitem{dong2016image}
C.~Dong, C.~C. Loy, K.~He, and X.~Tang, ``Image super-resolution using deep
  convolutional networks,'' \emph{TPAMI}, 2016.

\bibitem{zhou2018aware}
Y.~Zhou and L.~Shao, ``Aware attentive multi-view inference for vehicle
  re-identification,'' in \emph{CVPR}, 2018.

\bibitem{wang2017orientation}
Z.~Wang, L.~Tang, X.~Liu, Z.~Yao, S.~Yi, J.~Shao, J.~Yan, S.~Wang, H.~Li, and
  X.~Wang, ``Orientation invariant feature embedding and spatial temporal
  regularization for vehicle re-identification,'' in \emph{ICCV}, 2017.

\bibitem{shen2017learning}
Y.~Shen, T.~Xiao, H.~Li, S.~Yi, and X.~Wang, ``Learning deep neural networks
  for vehicle re-id with visual-spatio-temporal path proposals,'' in
  \emph{ICCV}, 2017.

\bibitem{he2016deep}
K.~He, X.~Zhang, S.~Ren, and J.~Sun, ``Deep residual learning for image
  recognition,'' in \emph{CVPR}, 2016.

\bibitem{huang2018munit}
X.~Huang, M.-Y. Liu, S.~Belongie, and J.~Kautz, ``Multimodal unsupervised
  image-to-image translation,'' in \emph{ECCV}, 2018.

\bibitem{miyato2018cgans}
T.~Miyato and M.~Koyama, ``cgans with projection discriminator,'' in
  \emph{ICLR}, 2018.

\bibitem{li2014deepreid}
W.~Li, R.~Zhao, T.~Xiao, and X.~Wang, ``Deepreid: Deep filter pairing neural
  network for person re-identification,'' in \emph{CVPR}, 2014.

\bibitem{gray2008viewpoint}
D.~Gray and H.~Tao, ``Viewpoint invariant pedestrian recognition with an
  ensemble of localized features,'' in \emph{ECCV}, 2008.

\bibitem{Cheng:BMVC11}
D.~S. Cheng, M.~Cristani, M.~Stoppa, L.~Bazzani, and V.~Murino, ``Custom
  pictorial structures for re-identification,'' in \emph{BMVC}, 2011.

\bibitem{zheng2015scalable}
L.~Zheng, L.~Shen, L.~Tian, S.~Wang, J.~Wang, and Q.~Tian, ``Scalable person
  re-identification: A benchmark,'' in \emph{ICCV}, 2015.

\bibitem{zheng2017unlabeled}
Z.~Zheng, L.~Zheng, and Y.~Yang, ``Unlabeled samples generated by gan improve
  the person re-identification baseline in vitro,'' in \emph{ICCV}, 2017.

\bibitem{liu2016deep}
X.~Liu, W.~Liu, T.~Mei, and H.~Ma, ``A deep learning-based approach to
  progressive vehicle re-identification for urban surveillance,'' in
  \emph{ECCV}, 2016.

\bibitem{liao2015person}
S.~Liao, Y.~Hu, X.~Zhu, and S.~Z. Li, ``Person re-identification by local
  maximal occurrence representation and metric learning,'' in \emph{CVPR},
  2015.

\bibitem{huang2017densely}
G.~Huang, Z.~Liu, L.~Van Der~Maaten, and K.~Q. Weinberger, ``Densely connected
  convolutional networks,'' in \emph{CVPR}, 2017.

\bibitem{hu2018squeeze}
J.~Hu, L.~Shen, and G.~Sun, ``Squeeze-and-excitation networks,'' in
  \emph{CVPR}, 2018.

\bibitem{dong2016accelerating}
C.~Dong, C.~C. Loy, and X.~Tang, ``Accelerating the super-resolution
  convolutional neural network,'' in \emph{ECCV}, 2016.

\bibitem{Li-2017-IJCAI}
W.~Li, X.~Zhu, and S.~Gong, ``Person re-identification by deep joint learning
  of multi-loss classification,'' \emph{IJCAI}, 2017.

\bibitem{Hermans-2017-arXiv}
A.~Hermans, L.~Beyer, and B.~Leibe, ``In defense of the triplet loss for person
  re-identification,'' \emph{arXiv}, 2017.

\bibitem{Zhang-2018-CVPR}
Y.~Zhang, T.~Xiang, T.~M. Hospedales, and H.~Lu, ``Deep mutual learning,'' in
  \emph{CVPR}, 2018.

\bibitem{Song-2018-CVPR}
C.~Song, Y.~Huang, W.~Ouyang, and L.~Wang, ``Mask-guided contrastive attention
  model for person re-identification,'' in \emph{CVPR}, 2018.

\bibitem{Cheng-2016-CVPR}
D.~Cheng, Y.~Gong, S.~Zhou, J.~Wang, and N.~Zheng, ``Person re-identification
  by multi-channel parts-based cnn with improved triplet loss function,'' in
  \emph{CVPR}, 2016.

\bibitem{Zheng-2018-TCSVT}
Z.~Zheng, L.~Zheng, and Y.~Yang, ``Pedestrian alignment network for large-scale
  person re-identification,'' \emph{TCSVT}, 2018.

\bibitem{Zhong-2018-CVPR}
Z.~Zhong, L.~Zheng, Z.~Zheng, S.~Li, and Y.~Yang, ``Camera style adaptation for
  person re-identification,'' in \emph{CVPR}, 2018.

\bibitem{zhang2017alignedreid}
X.~Zhang, H.~Luo, X.~Fan, W.~Xiang, Y.~Sun, Q.~Xiao, W.~Jiang, C.~Zhang, and
  J.~Sun, ``Alignedreid: Surpassing human-level performance in person
  re-identification,'' \emph{arXiv}, 2017.

\bibitem{sun2017svdnet}
Y.~Sun, L.~Zheng, W.~Deng, and S.~Wang, ``Svdnet for pedestrian retrieval,''
  \emph{arXiv}, 2017.

\bibitem{qian2017pose}
X.~Qian, Y.~Fu, T.~Xiang, W.~Wang, J.~Qiu, Y.~Wu, Y.-G. Jiang, and X.~Xue,
  ``Pose-normalized image generation for person re-identification,'' in
  \emph{ECCV}, 2018.

\bibitem{zhang2018unreasonable}
R.~Zhang, A.~A. Efros, E.~Shechtman, and O.~Wang, ``The unreasonable
  effectiveness of deep features as a perceptual metric,'' in \emph{CVPR},
  2018.

\bibitem{krizhevsky2012imagenet}
A.~Krizhevsky, I.~Sutskever, and G.~E. Hinton, ``Imagenet classification with
  deep convolutional neural networks,'' in \emph{NeurIPS}, 2012.

\end{thebibliography}

%\vspace{-17mm}
% \clearpage
%\newpage

\begin{IEEEbiography}[{\includegraphics[width=1in,height=1.25in,clip,keepaspectratio]{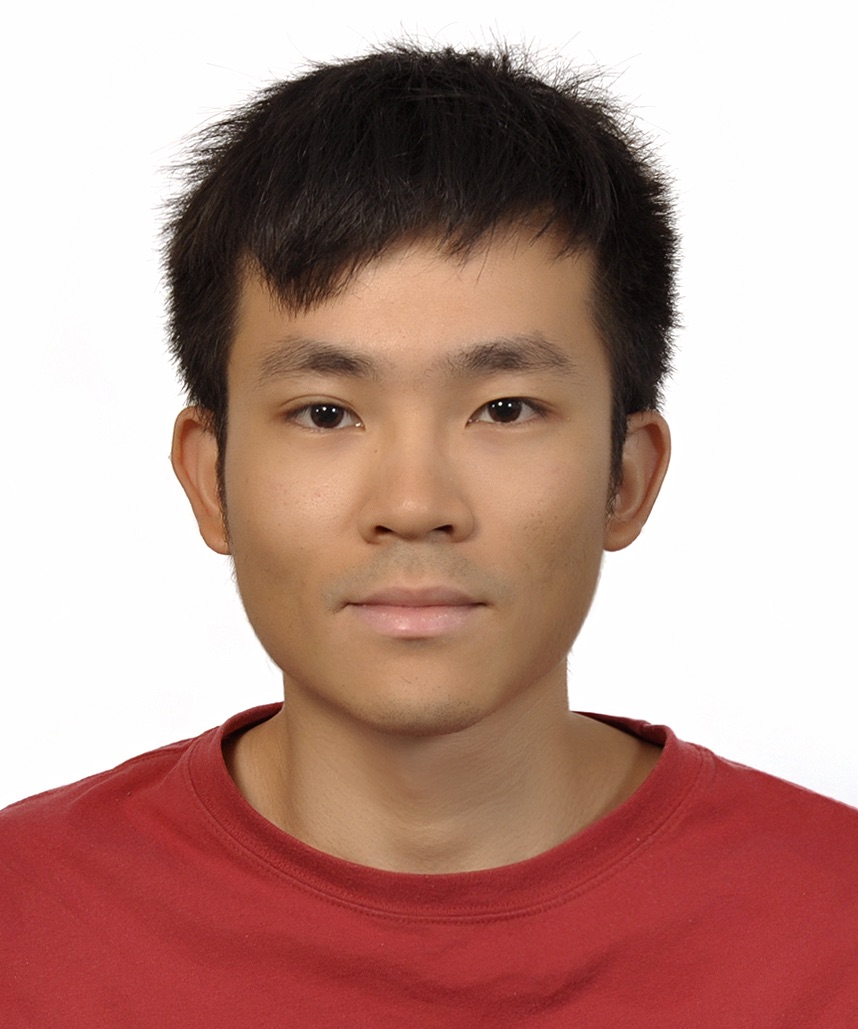}}]{Yu-Jhe Li} received the M.S. degree in Communication Engineering from National Taiwan University and the B.S. degree in Electrical Engineering and Computer Science from National Tsing Hua University in 2019 and 2017, respectively. His current research interests include computer vision and machine learning.
\end{IEEEbiography}
\begin{IEEEbiography}[{\includegraphics[width=1in,height=1.25in,clip,keepaspectratio]{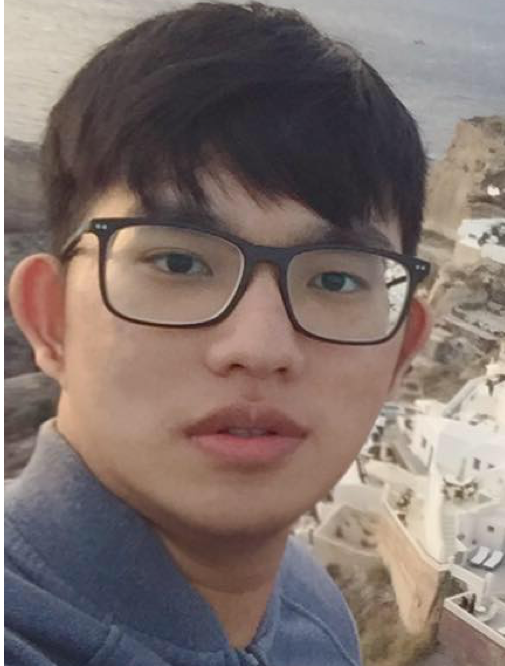}}]{Yun-Chun Chen} received the B.S. degree in Electrical Engineering from National Taiwan University, Taipei, Taiwan in 2018. He is currently a Ph.D. student in the Department of Computer Science at the University of Toronto. His research interests include computer vision, robotics, and machine learning.
\end{IEEEbiography}
\begin{IEEEbiography}[{\includegraphics[width=1in,height=1.25in,clip,keepaspectratio]{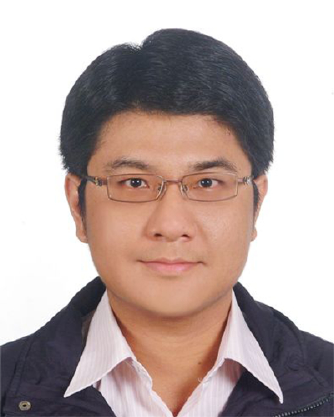}}]{Yen-Yu Lin} received the B.B.A. degree in Information Management, and the M.S. and Ph.D. degrees in Computer Science and Information Engineering from National Taiwan University in 2001, 2003, and 2010, respectively. He is a Professor with the Department of Computer Science, National Chiao Tung University since August 2019. Prior to that, he worked for the Research Center for Information Technology Innovation, Academia Sinica from January 2011 to July 2019. His research interests include computer vision, machine learning, and artificial intelligence.
\end{IEEEbiography}
\begin{IEEEbiography}[{\includegraphics[width=1in,height=1.25in,clip,keepaspectratio]{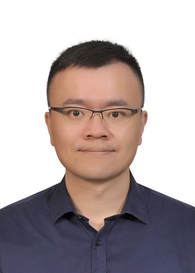}}]{Yu-Chiang Frank Wang} received the B.S. degree in Electrical Engineering from National Taiwan University, Taipei, Taiwan, in 2001, and the M.S. and Ph.D. degrees in Electrical and Computer Engineering from Carnegie Mellon University, Pittsburgh, PA, USA, in 2004 and 2009, respectively. He joined the Research Center for Information Technology Innovation (CITI), Academia Sinica, Taiwan, in 2009, as an Assistant Research Fellow, and later he was promoted to Associate Research Fellow, in 2013. From 2015 to 2017, he was the Deputy Director of CITI, Academia Sinica. He joins the Graduate Institute of Communication Engineering, Department of Electrical Engineering, National Taiwan University, Taipei, in 2017. He leads the Vision and Learning Lab, NTU, where he focuses on research topics in computer vision and machine learning. 
\end{IEEEbiography}

\end{document}